\documentclass[10pt,twocolumn,letterpaper]{article}

\usepackage{iccv}
\usepackage{times}
\usepackage{epsfig}
\usepackage{graphicx}
\usepackage{amsmath}
\usepackage{amssymb}
\usepackage{color}

\usepackage{caption}
\usepackage{subcaption}
\usepackage{booktabs}
\usepackage{multirow}
\usepackage{bm}
\usepackage[inline]{enumitem}
\usepackage{amsthm}
\usepackage{array}
\usepackage{dirtytalk}
\usepackage{appendix}

\newcommand{\myparagraph}[1]{
\vspace{0.1cm}\noindent
\textbf{#1}
\hspace{0.1cm}
}
 
\theoremstyle{definition}
\newtheorem{definition}{Definition}

\newcolumntype{V}{>{\centering\arraybackslash} m{.4\linewidth} }

\usepackage[pagebackref=true,breaklinks=true,letterpaper=true,bookmarks=false]{hyperref}

\iccvfinalcopy 

\begin{document}

\title{Towards a Visual Privacy Advisor:\\Understanding and Predicting Privacy Risks in Images}

\author{Tribhuvanesh Orekondy \qquad   Bernt Schiele \qquad Mario Fritz\vspace{0.5cm} \\
Max Planck Institute for Informatics\\
Saarland Informatics Campus\\
Saabr\"ucken, Germany\\
{\tt\small \{orekondy,schiele,mfritz\}@mpi-inf.mpg.de}
}

\maketitle

\begin{abstract}

With an increasing number of users sharing information online, privacy implications entailing such actions are a major concern.
For explicit content, such as user profile or GPS data, devices (\eg mobile phones) as well as web services (\eg facebook) offer to set privacy settings in order to enforce the users' privacy preferences.

We propose the first approach that extends this concept to image content in the spirit of a {\em Visual Privacy Advisor}.  First, we categorize personal information in images into 68 image attributes and collect a dataset, which allows us to train models that predict such information directly from images. Second, we run a user study to understand the privacy preferences of different users w.r.t. such attributes. Third, we propose models that predict user specific privacy score from images in order to enforce the users' privacy preferences.
Our model is trained to predict the user specific privacy risk and even outperforms the judgment of the users, who often fail to follow their own privacy preferences on image data.
\end{abstract}

\section{Introduction}
As more people obtain access to the internet, a large amount of personal information becomes accessible to \eg other users, web service providers and advertisers. To counter these problems, more and more devices (\eg mobile phone) and web services (\eg facebook) are equipped with mechanisms where the user can specify privacy settings to comply with his/her personal privacy preference.

\begin{figure}
\begin{center}
   \includegraphics[width=\linewidth]{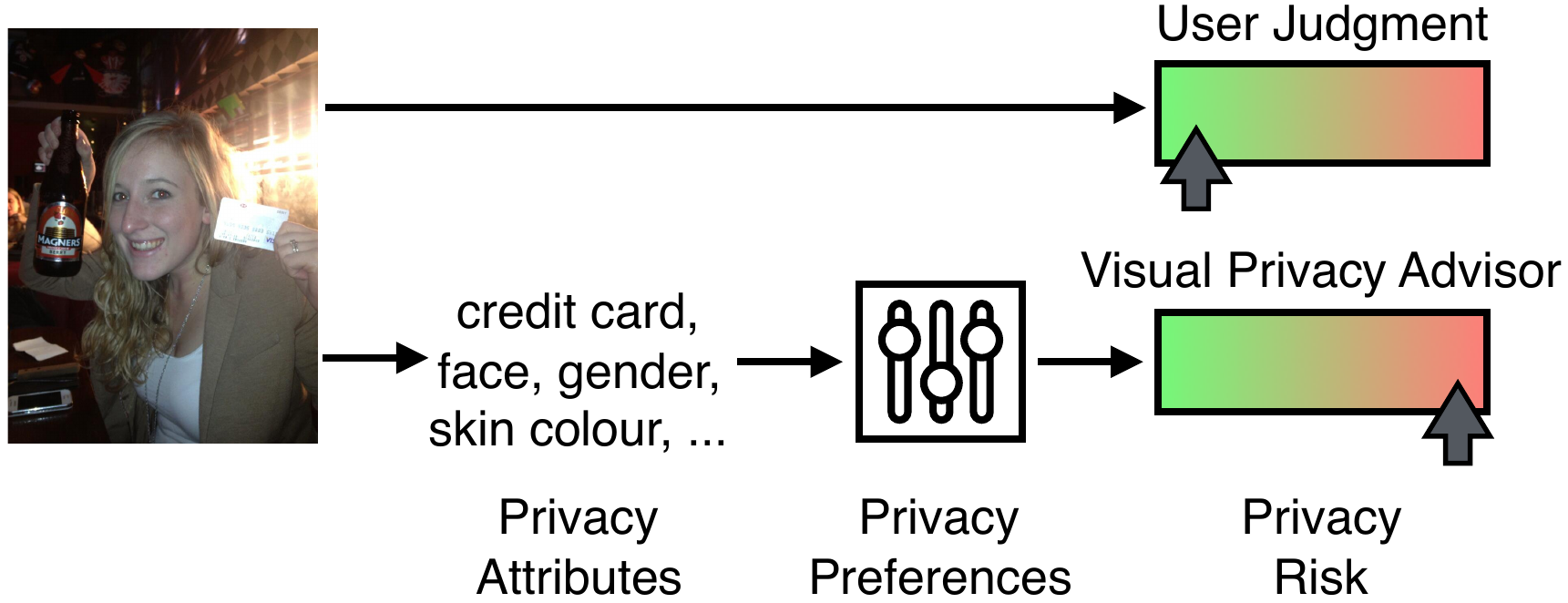}
\end{center}
   \caption{Users often fail to enforce their privacy preferences when sharing images online.
   We propose a first \emph{Visual Privacy Advisor} to provide user-specific privacy feedback.}
\label{fig:intro_ampel}
\end{figure}

While this has proven useful for explicit and textual information, we ask how this concept can generalize to visual content.
While users can be asked (as we also do in our study) to specify how comfortable they are releasing a certain type of image content, the actual presence of such content is implicit in the image and not readily available for a privacy preference enforcing mechanism nor the user. In fact -- as our study shows -- people frequently misjudge the privacy relevant information content in an image -- which leads to the failure of enforcing their own privacy preferences.

Hence, we work towards a {\it Visual Privacy Advisor} (\autoref{fig:intro_ampel}) that helps users enforce their privacy preferences and prevents leakage of private information. 
We approach this complex problem by first making personal information explicit by categorizing personal information into 68 image attributes.
Based on such attribute predictions and user privacy preferences, we infer a privacy score that can be used to prevent unintentional sharing of information.
Our model is trained to predict the user specific privacy risk and interestingly, it outperforms human judgment on the same images.

Our main contributions in this paper are as follows: 
\begin{enumerate*}[label=(\roman*)]
	\item To the best of our knowledge, we are the first to formulate the problem of identifying a diverse set of personal information in images and personalizing predictions to users based on their privacy preferences
	\item We provide a sizable dataset\footnote{Refer to project website: \url{https://tribhuvanesh.github.io/vpa/}} of 22k images annotated with 68 privacy attributes
	\item We conduct a user study and analyze the diversity of users' privacy preferences as well as the level to which they achieve to follow their privacy preferences on image data
	\item We propose the first model for Privacy Attribute Prediction. We also extend it to directly estimate user-specific privacy risks
	\item Finally, we show that our models outperform users in following their own privacy preferences on images
\end{enumerate*}

\section{Related Work}
Privacy is becoming an increasing concern \cite{young2009information, dwyer2007trust}, especially due to the rise of social networking websites allowing individuals to share personal information, without explaining consequences of these actions.
In this section, we discuss work that highlights these concerns and explores consequences of such actions.
We also discuss literature that deals with identifying private content in images and text.

\myparagraph{Identifying Personal Information}
There is a comparably small body of work that aims to recognize personal information.
Aura \etal \cite{aura2006scanning} explore this in the context of electronic documents, where they propose a tool to remove user names, identifiers, organization names and other private information from text-based documents with metadata. \cite{bier2014detection,geng2008using} study this in the context of textual email-content.
Bier \etal \cite{bier2014detection} model this as a privacy-classification problem, whereas  Geng \etal \cite{geng2008using} detect four types of personal information -- email addresses, telephone numbers, addresses and money.
The closest related work to ours is \cite{tonge2015privacy}, who are also motivated by unwanted disclosure and privacy violation on social media.
They approach the task as classifying if an image is public or private based on features extracted from a Convolutional Neural Network and user-generated tags for the image.
However, we later show that users have different notions of privacy and hence cannot be modeled as a binary classification problem.
Instead, we first tackle a more principled problem of predicting the privacy-sensitive elements present in images and use these in combination with users preferences to estimate privacy risk.

\myparagraph{Leakage and De-anonymization}
A problem closely related to ours is \emph{privacy leakage}, which deals with uncovering and analyzing methods leading to disclosure of personal information, rather than detection before such incidents.
\cite{krishnamurthy2009leakage, Krishnamurthy2011PrivacyLV} uncover privacy leakage when websites accidentally provide user information embedded in HTTP requests when contacting third-party aggregators.
As leakages can be user-intended, Yang \etal \cite{Yang2013AppIntentAS} explore this case in Android applications.
Some works \cite{Narayanan2009DeanonymizingSN, veiga2016privacy} study the case where users identity, location or other details can be de-anonymized when aggregating anonymized data across multiple social networks.
In contrast to these,  our approach is concerned about image content and privacy preferences.

\myparagraph{Privacy Preferences and Social Networks}
\cite{lenhart2007teens, gross2005information, krishnamurthy2008characterizing} study types of personal information disclosed on social networking websites.
Other tasks include preserving one's privacy while using social networks \cite{guha2008noyb, zhou2008preserving, li2011findu} and exploring privacy settings \cite{fang2010privacy, danezis2009inferring, liu2011analyzing}.
However in our user study, apart from collecting and analyzing user studies on privacy preferences for images, we additionally use them to train models based on image data.

\myparagraph{Privacy and Computer Vision}
Several works explore detecting individual privacy attributes such as license plates \cite{Zhou2012PrincipalVW, Zhang2006LearningBasedLP, Chang2004AutomaticLP}, age estimation from facial photographs \cite{Bauckhage2010AgeRI}, social relationships \cite{Wang2010SeeingPI}, face detection \cite{Sun2017FaceDU, Viola2001RobustRF}, landmark detection \cite{Zheng2009TourTW} and occupation recognition \cite{shao2013you}.
Apart from detecting attributes, some works introduce new privacy challenges in vision such as adversarial perturbations \cite{moosavi2016universal, papernot2016distillation}, privacy-preserving video capture \cite{aditya2016pic, pittaluga2015privacy, neustaedter2006blur, raval2014markit}, person re-identification \cite{ahmed2015improved, mclaughlin2016recurrent}, avoiding face detection \cite{wilber2016can, harvey2012cv}, full body re-identification \cite{oh2016faceless} and privacy-sensitive lifelogging \cite{Hoyle:2015:SLP:2702123.2702183,screenavoider2016chi}.
In this work, we present a new challenge in computer vision designed to help users assess privacy risk before sharing images on social media that encompasses a broad range of personal information in a single study.

\myparagraph{Datasets for Privacy Tasks}
Crucial to exploring privacy tasks are images revealing private details such as faces, names or opinions.
However, many available datasets do not contain a significant number of such images to effectively study privacy tasks.
Although some datasets \cite{gallagher_cvpr_08_clothing} contain such information, they are either too small or not representative of images on social networks.
The closest candidate is the PIPA dataset \cite{Zhang_2015_CVPR} with 37,107 Flickr images, proposed for people recognition in an unconstrained setting and does not include images covering many other privacy aspects such as license plates, political views or official identification documents.
In this paper, we introduce the first dataset of real-life images capturing important privacy-relevant attributes.

\section{The Visual Privacy  (VISPR) Dataset}
\label{sec:papds}

Mobile devices and social media platforms provide privacy settings, so that users can communicate their privacy preferences on the disclosure of different type of textual information. 
How does this concept transfer to image data?
We need to establish a similar concept of privacy relevant information types -- but now for {\it images}.
This will allow us to query users about their privacy preferences on the disclosure of various information types, as we will do in the next section.

Therefore, we propose in this section a categorization of personal information into 68 privacy attributes such as gender, tattoo, email address or fingerprint.
We collect a dataset of ~22k images that allows the study of privacy relevant attributes in images and the training of automatic recognizers.

\begin{figure*}
\begin{center}
\includegraphics[width=\textwidth]{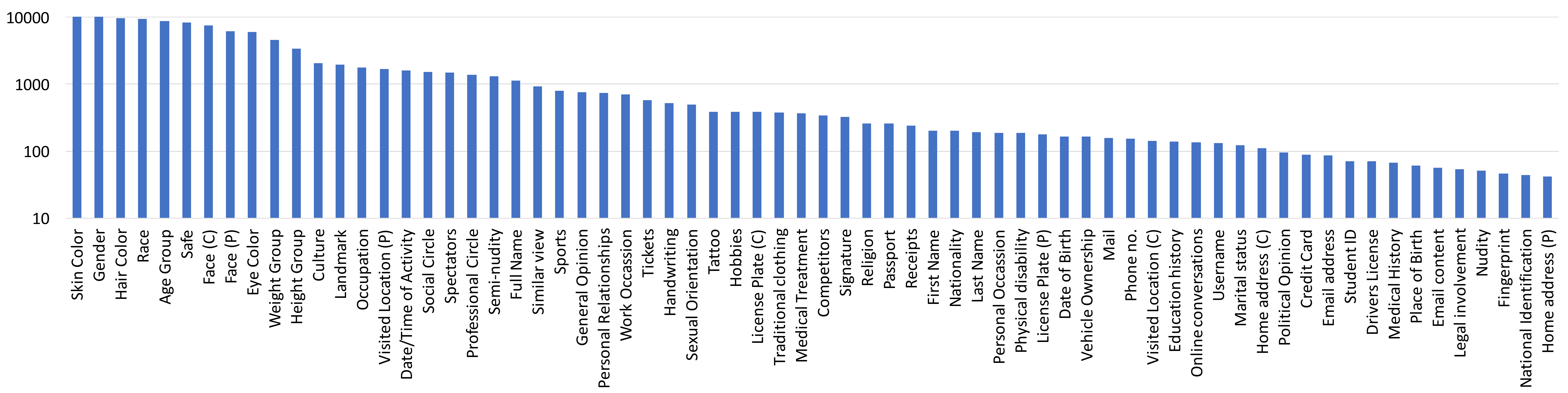}
\end{center}
   \caption{Label distribution in our dataset. $Y$-axis indicates the number of images.}
\label{fig:dataset_label_distribution}
\end{figure*}

\subsubsection*{Privacy Attributes}
As motivated before, we need to categorize different types of personal content in images -- akin to the privacy settings deployed in today's devices and services.
Therefore, we define a list of \emph{privacy  attributes} an image can disclose.

The primary challenge here is the lack of a standard list of privacy attributes.
We thus compile attributes from multiple sources.
First, we consolidate relevant attributes from the guidelines for handling \emph{Personally Identifiable Information} \cite{mccallister2010guide} provided in the EU Data Protection Directive 95/46/EC \cite{directive199595} and the US Privacy Act of 1974.
Second, we add relevant attributes from the rules on prohibiting sharing personal information on various social networking websites (\eg, Twitter, Reddit, Flickr).
Finally, we manually examine images that are shared on these websites and identify additional attributes.
As a result, we draft an initial set of 104 potential privacy attributes.
As discussed in the next section, these are reduced to 68 attributes (see \autoref{tab:dataset_statistics}) after pruning.

\subsubsection*{Annotation Setup}
The annotation was set up as a multi-label task to three annotators annotating independent sets of images.
A web-based tool was provided to select multiple options corresponding to the 104 privacy attributes per image.
Additionally, annotators could mark if they were unsure about their annotation.
In case none of the provided privacy labels applied, they were instructed to label the image as \emph{safe}, which we use as one of our privacy attributes.
Images were discarded if annotators were unsure, or
if the image contained a copyright watermark, 
was a historic photograph,
contained primarily non-English text,
or was of poor quality.

\subsubsection*{Data Collection and Annotation Procedure}
In this section, we discuss the steps taken to obtain the final set of 22k images annotated with 68 privacy attributes.

\myparagraph{Seed Sample}
We first gather 100k random images from the OpenImages dataset \cite{openimages}, a collection of $\sim$9 million Flickr images.
Using the definition and examples of the privacy attributes, the annotators annotate 10,000 images randomly selected from the downloaded images.

\myparagraph{Handling Imbalance}
Based on the label statistics from these 10,000 images, we add images to balance attributes with fewer than 100 occurrences.
These additional images are added by querying relevant OpenImages labels possibly representative of insufficient privacy attributes.

\myparagraph{Extended Search for Rare Classes}
In spite of using the above strategy, 37 attributes contain under 40 images.
We manually add images for these attributes by querying relevant keywords on Flickr.
We do not add multiple images from the same album.
For credit cards, we manually obtain ~50 high-quality images from Twitter, which are the only non-Flickr images in our dataset.

\myparagraph{Selected Attributes}
After annotating the dataset with the initial 104 labels, we discard 19 labels because either
\begin{enumerate*}[label=(\roman*)]
	\item images were difficult to obtain manually (\eg iris/retinal scan, insurance 	details) or
	\item the set of images did not clearly represent the attribute.
\end{enumerate*}
We additionally merge groups of attributes which capture similar concepts (\eg work and home phone number).
In the end, we obtain a dataset of 22,167 images, each annotated with one or more of 68 privacy attributes.

\myparagraph{Curation}
To reduce labeling mistakes, we organize the dataset into batches of images with each batch corresponding to a privacy attribute.
We curate attribute batches which either contain fewer than 500 images or are considered sensitive by users.

\myparagraph{Splits}
We perform a random 45-20-35 split with 10,000 training, 4,167 validation and 8,000 test images.
The final statistics of our dataset is presented in \autoref{tab:dataset_statistics}.
The labels and its distribution in our dataset is shown in \autoref{fig:dataset_label_distribution}.


\begin{table}
\centering
  \begin{tabular}{@{}lrrrr@{}}
  \toprule
  Split                   & All         & Train      & Val         & Test        \\ \midrule
  Images                  & 22,167      & 10,000     & 4,167       & 8,000       \\
  Labels                  & 115,742     & 51,799     & 22,026      & 41,917      \\
  Avg Labels/Image & 5.22 & 5.18 & 5.29 & 5.24 \\
  Max Images/Label      & 10,460      & 4,710      & 1,969       & 3,781       \\
  Min Images/Label      & 44          & 20         & 7           & 12          \\ \bottomrule
  \end{tabular}%
\caption{Dataset Statistics}
\label{tab:dataset_statistics}
\end{table}

\section{Understanding Privacy Risks}
\label{sec:papds_user_preferences}
In this section, we explore how users' personal privacy preferences relate to the attributes in Section \ref{sec:privacy_preferences}.
Furthermore, we study how good users are at enforcing their own privacy preferences on visual data when making judgments based on image data in Section \ref{sec:user_study_img}.

\subsection{Understanding Users' Privacy Preferences}
\label{sec:privacy_preferences}
In this section, we study the degree to which various users are sensitive to the privacy attributes discussed in Section \ref{sec:papds}.

\myparagraph{User Study} 
We present each user with a series of 72 questions in a randomized order.
Each of these questions corresponds to either exactly one of 67 privacy attributes (excluding the safe attribute) or a control question.
In each question, the users are asked how much they would find their privacy violated if they accidentally shared details of a particular attribute publicly online.
For instance:
\say{How much would you find your privacy violated if you accidentally shared details on personal occasions you have attended (like a birthday party or friend's wedding).}
Responses for the question are collected on a scale of 1 to 5, where:
\begin{enumerate*}[label=(\arabic*)]
	\item Privacy is not violated
    \item Privacy is slightly violated
    \item Privacy is somewhat violated
    \item Privacy is violated
    \item Privacy is extremely violated.
\end{enumerate*}
We treat these responses as users privacy preference for this particular privacy attribute.

\myparagraph{Participants}
We collect responses of 305 unique AMT workers in this survey.
Out of the 305 respondents, 59\% were male, 78\% were under 40 years of age with 57\%  from USA and 38\% from India.
Additionally, 75\% were regular Facebook users, 80\% and 44\% reported to be aware of and have used Twitter and Flickr at least once.

\begin{figure*}
\begin{center}
   \includegraphics[width=\textwidth]{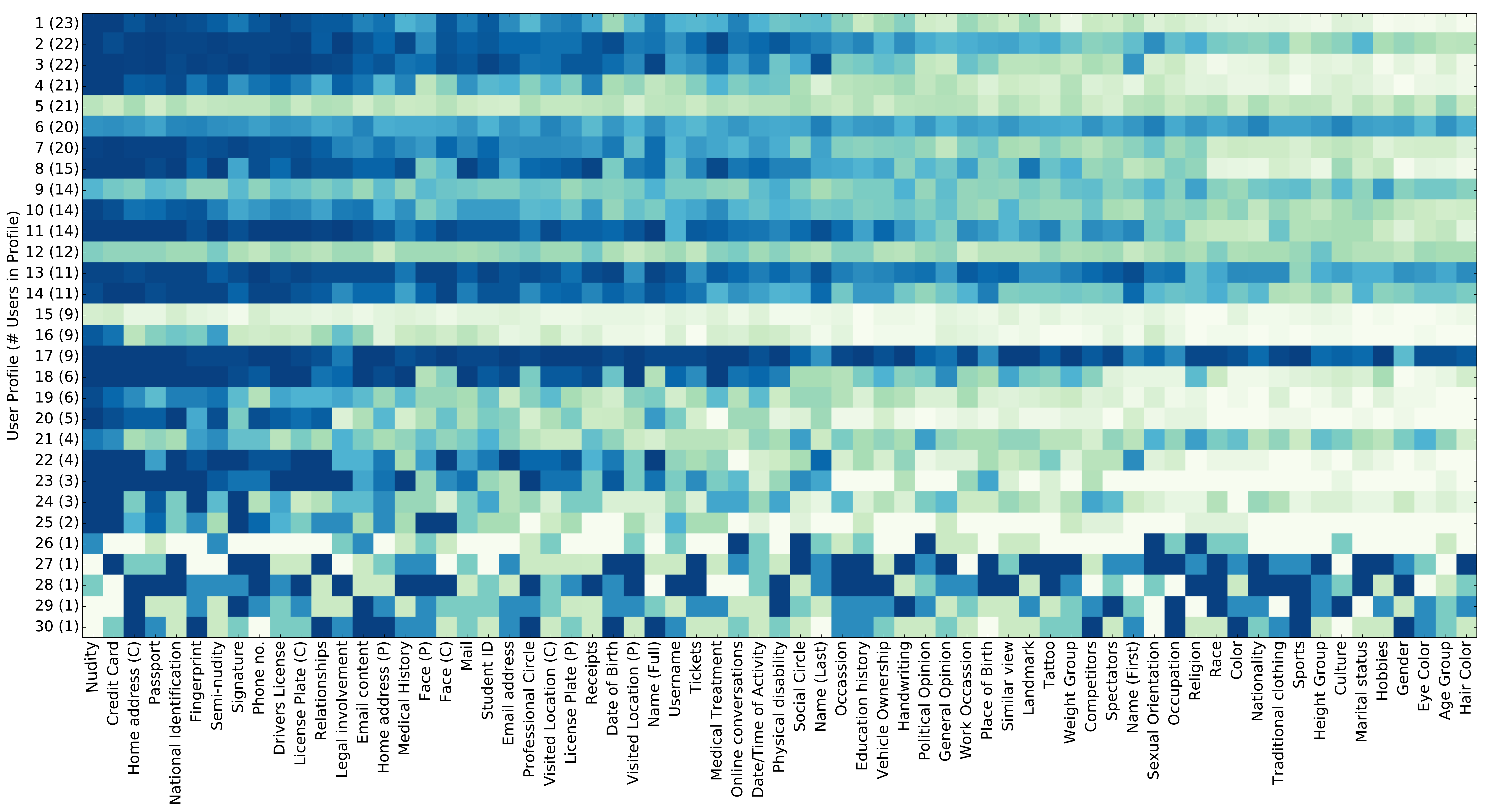}
\end{center}
   \caption{Privacy preferences of user profiles for the privacy attributes.
   Darker colors represent higher privacy-sensitivity to attributes.
   Each row corresponds to one of the 30 profiles and the number in brackets on the $Y$-axis represents the number of users mapped to the profile.
   Rows are ordered based on number of users linked to the profile. }
\label{fig:user_profiles}
\end{figure*}

\myparagraph{Analysis}
In order to understand the diversity in users' privacy preferences, we first cluster the users based on their preferences into \emph{user privacy profiles}.
We cluster using $K$-means and choose $K$ based on silhouette score \cite{rousseeuw1987silhouettes}, which considers distance between points within the cluster and additionally distance between points and their neighbouring cluster.
We choose $K=30$ as this yields the lowest silhouette score.
This enables visualizing the preferences over the attributes, as seen in \autoref{fig:user_profiles}, where each row represents the preferences for one of the 30 user profiles (ordered based on number of users associated with the profile).
We observe from this study:
\begin{enumerate*}[label=(\roman*)]
	\item Users show a wide variety of preferences. This supports requiring user-specific privacy risk predictions.
    \item The \emph{majority} (Profiles 1-4, 7-11, 13-14, 18-20 in \autoref{fig:user_profiles}) display a similar order of sensitivity to the attributes
    \item A \emph{minority} (Profiles 21-30) of users are particularly sensitive to some attributes such as their political view, sexual orientation or religion
	\item The \emph{uniformly-sensitive users} (Profiles 5, 6, 12, 15, 17) are uniformly sensitive to all attributes even though to different degrees.
\end{enumerate*}

\subsection{Users and Visual Privacy Judgment}
\label{sec:user_study_img} 
In this study, we first ask participants to judge their personal privacy risk based on images representing an attribute (providing a visual privacy risk score) and afterwards asking the actual user's privacy preferences for the same attribute (providing a desired or explicit privacy risk score).
Hence, we study how good users are at assessing their personal privacy risks based on images.

\myparagraph{User Study} 
In this study, we split the survey into two parts.
In the first part, the users are shown a group of 3-6 images.
Given the sensitive nature of attributes, we cannot obtain or ask users to rate their personal images and hence use images from the dataset.
They are asked how comfortable they are sharing such images publicly, considering they are the subject in these images.
Responses are collected on a scale of 1 to 5, where:
\begin{enumerate*}[label=(\arabic*)]
	\item Extremely comfortable
    \item Slightly comfortable
    \item Somewhat comfortable
    \item Not comfortable
    \item Extremely uncomfortable.
\end{enumerate*}
Each group of images represents one of the 68 privacy attributes.
In most cases, the attributes occur isolated and are the most prominent visual cue in the image.
We refer to these responses as \emph{human visual privacy score}.
The second part is identical to questions and the setting in the previous user-study on privacy preferences.
Each question is designed to obtain the privacy preference of the user for each attribute.
As before, the user rates on a scale of 1 (Not Violated) to 5 (Extremely Violated).
We refer to these responses as \emph{privacy preference score}.

\myparagraph{Participants}
We split the study into two parts to prevent user fatigue.
Each part contains only half of the attributes.
We obtain 50 unique responses for this survey from AMT.
In each of these parts, roughly: 70\%  of the respondents were under 40 years, 57\% were male and 87\% were from USA.
Additionally, 80\% responded that they use Facebook, 84\% Twitter and 46\% Flickr.

\begin{figure}
\begin{center}
   \includegraphics[width=0.7\linewidth]{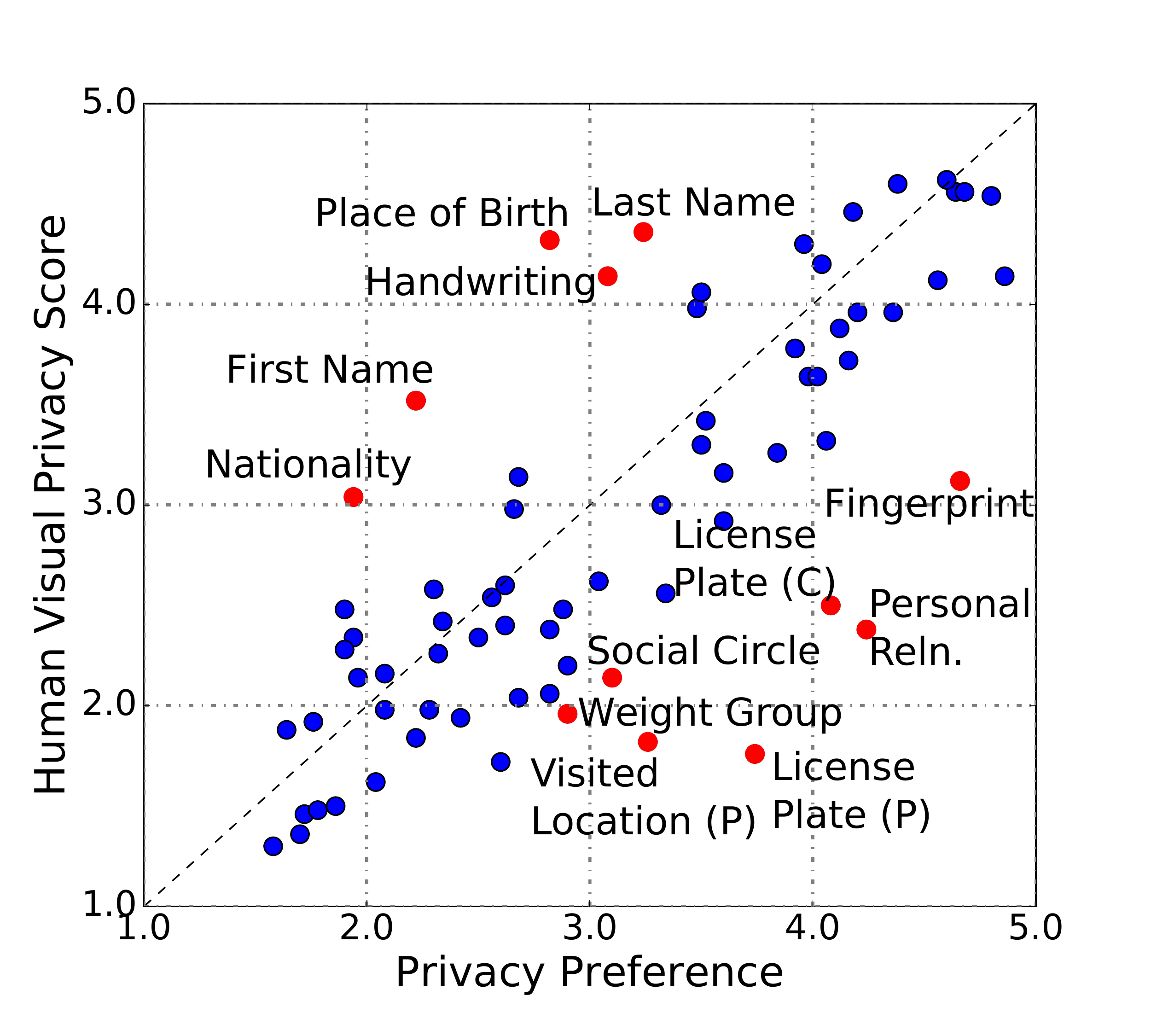}
\end{center}
   \caption{Users are asked to rate on a scale of 1 (Not violated) to 5 (Extremely violated) how much an attribute affects their privacy.
   $X$-axis denotes their desired privacy preference and $Y$-axis denotes their evaluation of risk on images.
   The red markers indicate privacy attributes with highly underestimated or overestimated user ratings}
\label{fig:img_vs_attr}
\end{figure}

\myparagraph{Analysis}
We compute for each attribute average privacy preference score and human visual scores, and visualized them as a scatter plot in \autoref{fig:img_vs_attr}.
From the results, we observe:
\begin{enumerate*}[label=(\roman*)]
	\item The off-diagonal data points show  a clear inconsistency in the users between the required privacy preference and their judgment of privacy risk in images.
    \item For cases close to the diagonal, like credit cards, passport and national identification documents, users display consistent behaviour on images and attributes.
    \item However, when photographs are natural scenes containing people or vehicles, users underestimate (below diagonal) the privacy score, such as in the case of family photographs or cars displaying license plate numbers.
    We speculate this is indicative of personal photographs commonly shared online.
    \item They overestimate (above diagonal) the privacy risk of some photographs showing birth place or their name. 
    We speculate this is because the photographs are often official documents, making users more cautious.
\end{enumerate*}

\section{Predicting Privacy Risks}
\label{sec:pap}

\begin{figure}
\begin{center}
   \includegraphics[width=1.0\linewidth]{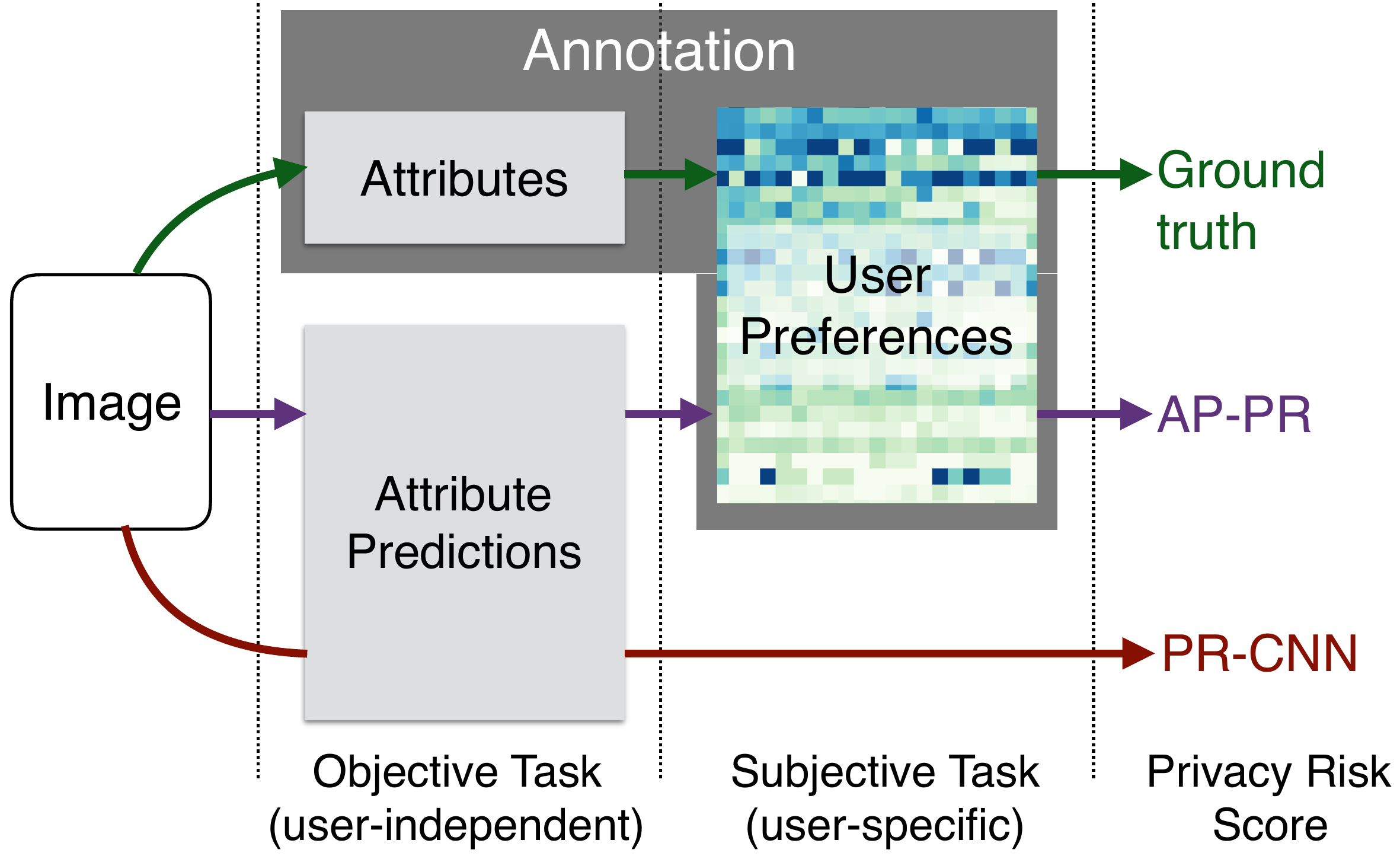}
\end{center}
   \caption{We learn an end-to-end model for user-specific privacy risk estimation.}
\label{fig:ppp_prcnn_overview}
\end{figure}

In this section, we make a step towards our overall goal of a {\it Visual Privacy Advisor}. 
As illustrated in \autoref{fig:ppp_prcnn_overview}, we follow a similar paradigm \eg on social networks that defines privacy risk based on both the content type and user-specific privacy settings.
In our case, the content type is described by (user-independent) attributes in the previous section. We combine these with the user-specific privacy preferences to determine if the image contains a privacy violation. 

We describe our model for privacy attribute prediction in Section \ref{sec:pap}, followed by our approaches to personalized privacy risk prediction in Section \ref{sec:pprp}.
We conclude with a comparison of human judgment of privacy risks in images against the prediction of our proposed models in Section \ref{sec:hvm}.

\label{sec:prs}

\subsection{Privacy Attribute Prediction}
\label{sec:pap}

In this section, we define the \emph{user-independent} task of predicting privacy attributes from images.
Then, we present and evaluate different methods on our new VISPR dataset.

\myparagraph{Task}
We propose the task of \emph{Privacy Attribute Prediction},  which is to predict one or more of 68 privacy attributes based on an image.
This can be seen as a multilabel classification problem that recognizes different type of personal information visual data and therefore has the potential to make this information explicit.
\autoref{fig:intro_ampel} shows multiple examples for this task.
The task is challenging due to image diversity, subtle cues and high level semantics.

\myparagraph{Metric}
To assess performance of methods for this task, we compute the Average Precision (AP) per class, which is the area under Precision-Recall curve for the attribute.
Additionally, the overall performance of a method is given by Class-based Mean Average Precision (C-MAP), the average of the AP score across all 68 attributes.

\begin{figure*}
\centering
\begin{tabular}{ @{} c  >{\centering\arraybackslash} m{4cm}  >{\centering\arraybackslash} m{4cm} >{\centering\arraybackslash} m{4cm} @{}}
 & True Positives & False Positives & False Negatives \\
 & & & \\
Tattoo 
&  \includegraphics[width=0.1\textwidth]{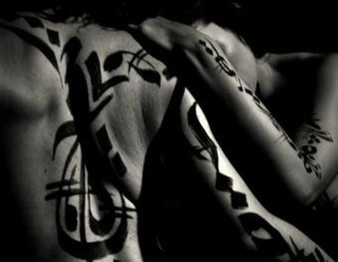}
   \includegraphics[width=0.1\textwidth]{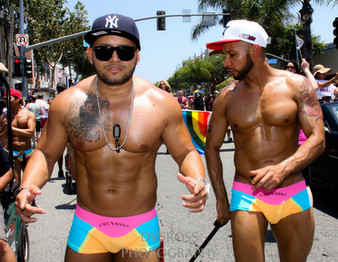} \qquad 
&  \includegraphics[width=0.1\textwidth]{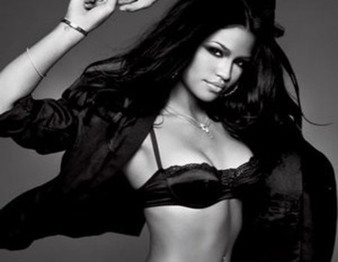}
   \includegraphics[width=0.1\textwidth]{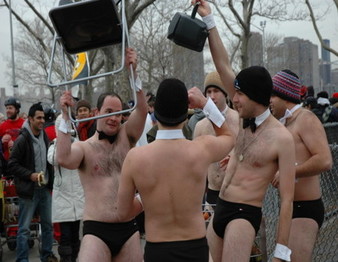}
&  \includegraphics[width=0.1\textwidth]{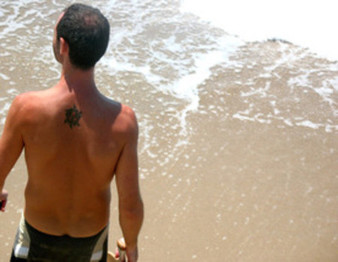}
   \includegraphics[width=0.1\textwidth]{fig/pap_qual/tattoo_fn_2.jpg} \\
\begin{tabular}[c]{@{}l@{}}Physical\\ Disability\end{tabular} 
&  \includegraphics[width=0.1\textwidth]{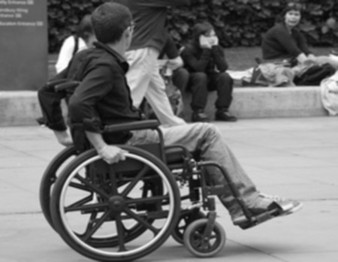}
   \includegraphics[width=0.1\textwidth]{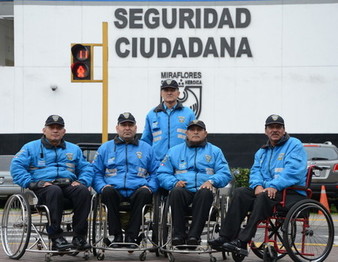} \qquad
&  \includegraphics[width=0.1\textwidth]{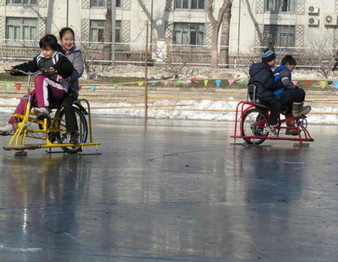}
   \includegraphics[width=0.1\textwidth]{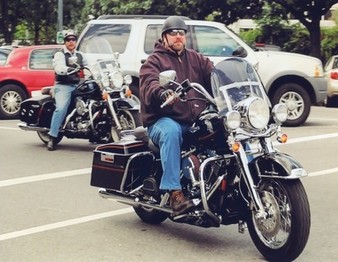}
&  \includegraphics[width=0.1\textwidth]{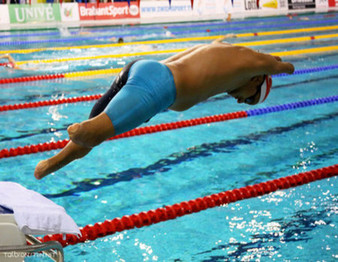}
   \includegraphics[width=0.1\textwidth]{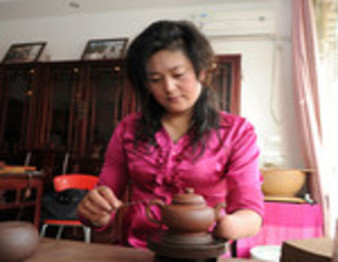} \\
Landmark 
&  \includegraphics[width=0.1\textwidth]{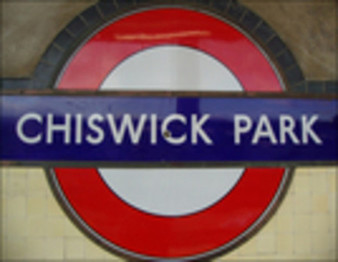}
   \includegraphics[width=0.1\textwidth]{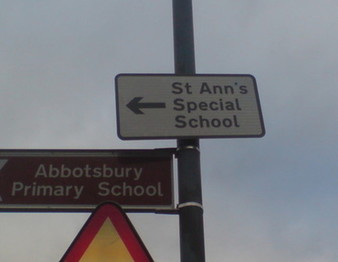} \qquad
&  \includegraphics[width=0.1\textwidth]{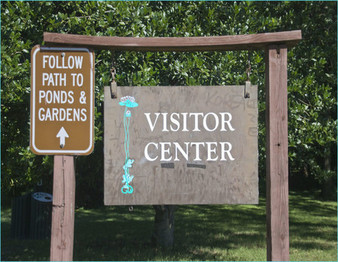}
   \includegraphics[width=0.1\textwidth]{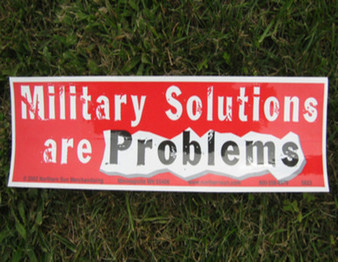}
&  \includegraphics[width=0.1\textwidth]{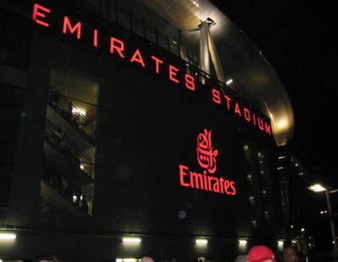}
   \includegraphics[width=0.1\textwidth]{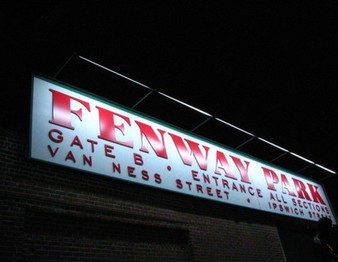} \\
\end{tabular}
\caption{Qualitative Results of our Privacy Attribute Prediction method}
\label{fig:pap_qualitative}
\end{figure*}

\begin{table}
\centering
\begin{tabular}{@{}lll@{}}
\toprule
Training                    & Features  & C-MAP \\ \midrule
\multirow{3}{*}{SVM}        & CaffeNet  & 37.93 \\
                            & GoogleNet & 39.88 \\
                            & Resnet-50 & 40.50  \\ \midrule
\multirow{3}{*}{End-to-End} & CaffeNet  & 42.99 \\
                            & GoogleNet & 43.29 \\
                            & Resnet-50 & 47.45 \\ \bottomrule
\end{tabular}
\caption{Accuracy of our methods given by Class-based Mean Average Precision, evaluated on test}
\label{tab:pap_cmap}
\end{table}

\myparagraph{Methods}
\label{sec:pap_methods_eval}
We experiment with three types of visual features extracted from CNNs -- CaffeNet \cite{jia2014caffe}, GoogleNet \cite{szegedy2015going} and ResNet-50 \cite{he2016deep}.
First, we train a linear SVM model using features from the layer preceding the last fully-connected layer of these CNNs.
In a pilot study, we found that the multilabel SVM with smoothed hinge loss \cite{lapin2016loss} yields better results than SVM multi-label prediction \cite{crammer2003family} and cross-entropy loss.
Second, we fine-tune the CNNs initialized with pretrained ImageNet models, based on a multi-label classification loss with sigmoid activations.

\begin{figure*}
\begin{center}
   \includegraphics[width=\textwidth]{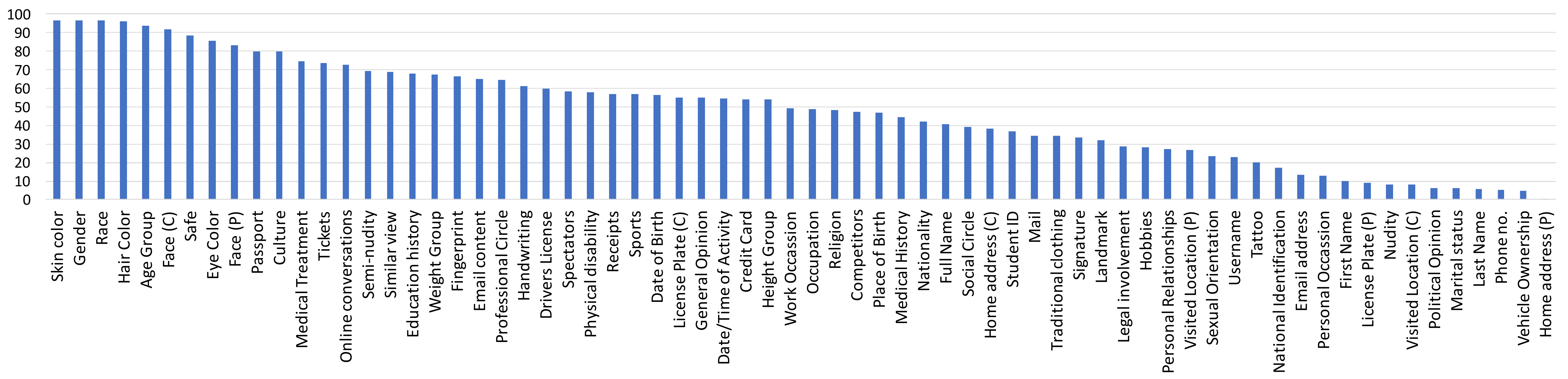}
\end{center}
   \caption{Average Precision (AP) Scores for the privacy attributes from our method}
\label{fig:pap_prec_recall}
\end{figure*}

\myparagraph{Results}
Quantitative results of our method are shown in \autoref{tab:pap_cmap} and qualitative results in \autoref{fig:pap_qualitative} (more discussed in supplementary).
We additionally present the Average Precision scores per class in \autoref{fig:pap_prec_recall}.
We make the following observations:
\begin{enumerate*}[label=(\roman*)]
	\item The CNN performs well in attributes such as tickets, passports, medical treatment that correlated well with scenes (\eg airport, hospital).
    It also performs well in recognizing attributes which are human-centric, such as faces, gender and age.
    \item Fine-grained differences cause confusions such as predicting student IDs for drivers licenses or differentiating between street and other signboards.
    \item We observe failure modes due to small details in the image, such as tattoos, marriage rings or a credit card in the hands of a child.
    \item Another shortcoming is not being able to recognize relationship-based attributes (\eg, personal or social relationships, vehicle ownership) which requires reasoning based on interaction of multiple visual cues in an image rather than just their presence.
\end{enumerate*}

\subsection{Personalizing Privacy Risk Prediction}
\label{sec:pprp}
In the previous section, we discussed predicting privacy attributes in images, a task independent of user privacy preferences.
In this section, we investigate \emph{user-specific} visual privacy feedback. 
The goal is to compute a  {\it privacy risk score} per image, representing the risk of privacy leakage for the particular user.

\myparagraph{Task}
As illustrated in \autoref{fig:ppp_prcnn_overview}, we combine privacy attributes (user independent) together with the privacy preferences based on these attributes (user specific) to arrive at the privacy risk score.
As we allow the users to give scores for each attributes based on their privacy preferences, we define the following {\it privacy risk score}.

\theoremstyle{definition}
\begin{definition}{\it Privacy Risk Score.}
For some image $\bm{x}$, attributes $\bm{y} \in [0, 1]^A$ and user preference $\bm{u} \in [0, 5]^A$, the privacy risk score of image $\bm{x}$ containing attributes $\bm{y}$ on user $\bm{u}$ is $\max_a \bm{y}_a\bm{u}_a$
\end{definition}

This represents the user-specific score of the most sensitive attribute, most likely to be present in an image.
As a result, the privacy-risk score is comparable to the preference-score: 1 (Not Sensitive) to 5 (Extremely Sensitive). 
As illustrated in \autoref{fig:ppp_prcnn_overview}, we compute the ground-truth privacy risk score based on ground-truth attribute annotation for an image (represented as a $k$-hot vector $\bm{y} \in \{0, 1\}^A$) and privacy preferences of users.

\myparagraph{Method: Attribute Prediction-Based Privacy Risk (AP-PR)}
Our first method performs Attributed-Based Privacy Risk ({\it AP-PR}) prediction.
As illustrated in \autoref{fig:ppp_prcnn_overview}, we combine the privacy attribute prediction and the profile's privacy preferences (that we can assume as provided by users at test time) to compute the privacy risk score as defined above.

\myparagraph{Method: Privacy Risk CNN (PR-CNN)}
We propose a Privacy Risk CNN ({\it PR-CNN}) that does not directly use the user profile's privacy preferences -- but only indirectly via the ground-truth.
The key observation is that AP-PR scores suffer from erroneous attribute predictions (see \autoref{fig:pap_prec_recall}).
Therefore, we extend the the privacy attribute prediction network by additional fully-connected layers to directly predict the privacy risk score.
A parameter search yielded best results using additional two fully-connected hidden layers of 128 neurons, each followed by sigmoid activations.
We finetune this network from our Googlenet Privacy Attribute Prediction network for 30 user profiles described in Section \ref{sec:papds_user_preferences} and a Euclidean loss.

\begin{table}
\centering
\begin{tabular}{@{}lllllll@{}}
\toprule
         & \multirow{2}{*}{L1-Error}               &  & \multicolumn{4}{c}{MAP}                                           \\ \cmidrule(lr){4-7}
         &        &  & 1+             & 2+             & 3+             & 4+             \\ \midrule
AP-PR & 0.656          &  & \textbf{94.94} & \textbf{94.27} & 87.97          & 77.89          \\
PR-CNN   & \textbf{0.637} &  & 94.35          & 93.65          & \textbf{88.14} & \textbf{78.38} \\ \bottomrule
\end{tabular}
\caption{Evaluation of Personalized Privacy Risk}
\label{tab:ppp_eval}
\end{table}

\begin{figure}[t]
\begin{center}
   \includegraphics[width=1.0\linewidth]{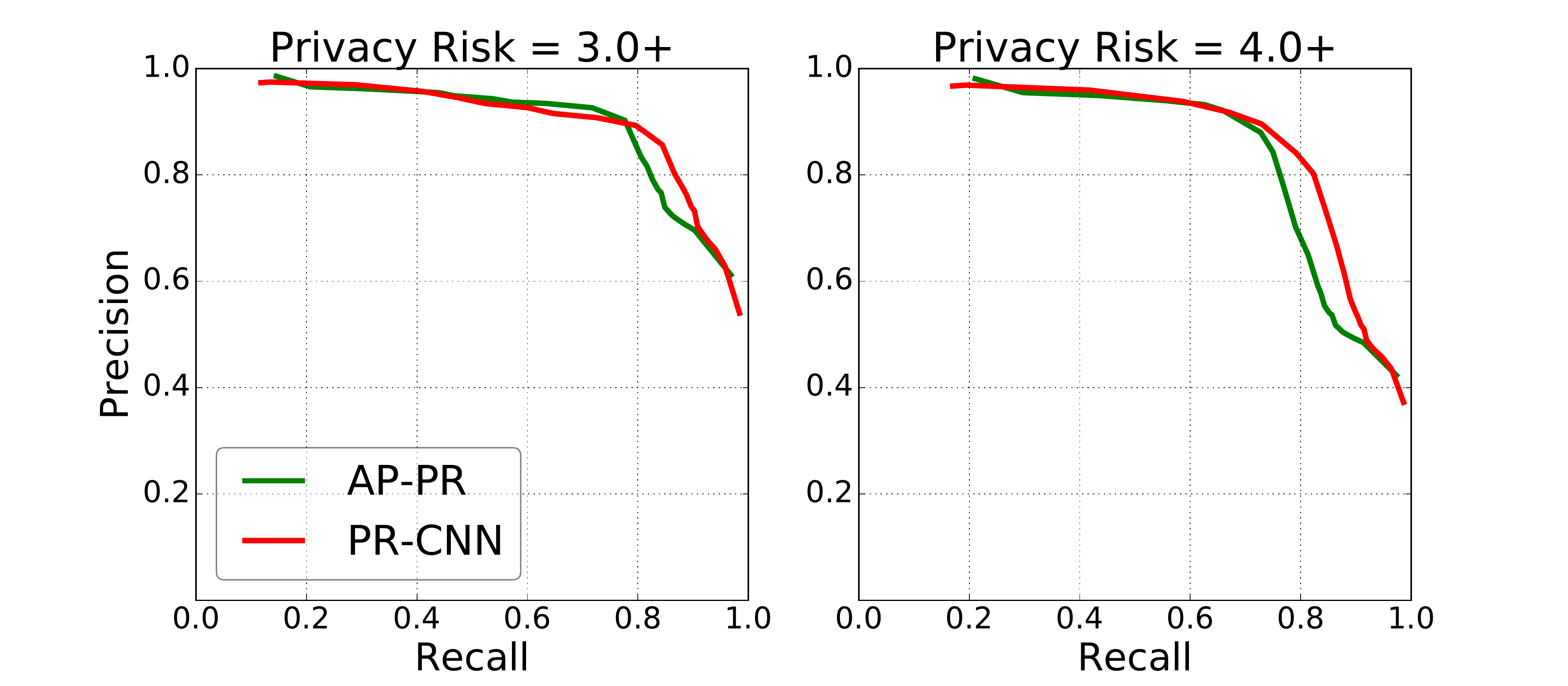}
\end{center}
   \caption{Performance of our approach in predicting Privacy Risks of images.
   Our approach performs better on high privacy-risk images.}
\label{fig:ppp_pr_graphs}
\end{figure}

\myparagraph{Evaluation}
We use two metrics for evaluation.
First, the $L1$ error averaged over all images and profiles; it represents the mean absolute difference between the ratings.
Secondly, we calculate the Precision-Recall curves for varying thresholds of sensitivity which indicates how well our models detect images above a certain true privacy risk.
By calculating the area under the Precision-Recall curves over all user profiles, we additionally report the Mean Average Precision (MAP).

In our experiments, we use the previously introduced user-profiles instead of individual users in order to cater to all the diverse privacy preferences equally that we have seen in the previous section.
We assign a privacy risk score of 0.5 for the \emph{safe} attribute for all profiles.

The evaluation of our approach on these metrics is presented in \autoref{tab:ppp_eval}.
Each graph in \autoref{fig:ppp_pr_graphs} represents PR curves over the ground-truth thresholded to obtain a particular risk interval, such that any score above this threshold is considered private.
This allows us to estimate performance of methods at various levels of sensitivity.
We then obtain the PR-curves for each sensitivity interval by thresholding scores estimated by AP-PR and PR-CNN.

From these results, we observe:
\begin{enumerate*}[label=(\roman*)]
	\item PR-CNN performs better in predicting risk compared to using the intermediate attributes predictions.
	Notably, the prediction is on average less than one step on the scale from 1 to 5 away from the true  privacy risk.
	\item Moreover, it is better at detecting high-risk images, as shown in \autoref{fig:ppp_pr_graphs}.
    In particular, we notice better recall for high-risk images.
    We discuss profile-specific PR curves in the supplementary material.
\end{enumerate*}

\subsection{Humans vs. Machine}
\label{sec:hvm}

\begin{figure}[t]
\begin{center}
   \includegraphics[width=1.0\linewidth]{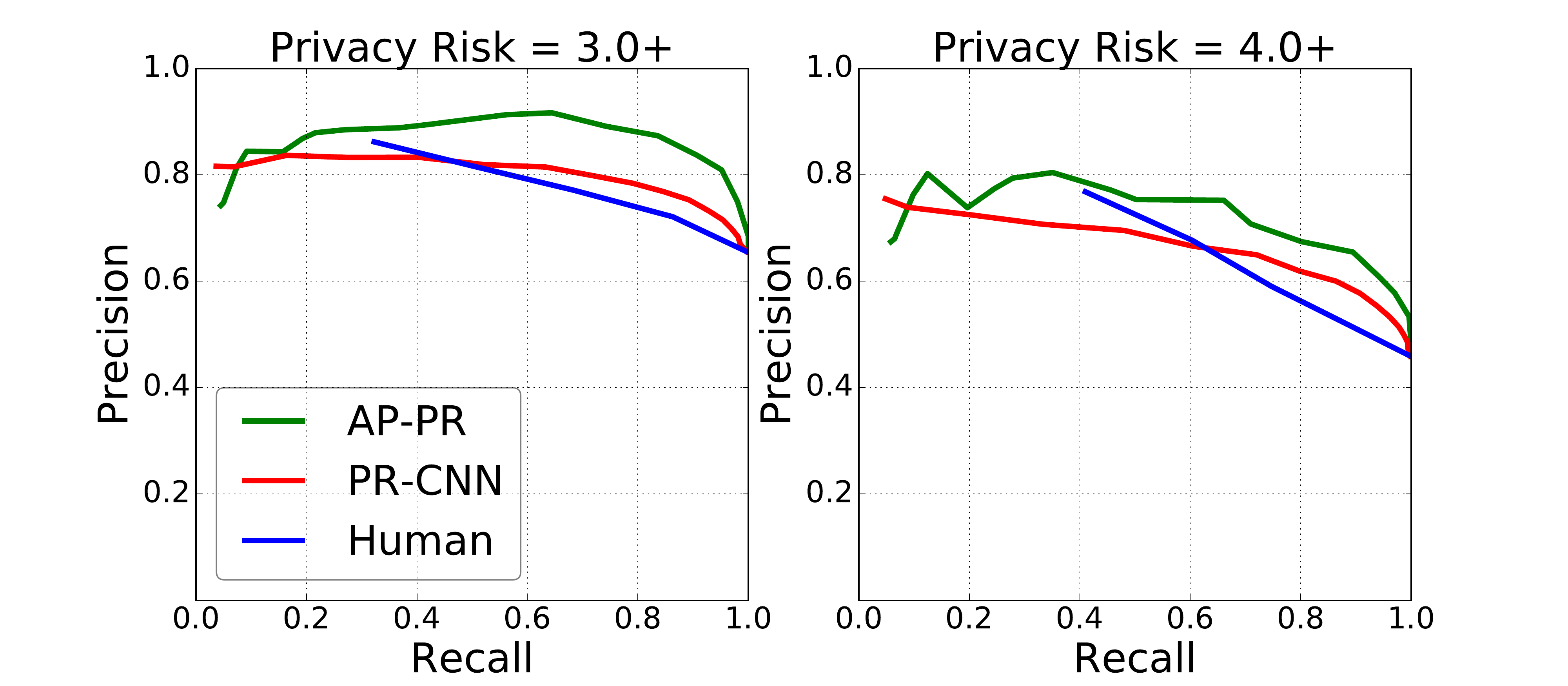}
\end{center}
   \caption{The Precision-Recall curves of three risk estimations are displayed -- users implicitly evaluating risk from images and our two methods AP-PR and PR-CNN.
   }
\label{fig:human_cnn_pr}
\end{figure}

In Section \ref{sec:papds_user_preferences}, we have shown inconsistency in users' privacy preferences and their assessment of privacy risks in images.
In this section, we compare our proposed approach for evaluating privacy risk against human judgments.

In our second user study (\autoref{sec:user_study_img}), for each attribute, users first assessed their personal privacy risk on images (providing a visual privacy risk score) and later rated their privacy preference (providing a desired privacy risk score).
We have computed scores with our privacy risk models AP-PR and PR-CNN on those very same images.

As a result, for each image, we have
\begin{enumerate*}[label=(\alph*)]
	\item users' privacy preference
    \item users' privacy risk judgment from images
    \item our AP-PR privacy risk score from images
    \item our PR-CNN privacy risk score from images.
\end{enumerate*}
All these scores are on a scale of 1 (Not Sensitive) to 5 (Extremely Sensitive).
Using the users desired preference as the ground-truth, we now ask: \emph{who is better at reproducing the user's desired privacy preference on images?}
As from the previous section, we use precision-recall and $L_1$-error as metrics to compare the desired preference score (a) and predicted privacy risk score for evaluation (b, c, d).

The precision-recall-curves for the three candidates are presented in \autoref{fig:human_cnn_pr}.
Evaluation using the $L_1$-error is discussed in the supplementary material.
We observe:
\begin{enumerate*}[label=(\roman*)]
	\item AP-PR achieves better precision-recall for the task than PR-CNN and -- remarkably -- is even {\it  consistently better than the users' image-based judgment}.
    \item On average, the PR-CNN estimates privacy risks  ($L_1$ error = 1.03) slightly better than the user's image-based judgment ($L_1$ error = 1.1) and AP-PR ($L_1$ error = 1.27).
\end{enumerate*}

\section{Conclusion}
We have extended the concept of privacy settings to visual content and have presented work towards a \emph{Visual Privacy Advisor} that can provide feedback to the users based on their privacy preferences.
The significance of this research direction is highlighted by our user study which shows users often fail to enforce their own privacy preferences when judging image content.
Our survey also captures typical privacy preference profiles that show a surprising level of diversity.
Our new VISPR dataset allowed us to train visual models that recognize privacy attributes, predict privacy risk scores and detect images that conflict with user's privacy.
In particular, a final comparison of human vs. machine prediction of privacy risks on images, shows an improvement by our model over human judgment.
This highlights the feasibility and future opportunities of the overarching goal -- a \emph{Visual Privacy Advisor}.

\section*{Acknowledgement}

\noindent
This research was supported by the German Research Foundation (DFG CRC 1223).
We thank Paarijaat Aditya, Philipp M\"uller and Julian Steil for advice on the user study.
We also thank Dr. Mykhaylo Andriluka and Seong Joon Oh for valuable feedback on the paper.

{\small
\bibliographystyle{ieee}
\bibliography{paper}

\begin{thebibliography}{10}\itemsep=-1pt

\bibitem{aditya2016pic}
P.~Aditya, R.~Sen, P.~Druschel, S.~Joon~Oh, R.~Benenson, M.~Fritz, B.~Schiele,
  B.~Bhattacharjee, and T.~T. Wu.
\newblock I-pic: A platform for privacy-compliant image capture.
\newblock In {\em Proceedings of the 14th Annual International Conference on
  Mobile Systems, Applications, and Services}, pages 235--248. ACM, 2016.

\bibitem{ahmed2015improved}
E.~Ahmed, M.~Jones, and T.~K. Marks.
\newblock An improved deep learning architecture for person re-identification.
\newblock In {\em Proceedings of the IEEE Conference on Computer Vision and
  Pattern Recognition}, pages 3908--3916, 2015.

\bibitem{aura2006scanning}
T.~Aura, T.~A. Kuhn, and M.~Roe.
\newblock Scanning electronic documents for personally identifiable
  information.
\newblock In {\em Proceedings of the 5th ACM workshop on Privacy in electronic
  society}, pages 41--50. ACM, 2006.

\bibitem{Bauckhage2010AgeRI}
C.~Bauckhage, A.~Jahanbekam, and C.~Thurau.
\newblock Age recognition in the wild.
\newblock In {\em ICPR}, 2010.

\bibitem{bier2014detection}
C.~Bier and J.~Prior.
\newblock Detection and labeling of personal identifiable information in
  e-mails.
\newblock In {\em IFIP International Information Security Conference}, pages
  351--358. Springer, 2014.

\bibitem{Chang2004AutomaticLP}
S.-L. Chang, L.-S. Chen, Y.-C. Chung, and S.-W. Chen.
\newblock Automatic license plate recognition.
\newblock {\em IEEE Trans. Intelligent Transportation Systems}, 5:42--53, 2004.

\bibitem{crammer2003family}
K.~Crammer and Y.~Singer.
\newblock A family of additive online algorithms for category ranking.
\newblock {\em Journal of Machine Learning Research}, 3(Feb):1025--1058, 2003.

\bibitem{danezis2009inferring}
G.~Danezis.
\newblock Inferring privacy policies for social networking services.
\newblock In {\em Proceedings of the 2nd ACM workshop on Security and
  artificial intelligence}, pages 5--10. ACM, 2009.

\bibitem{directive199595}
E.~Directive.
\newblock 95/46/ec of the european parliament and of the council of 24 october
  1995 on the protection of individuals with regard to the processing of
  personal data and on the free movement of such data.
\newblock {\em Official Journal of the EC}, 23(6), 1995.

\bibitem{dwyer2007trust}
C.~Dwyer, S.~Hiltz, and K.~Passerini.
\newblock Trust and privacy concern within social networking sites: A
  comparison of facebook and myspace.
\newblock {\em AMCIS 2007 proceedings}, page 339, 2007.

\bibitem{fang2010privacy}
L.~Fang and K.~LeFevre.
\newblock Privacy wizards for social networking sites.
\newblock In {\em Proceedings of the 19th international conference on World
  wide web}, pages 351--360. ACM, 2010.

\bibitem{gallagher_cvpr_08_clothing}
A.~Gallagher and T.~Chen.
\newblock Clothing cosegmentation for recognizing people.
\newblock In {\em Proc. CVPR}, 2008.

\bibitem{geng2008using}
L.~Geng, L.~Korba, X.~Wang, Y.~Wang, H.~Liu, and Y.~You.
\newblock Using data mining methods to predict personally identifiable
  information in emails.
\newblock In {\em International Conference on Advanced Data Mining and
  Applications}, pages 272--281. Springer, 2008.

\bibitem{gross2005information}
R.~Gross and A.~Acquisti.
\newblock Information revelation and privacy in online social networks.
\newblock In {\em Proceedings of the 2005 ACM workshop on Privacy in the
  electronic society}, pages 71--80. ACM, 2005.

\bibitem{guha2008noyb}
S.~Guha, K.~Tang, and P.~Francis.
\newblock Noyb: Privacy in online social networks.
\newblock In {\em Proceedings of the first workshop on Online social networks},
  pages 49--54. ACM, 2008.

\bibitem{harvey2012cv}
A.~Harvey.
\newblock Cv dazzle: Camouflage from computer vision.
\newblock {\em Technical report}, 2012.

\bibitem{he2016deep}
K.~He, X.~Zhang, S.~Ren, and J.~Sun.
\newblock Deep residual learning for image recognition.
\newblock In {\em Proceedings of the IEEE Conference on Computer Vision and
  Pattern Recognition}, pages 770--778, 2016.

\bibitem{Hoyle:2015:SLP:2702123.2702183}
R.~Hoyle, R.~Templeman, D.~Anthony, D.~Crandall, and A.~Kapadia.
\newblock Sensitive lifelogs: A privacy analysis of photos from wearable
  cameras.
\newblock In {\em Proceedings of the 33rd Annual ACM Conference on Human
  Factors in Computing Systems}, CHI '15, pages 1645--1648, New York, NY, USA,
  2015. ACM.

\bibitem{jia2014caffe}
Y.~Jia, E.~Shelhamer, J.~Donahue, S.~Karayev, J.~Long, R.~Girshick,
  S.~Guadarrama, and T.~Darrell.
\newblock Caffe: Convolutional architecture for fast feature embedding.
\newblock In {\em Proceedings of the 22nd ACM international conference on
  Multimedia}, pages 675--678. ACM, 2014.

\bibitem{screenavoider2016chi}
M.~Korayem, R.~Templeman, D.~Chen, D.~Crandall, and A.~Kapadia.
\newblock Enhancing lifelogging privacy by detecting screens.
\newblock In {\em ACM CHI Conference on Human Factors in Computing Systems
  (CHI)}, 2016.

\bibitem{openimages}
I.~Krasin, T.~Duerig, N.~Alldrin, V.~Ferrari, S.~Abu-El-Haija, A.~Kuznetsova,
  H.~Rom, J.~Uijlings, S.~Popov, A.~Veit, S.~Belongie, V.~Gomes, A.~Gupta,
  C.~Sun, G.~Chechik, D.~Cai, Z.~Feng, D.~Narayanan, and K.~Murphy.
\newblock Openimages: A public dataset for large-scale multi-label and
  multi-class image classification.
\newblock {\em Dataset available from https://github.com/openimages}, 2017.

\bibitem{Krishnamurthy2011PrivacyLV}
B.~Krishnamurthy, K.~Naryshkin, and C.~E. Wills.
\newblock Privacy leakage vs. protection measures: the growing disconnect.
\newblock In {\em Proceedings of the Web}, volume~2, pages 1--10, 2011.

\bibitem{krishnamurthy2008characterizing}
B.~Krishnamurthy and C.~E. Wills.
\newblock Characterizing privacy in online social networks.
\newblock In {\em Proceedings of the first workshop on Online social networks},
  pages 37--42. ACM, 2008.

\bibitem{krishnamurthy2009leakage}
B.~Krishnamurthy and C.~E. Wills.
\newblock On the leakage of personally identifiable information via online
  social networks.
\newblock In {\em Proceedings of the 2nd ACM workshop on Online social
  networks}, pages 7--12. ACM, 2009.

\bibitem{lapin2016loss}
M.~Lapin, M.~Hein, and B.~Schiele.
\newblock Loss functions for top-k error: Analysis and insights.
\newblock In {\em CVPR}, 2016.

\bibitem{lenhart2007teens}
A.~Lenhart and M.~Madden.
\newblock Teens, privacy and online social networks: How teens manage their
  online identities and personal information in the age of myspace.
\newblock 2007.

\bibitem{li2011findu}
M.~Li, N.~Cao, S.~Yu, and W.~Lou.
\newblock Findu: Privacy-preserving personal profile matching in mobile social
  networks.
\newblock In {\em INFOCOM, 2011 Proceedings IEEE}, pages 2435--2443. IEEE,
  2011.

\bibitem{liu2011analyzing}
Y.~Liu, K.~P. Gummadi, B.~Krishnamurthy, and A.~Mislove.
\newblock Analyzing facebook privacy settings: user expectations vs. reality.
\newblock In {\em Proceedings of the 2011 ACM SIGCOMM conference on Internet
  measurement conference}, pages 61--70. ACM, 2011.

\bibitem{mccallister2010guide}
E.~McCallister.
\newblock {\em Guide to protecting the confidentiality of personally
  identifiable information}.
\newblock Diane Publishing, 2010.

\bibitem{mclaughlin2016recurrent}
N.~McLaughlin, J.~Martinez~del Rincon, and P.~Miller.
\newblock Recurrent convolutional network for video-based person
  re-identification.
\newblock In {\em Proceedings of the IEEE Conference on Computer Vision and
  Pattern Recognition}, pages 1325--1334, 2016.

\bibitem{moosavi2016universal}
S.-M. Moosavi-Dezfooli, A.~Fawzi, O.~Fawzi, and P.~Frossard.
\newblock Universal adversarial perturbations.
\newblock {\em arXiv preprint arXiv:1610.08401}, 2016.

\bibitem{Narayanan2009DeanonymizingSN}
A.~Narayanan and V.~Shmatikov.
\newblock De-anonymizing social networks.
\newblock {\em CoRR}, abs/0903.3276, 2009.

\bibitem{neustaedter2006blur}
C.~Neustaedter, S.~Greenberg, and M.~Boyle.
\newblock Blur filtration fails to preserve privacy for home-based video
  conferencing.
\newblock {\em ACM Transactions on Computer-Human Interaction (TOCHI)},
  13(1):1--36, 2006.

\bibitem{oh2016faceless}
S.~J. Oh, R.~Benenson, M.~Fritz, and B.~Schiele.
\newblock Faceless person recognition: Privacy implications in social media.
\newblock In {\em European Conference on Computer Vision}, pages 19--35.
  Springer, 2016.

\bibitem{papernot2016distillation}
N.~Papernot, P.~McDaniel, X.~Wu, S.~Jha, and A.~Swami.
\newblock Distillation as a defense to adversarial perturbations against deep
  neural networks.
\newblock In {\em Security and Privacy (SP), 2016 IEEE Symposium on}, pages
  582--597. IEEE, 2016.

\bibitem{pittaluga2015privacy}
F.~Pittaluga and S.~J. Koppal.
\newblock Privacy preserving optics for miniature vision sensors.
\newblock In {\em Proceedings of the IEEE Conference on Computer Vision and
  Pattern Recognition}, pages 314--324, 2015.

\bibitem{raval2014markit}
N.~Raval, A.~Srivastava, K.~Lebeck, L.~Cox, and A.~Machanavajjhala.
\newblock Markit: Privacy markers for protecting visual secrets.
\newblock In {\em Proceedings of the 2014 ACM International Joint Conference on
  Pervasive and Ubiquitous Computing: Adjunct Publication}, pages 1289--1295.
  ACM, 2014.

\bibitem{rousseeuw1987silhouettes}
P.~J. Rousseeuw.
\newblock Silhouettes: a graphical aid to the interpretation and validation of
  cluster analysis.
\newblock {\em Journal of computational and applied mathematics}, 20:53--65,
  1987.

\bibitem{shao2013you}
M.~Shao, L.~Li, and Y.~Fu.
\newblock What do you do? occupation recognition in a photo via social context.
\newblock In {\em Proceedings of the IEEE International Conference on Computer
  Vision}, pages 3631--3638, 2013.

\bibitem{Sun2017FaceDU}
X.~Sun, P.~Wu, and S.~C.~H. Hoi.
\newblock Face detection using deep learning: An improved faster rcnn approach.
\newblock {\em CoRR}, abs/1701.08289, 2017.

\bibitem{szegedy2015going}
C.~Szegedy, W.~Liu, Y.~Jia, P.~Sermanet, S.~Reed, D.~Anguelov, D.~Erhan,
  V.~Vanhoucke, and A.~Rabinovich.
\newblock Going deeper with convolutions.
\newblock In {\em Proceedings of the IEEE Conference on Computer Vision and
  Pattern Recognition}, pages 1--9, 2015.

\bibitem{tonge2015privacy}
A.~Tonge and C.~Caragea.
\newblock Privacy prediction of images shared on social media sites using deep
  features.
\newblock {\em arXiv preprint arXiv:1510.08583}, 2015.

\bibitem{veiga2016privacy}
M.~H. Veiga and C.~Eickhoff.
\newblock Privacy leakage through innocent content sharing in online social
  networks.
\newblock {\em arXiv preprint arXiv:1607.02714}, 2016.

\bibitem{Viola2001RobustRF}
P.~A. Viola and M.~J. Jones.
\newblock Robust real-time face detection.
\newblock {\em International Journal of Computer Vision}, 57:137--154, 2001.

\bibitem{Wang2010SeeingPI}
G.~Wang, A.~C. Gallagher, J.~Luo, and D.~A. Forsyth.
\newblock Seeing people in social context: Recognizing people and social
  relationships.
\newblock In {\em ECCV}, 2010.

\bibitem{wilber2016can}
M.~J. Wilber, V.~Shmatikov, and S.~Belongie.
\newblock Can we still avoid automatic face detection?
\newblock In {\em Applications of Computer Vision (WACV), 2016 IEEE Winter
  Conference on}, pages 1--9. IEEE, 2016.

\bibitem{Yang2013AppIntentAS}
Z.~Yang, M.~Yang, Y.~Zhang, G.~Gu, P.~Ning, and X.~S. Wang.
\newblock Appintent: analyzing sensitive data transmission in android for
  privacy leakage detection.
\newblock In {\em ACM Conference on Computer and Communications Security},
  2013.

\bibitem{young2009information}
A.~L. Young and A.~Quan-Haase.
\newblock Information revelation and internet privacy concerns on social
  network sites: a case study of facebook.
\newblock In {\em Proceedings of the fourth international conference on
  Communities and technologies}, pages 265--274. ACM, 2009.

\bibitem{Zhang2006LearningBasedLP}
H.~Zhang, W.~Jia, X.~He, and Q.~Wu.
\newblock Learning-based license plate detection using global and local
  features.
\newblock In {\em ICPR}, 2006.

\bibitem{Zhang_2015_CVPR}
N.~Zhang, M.~Paluri, Y.~Taigman, R.~Fergus, and L.~Bourdev.
\newblock Beyond frontal faces: Improving person recognition using multiple
  cues.
\newblock In {\em The IEEE Conference on Computer Vision and Pattern
  Recognition (CVPR)}, June 2015.

\bibitem{Zheng2009TourTW}
Y.~Zheng, M.~Zhao, Y.~Song, H.~Adam, U.~Buddemeier, A.~Bissacco, F.~Brucher,
  T.-S. Chua, and H.~Neven.
\newblock Tour the world: Building a web-scale landmark recognition engine.
\newblock In {\em 2009 IEEE Conference on Computer Vision and Pattern
  Recognition}, 2009.

\bibitem{zhou2008preserving}
B.~Zhou and J.~Pei.
\newblock Preserving privacy in social networks against neighborhood attacks.
\newblock In {\em Data Engineering, 2008. ICDE 2008. IEEE 24th International
  Conference on}, pages 506--515. IEEE, 2008.

\bibitem{Zhou2012PrincipalVW}
W.~Zhou, H.~Li, Y.~Lu, and Q.~Tian.
\newblock Principal visual word discovery for automatic license plate
  detection.
\newblock {\em IEEE Trans. Image Processing}, 21:4269--4279, 2012.

\end{thebibliography}
}

\section*{Changelog}
\paragraph*{Version 2 (31-July-2017)}
\begin{itemize}[noitemsep,topsep=0pt,parsep=0pt,partopsep=0pt]
  \item New teaser figure
  \item Improving writing
  \item Citing more related work
  \item Additional information on User Study
  \item Project web-page link
\end{itemize}


\newpage
\appendix
\appendixpage

\section{Privacy Attributes and Examples}

A complete list of privacy attributes with descriptions and an example image is given in  \autoref{tab:definitions}.
We consider all these cases when viewing the image in its original high-resolution form.
We use these definitions to any subject in the image -- either in the foreground or background.
Using these definitions, attributes can be typically inferred from an image in multiple ways: 
\begin{enumerate*}[label=(\alph*)]
	\item \emph{Direct}: it is explicitly mentioned, such as in a form or document (\eg gender on an identity card)
    \item \emph{Visual}: based on visual cues (\eg gender from clothing or facial features)
    \item \emph{Reasoning}: it is inferred by some additional reasoning (\eg relationships based on age differences between multiple people).
\end{enumerate*} 
Dataset is available on the project website: \url{https://tribhuvanesh.github.io/vpa/}.

\begin{table*}
\centering
\begin{tabular}{p{0.1\textwidth}p{0.2\textwidth}p{0.45\textwidth}m{0.12\textwidth}}
\toprule
Group & Attribute & Description & Examples \\ \midrule
Personal Description        
& Gender         
& Subject's gender is clearly visible using one or more gender-specific discriminative visual cues such as more than 50\% body being visible, clothing, facial/head hair or colored nails.
& \includegraphics[width=0.1\textwidth]{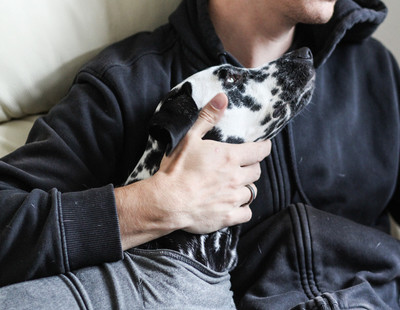}        
\\

& Eye Color         
& If eyes are visible and can be categorized as one of: brown, hazel, blue or green.           
& \includegraphics[width=0.1\textwidth]{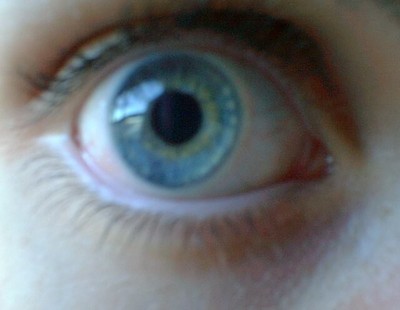}    
\\

& Hair Color         
& Subject's head hair color is visible
& \includegraphics[width=0.1\textwidth]{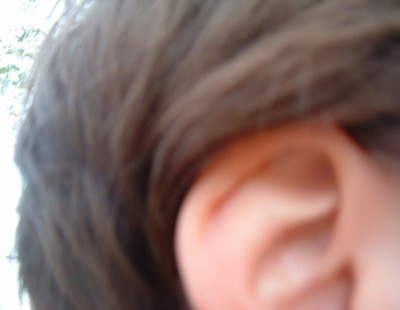}     
\\

& Fingerprint         
& Fingerprint is visible through either a close-up shot of one's finger or adu -sh .
n imprint on some surface.
& \includegraphics[width=0.1\textwidth]{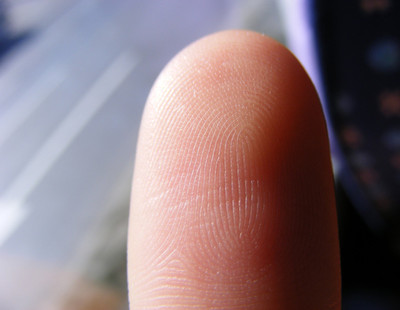}    
\\

& Signature         
& Complete signature is visible in an image, such as in a form or document
& \includegraphics[width=0.1\textwidth]{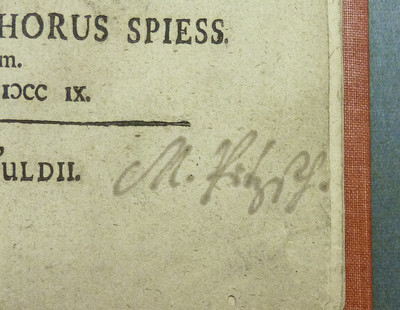}   
\\

& Face (Complete)         
& A face is completely visible. Also includes photographs of faces on identity cards, documents or billboards.
& \includegraphics[width=0.1\textwidth]{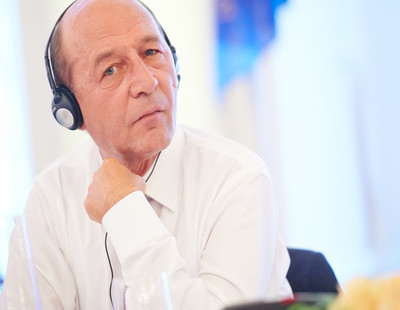}  
\\

& Face (Partial)         
& Less than 70\% of the face is visible or there is occlusion, such as when the subject is wearing sun-glasses.
& \includegraphics[width=0.1\textwidth]{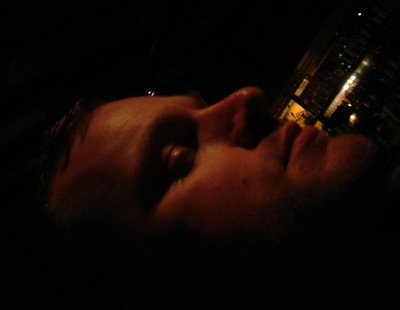}
\\

& Tattoo         
& Subject displays either a tattoo or body paint.           
& \includegraphics[width=0.1\textwidth]{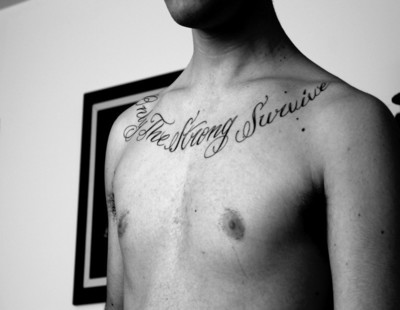}  
\\

& Nudity (Partial)         
& Subject appears in undergarments           
& \includegraphics[width=0.1\textwidth]{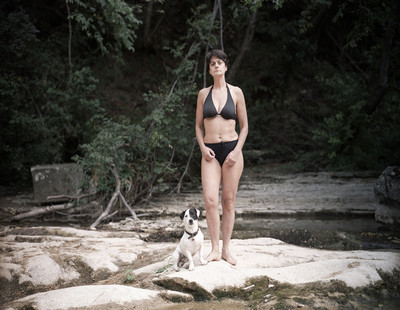}
\\
& Nudity (Complete)         
& Human subject appears without clothing
& \includegraphics[width=0.1\textwidth]{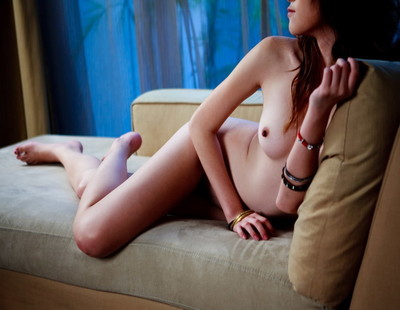}
\\

& Race         
& Any subject in the photograph can be categorized into one of Caucasian, Asian or Negroid.       
& \includegraphics[width=0.1\textwidth]{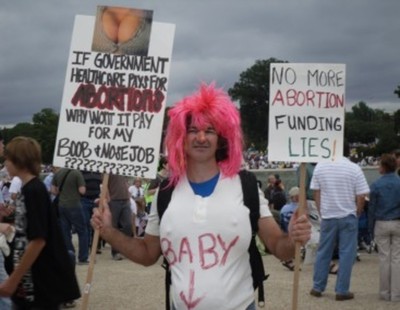}
\\

& (Skin) Color         
& One's skin color can be categorized into one of White, Brown or Black.           
& \includegraphics[width=0.1\textwidth]{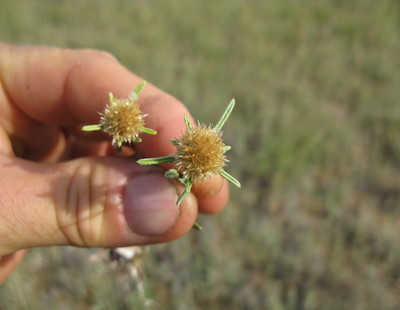}
\\

& Traditional Clothing         
& Subject appears in clothing which is indicative of a particular region or country \eg dirndl, sari.
& \includegraphics[width=0.1\textwidth]{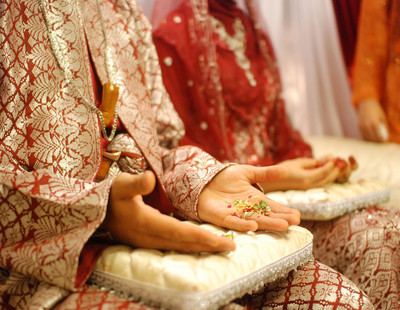}
\\ \bottomrule
\end{tabular}
\caption{List of Privacy Attributes including their definitions and examples}
\label{tab:definitions}
\end{table*}

\begin{table*}
\centering
\begin{tabular}{p{0.1\textwidth}p{0.2\textwidth}p{0.45\textwidth}m{0.12\textwidth}}
\toprule
Group & Attribute & Description & Examples \\ \midrule

& Full Name        
& A recognizable full name which appears in the context of a form, document or a badge. Also includes if the name can be inferred from a signature.
& \includegraphics[width=0.1\textwidth]{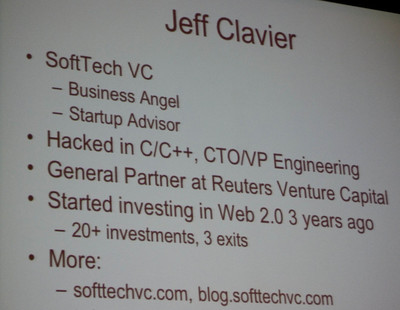}
\\

& Name (First)         
& Only if the first name is visible on a form, document, badge or clothing.          
& \includegraphics[width=0.1\textwidth]{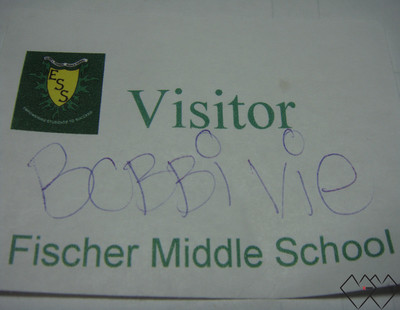}
\\

& Name (Last)          
& Only if the last name is visible on a form, document, badge or clothing.    
& \includegraphics[width=0.1\textwidth]{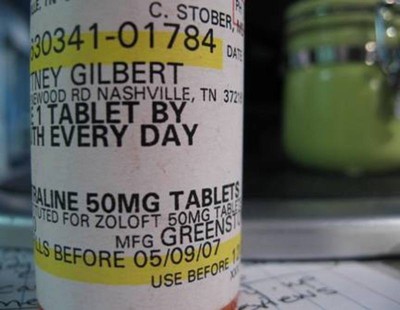}
\\

& Place of Birth          
& Place of Birth is explicitly mentioned, such as in a form or in an identification document.
& \includegraphics[width=0.1\textwidth]{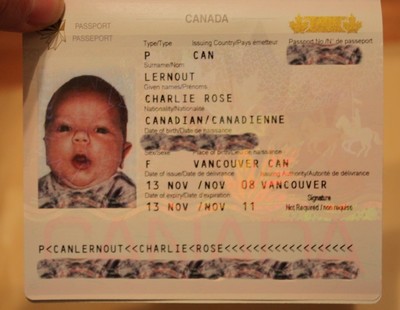}
\\

& Date of Birth          
& Date of Birth is explicitly mentioned in writing. Includes year, month or the day of birth.
& \includegraphics[width=0.1\textwidth]{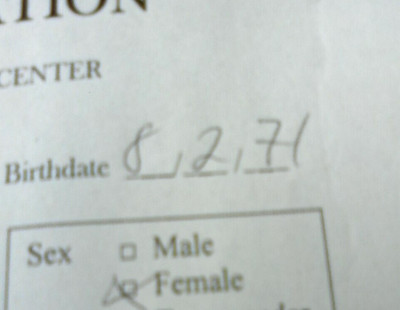}
\\

& Nationality          
& A passport indicating country is clearly visible. Includes the case if a subject appears holding a country's flag or wearing a uniform bearing the flag (such as a soldier or an international athlete).
& \includegraphics[width=0.1\textwidth]{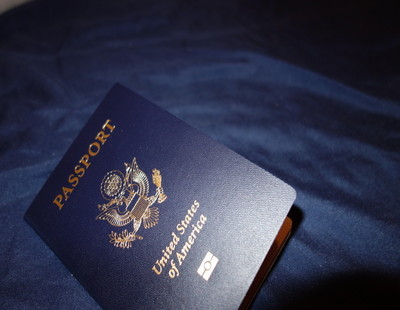}
\\

& Handwriting          
& Hand-written text on any surface.
& \includegraphics[width=0.1\textwidth]{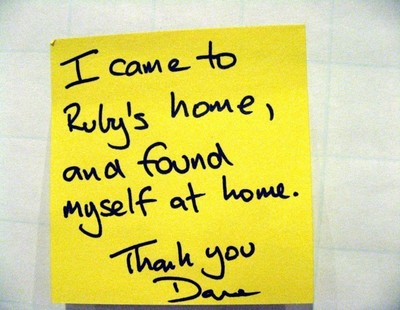}
\\

& Marital status          
& A subject is wearing an engagement ring. Includes wedding photographs taken of the bride and groom.
& \includegraphics[width=0.1\textwidth]{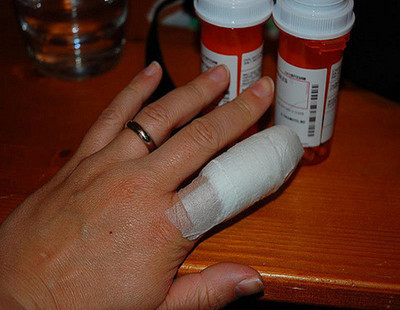}
\\ \midrule
Documents
& National Identification          
& Documents such as a Green Card or a European national identity card, not including passports.
& \includegraphics[width=0.1\textwidth]{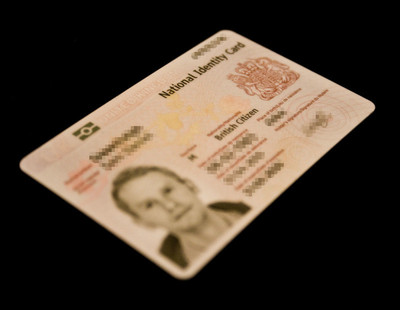}
\\

& Credit Card          
& Either the front or back of a credit card. Includes cases when the card is partially visible \eg in someone's hand or in a shredded form
& \includegraphics[width=0.1\textwidth]{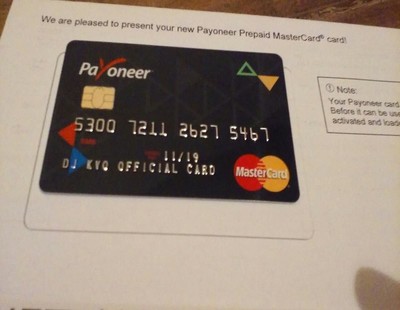}
\\

& Passport          
& A photograph of any page in the passport or its front cover.
& \includegraphics[width=0.1\textwidth]{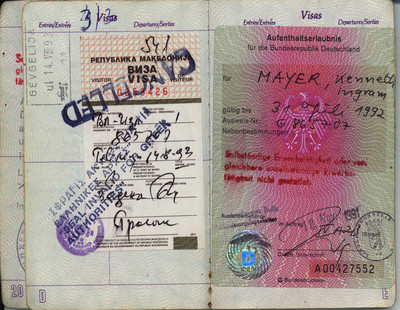}
\\

& Drivers License          
& Either front or back of a drivers license or a driving permit.
& \includegraphics[width=0.1\textwidth]{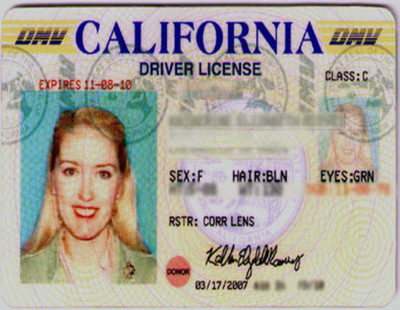}
\\

& Student ID          
& Front or back of a student identity card, with at least the name of a school, college or university clearly readable.
& \includegraphics[width=0.1\textwidth]{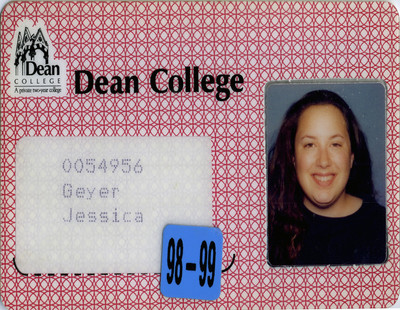}
\\ \bottomrule
\end{tabular}
\end{table*}

\begin{table*}
\centering
\begin{tabular}{p{0.1\textwidth}p{0.2\textwidth}p{0.45\textwidth}m{0.12\textwidth}}
\toprule
Group & Attribute & Description & Examples \\ \midrule

& Mail          
& Contents of a mail or the envelope.
& \includegraphics[width=0.1\textwidth]{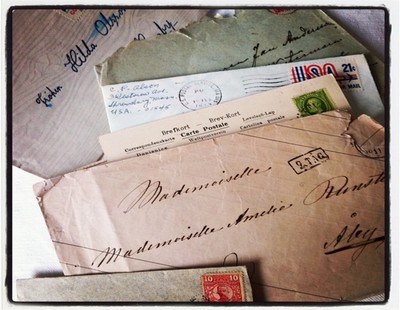}
\\

& Receipts          
& Purchase receipts indicating a financial transaction with an amount clearly visible, \eg a restaurant receipt.
& \includegraphics[width=0.1\textwidth]{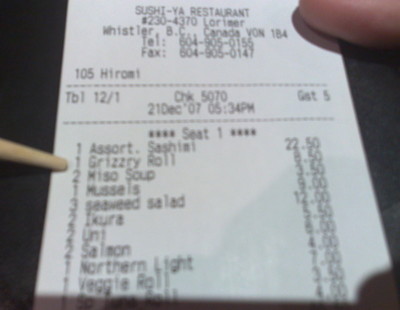}
\\

& Tickets          
& A travel, movie or concert ticket which specifies travel location or an event.
& \includegraphics[width=0.1\textwidth]{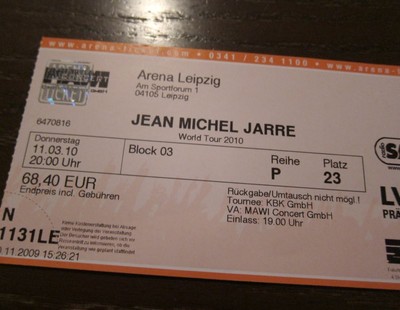}
\\ \midrule
Health
& Physical disability          
& Subject appears with a permanent physical disability \eg an amputee or a person in a wheelchair.
& \includegraphics[width=0.1\textwidth]{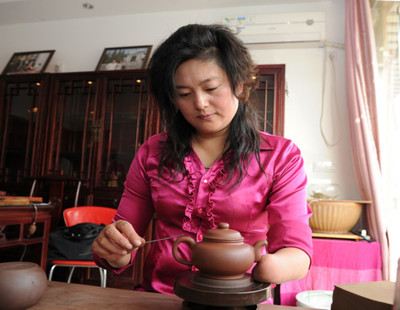}
\\

& Medical Treatment          
& Subject appears either with an injury or indicates hospital admittance.
& \includegraphics[width=0.1\textwidth]{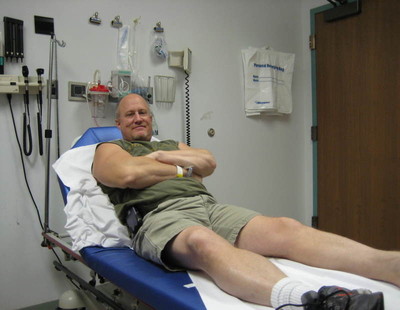}
\\

& Medical History          
& Photographs of medicine or medical prescriptions.
& \includegraphics[width=0.1\textwidth]{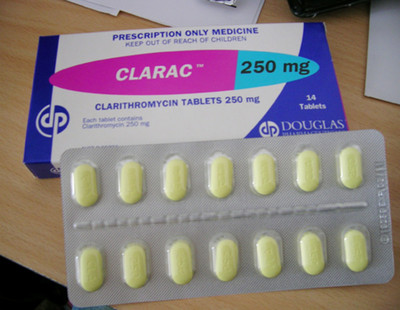}
\\ \midrule
Employment
& Occupation          
& Subject appears in a distinguishable occupation-specific uniform \eg doctor, policemen, construction worker.
& \includegraphics[width=0.1\textwidth]{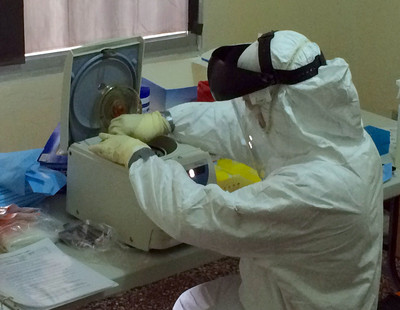}
\\

& Work Occasion          
& Subject is photographed while giving a talk, presentation, attending a work-related or broad-casting event. Includes photographs of people in formal attire in an office.
& \includegraphics[width=0.1\textwidth]{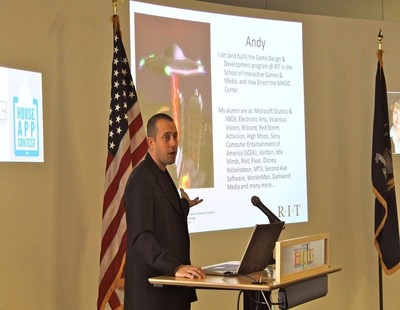}
\\ \midrule
Personal Life
& Religion          
& Subject appears associated with a distinguishable religious symbol, religion-specific clothing or at a religious location.
& \includegraphics[width=0.1\textwidth]{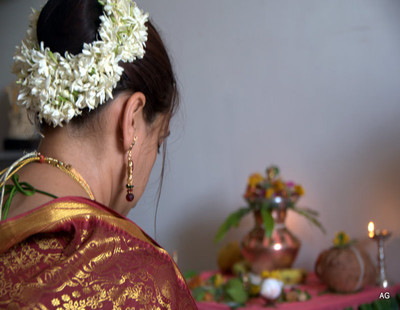}
\\ 

& Sexual Orientation          
& Two subjects are photographed in an intimate setting
& \includegraphics[width=0.1\textwidth]{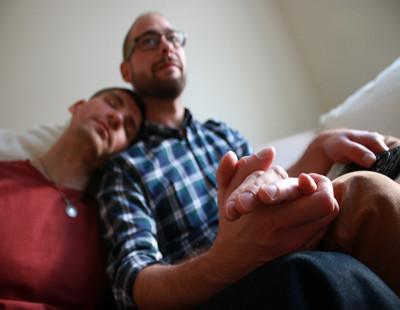}
\\

& Culture          
& Subjects appear celebrating a traditional festival or attending an art or culture related activity \eg concert, play.
& \includegraphics[width=0.1\textwidth]{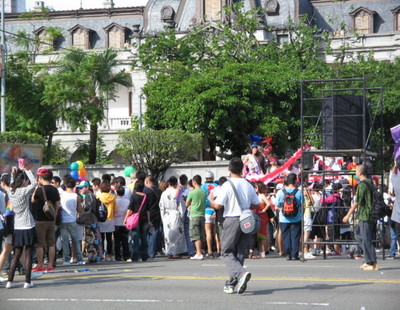}
\\

& Hobbies          
& A non-professional related activity of a subject is visible \eg playing a musical instrument, taking photographs.
& \includegraphics[width=0.1\textwidth]{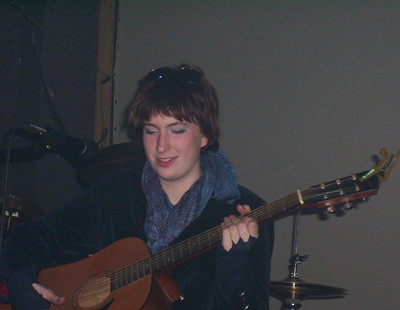}
\\

& Sports          
& Subject appears taking part in an indoor or outdoor sports activity
& \includegraphics[width=0.1\textwidth]{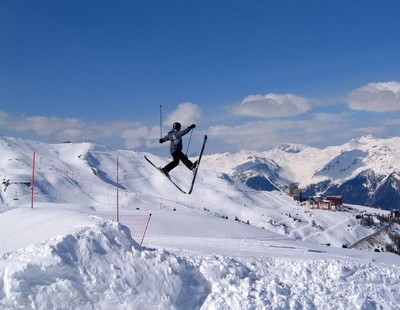}
\\ \bottomrule
\end{tabular}
\end{table*}

\begin{table*}
\centering
\begin{tabular}{p{0.1\textwidth}p{0.2\textwidth}p{0.45\textwidth}m{0.12\textwidth}}
\toprule
Group & Attribute & Description & Examples \\ \midrule

& Education history          
& Photographs contains cues indicating subject's education history, such as at a graduation ceremony, clothing indicating university or an academic or school certificate
& \includegraphics[width=0.1\textwidth]{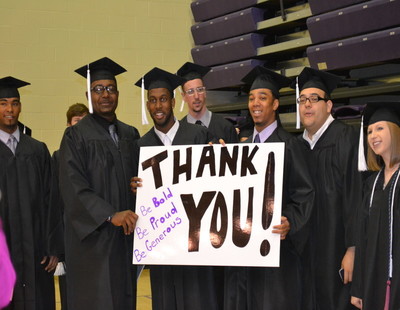}
\\ 

& Legal involvement          
& Photographs indicating subject's involvement with law-related activities \eg someone being arrested, in a court hearing.
& \includegraphics[width=0.1\textwidth]{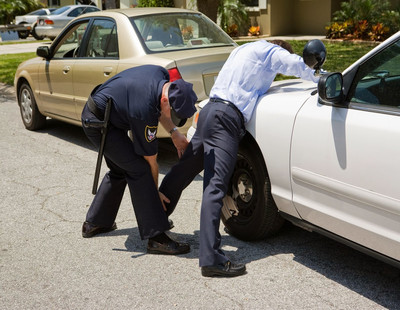}
\\

& Personal Occasion          
& Photographs of people celebrating a personal occasion with friends or family members \eg wedding, birthday.
& \includegraphics[width=0.1\textwidth]{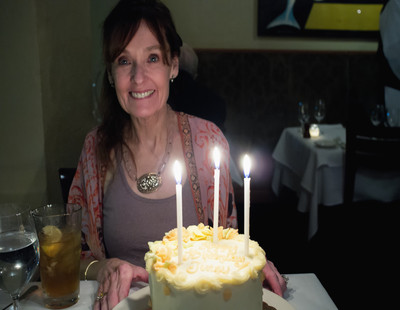}
\\

& General Opinion          
& Subject appears associated with a placard or clothing indicating opinion on general issues \eg wars, taxes, LGBT rights.
& \includegraphics[width=0.1\textwidth]{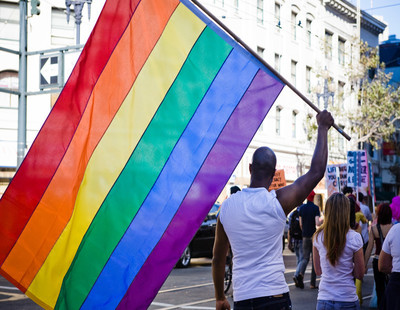}
\\

& Political Opinion          
& Subject appears with either clothing, placard or in a crowd at a political rally.
& \includegraphics[width=0.1\textwidth]{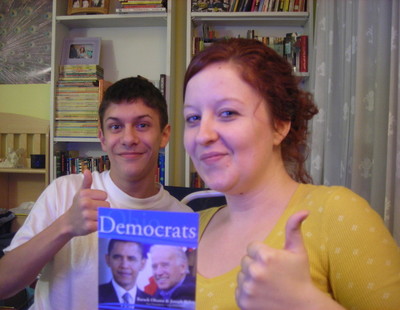}
\\ \midrule
Relationships
& Personal Relationships          
& Photographs of people in a visually-identifiable personal relationship \eg mother-son, husband-wife.
& \includegraphics[width=0.1\textwidth]{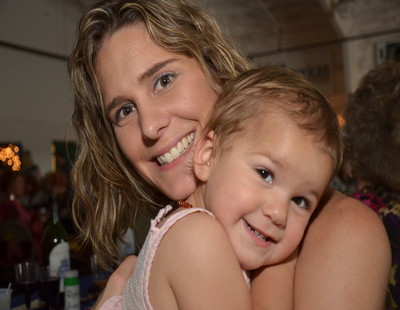}
\\

& Social Circle          
& Subjects of the same age-group photographed in a casual setting \eg friends at a party, walking together on a street.
& \includegraphics[width=0.1\textwidth]{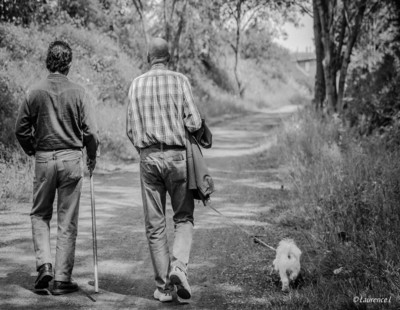}
\\

& Professional Circle        
& A group of people who share an occupation (\eg a group of policemen) or who are dressed for a professional event (\eg a conference or meeting).
& \includegraphics[width=0.1\textwidth]{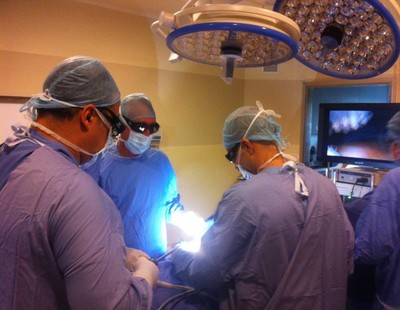}
\\

& Competitors          
& A group of people taking part in team sports. Also includes the case when subjects belong to the same team.
& \includegraphics[width=0.1\textwidth]{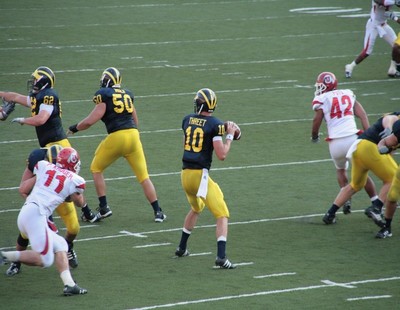}
\\

& Spectators          
& A group of people spectating an event such as a concert or play.
& \includegraphics[width=0.1\textwidth]{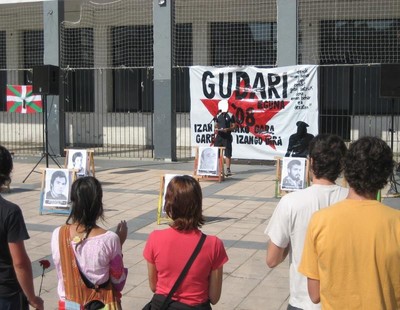}
\\

& Similar view          
& A group of people at a rally or a protest who share opinions on a general issue. Only includes the case when placards or clothing denoting a cause or rallying for a political party is visible.
& \includegraphics[width=0.1\textwidth]{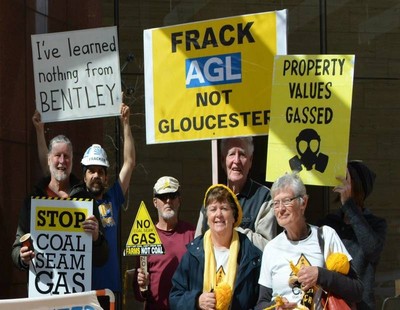}
\\ \midrule
Whereabouts
& Visited Landmark          
& Photograph contains text indicating a business' name, street sign or a well-known landmark.
& \includegraphics[width=0.1\textwidth]{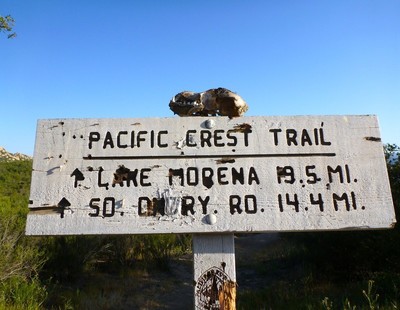}
\\ \bottomrule
\end{tabular}
\end{table*}

\begin{table*}
\centering
\begin{tabular}{p{0.1\textwidth}p{0.2\textwidth}p{0.45\textwidth}m{0.12\textwidth}}
\toprule
Group & Attribute & Description & Examples \\ \midrule

& Visited Location (Complete)          
& Text indicating a \emph{complete} address (\eg restaurant receipt with the address of the restaurant) or a screen-shot of GPS-based location.
& \includegraphics[width=0.1\textwidth]{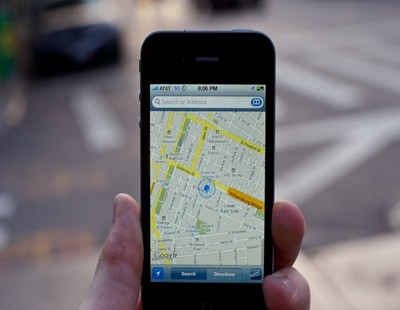}
\\

& Visited Location (Partial)          
& Text which partially indicates the subject's location, such as street name, city or country where the photograph was taken.
& \includegraphics[width=0.1\textwidth]{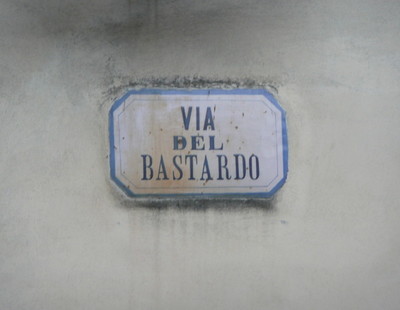}
\\

& Home address (Complete)          
& Photograph containing a complete non-commercial postal address.
& \includegraphics[width=0.1\textwidth]{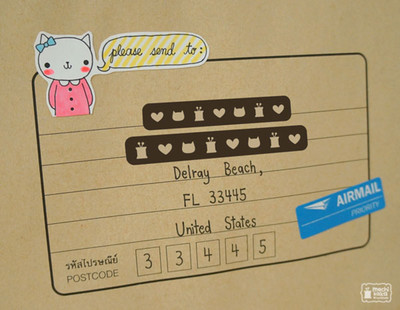}
\\

& Home address (Partial)          
& Photograph containing a partial non-commercial postal address.
& \includegraphics[width=0.1\textwidth]{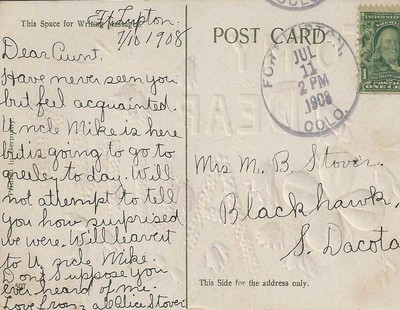}
\\

& Date/Time of Activity          
& Photograph contains information of date and/or time of subject's location or activity such as a time-stamp watermark in an image, or a clock in the photograph.
& \includegraphics[width=0.1\textwidth]{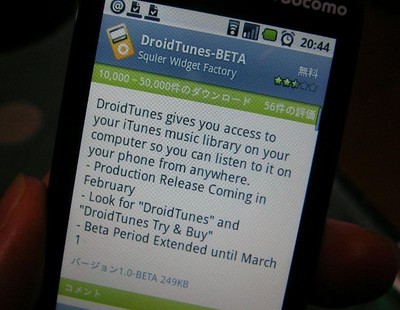}
\\

& Phone no.          
& A phone number that is visible in the photograph (either personal or commercial).
& \includegraphics[width=0.1\textwidth]{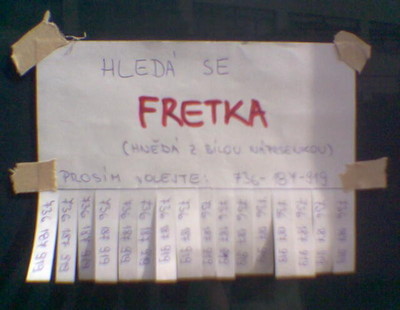}
\\ \midrule
Internet Activity
& Username          
& A screen shot of a website which mentions any username or internet handles.
& \includegraphics[width=0.1\textwidth]{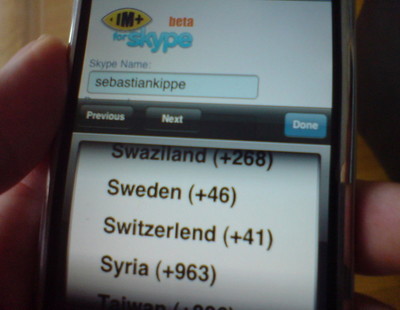}
\\

& Email address          
& Any complete valid email-address that appears in a photograph or a screen-shot.
& \includegraphics[width=0.1\textwidth]{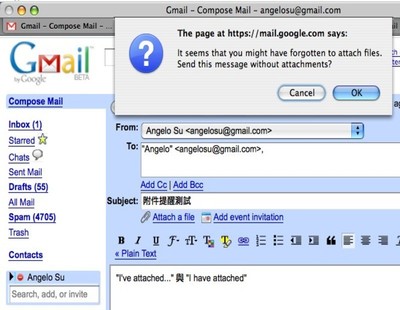}
\\

& Email content          
& Screenshots of emails including the subject of the email, or parts of the email body content.
& \includegraphics[width=0.1\textwidth]{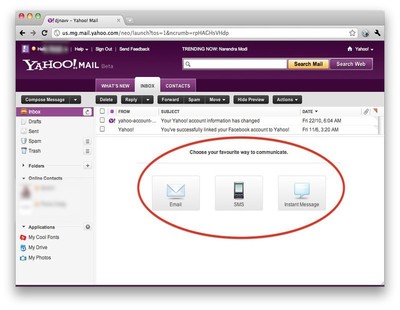}
\\

& Online conversations          
& Screenshots of online conversations, posts, tweets or internet activity by any user.
& \includegraphics[width=0.1\textwidth]{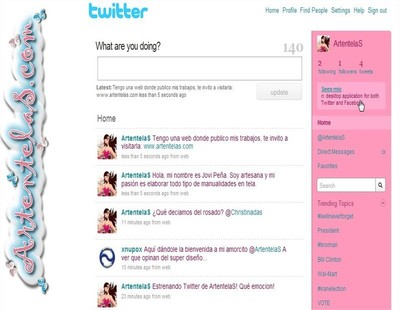}
\\ \midrule
Automobile
& Vehicle Ownership          
& Photograph of a person riding a motor vehicle.
& \includegraphics[width=0.1\textwidth]{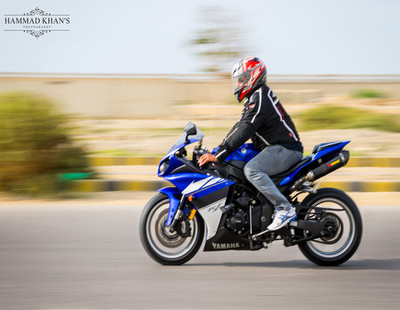}
\\

& License Plate (Complete)          
& A  clearly visible license plate or registration number of any motor vehicle.
& \includegraphics[width=0.1\textwidth]{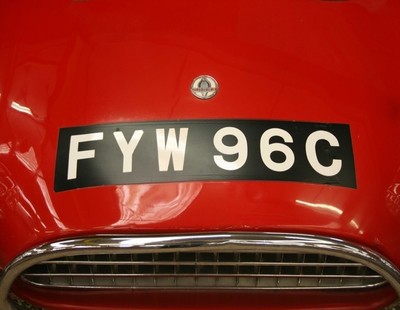}
\\

& License Plate (Partial)          
& A partial license plate or registration number of any motor vehicle
& \includegraphics[width=0.1\textwidth]{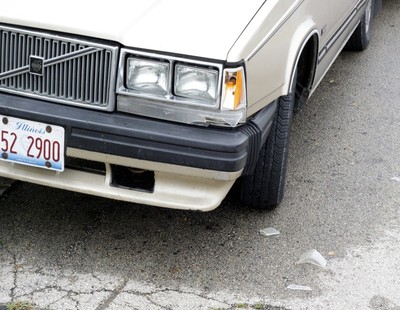}
\\
\bottomrule
\end{tabular}
\end{table*}


\section{Additional Details on User Study}
\label{sec:appendix_user_study}

In this section, we provide additional details on the user study discussed in Section 4.

\subsection{Understanding Users' Privacy Preferences}

\begin{figure}
\begin{center}
   \includegraphics[width=\linewidth]{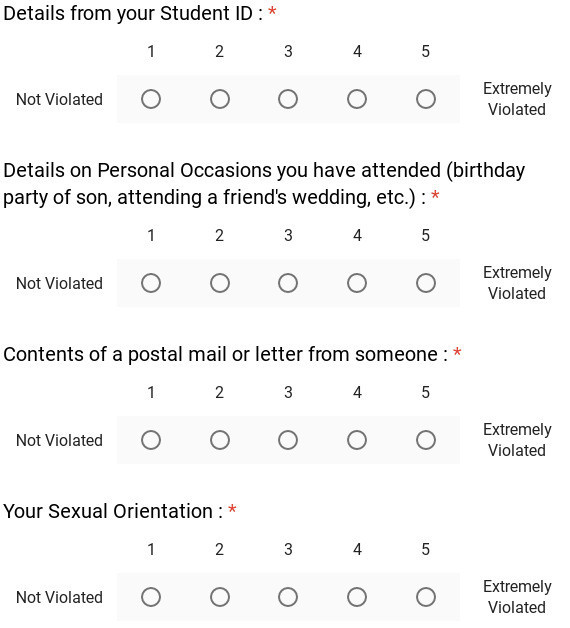}
\end{center}
   \caption{Questions from user study to understand privacy preferences}
\label{fig:p1_1}
\end{figure}

The task in this user study is to obtain user preferences over the 67 privacy attributes (excludes the attribute \emph{safe}).
The questionnaire instructs the user on a fictitious website (similar to Flickr or Twitter), where content posted is by default visible to everyone else on the platform.
By unintentionally posting information about a particular attribute, the user exposes private information comprising his/her anonymity.
Each question is a verbal description of one of the attributes (Figure \ref{fig:p1_1}).
We collect responses on a scale of 1-5 of how much the user finds his/her privacy violated as a consequence of this action.

\subsubsection*{Instructions provided to the Users}

\begin{em}
In this academic survey we want to understand how sensitive you are to certain details of your personal or private life. For instance, are you more comfortable sharing your full name, gender or details on your personal relationships?

We refer to these details of your personal or private life as "Personally Identifiable Information" (PII).

PII is information that can be used on its own or with other information to identify, contact, or locate a single person, or to identify an individual. Such information could be one or more of your: Full Name, Home Address, Political Opinion, etc.

 Following this description are a list of PIIs. For each of these PIIs, consider the following situation:
 On an online public platform, you create an anonymous account. On this platform, once you post something, you cannot delete it. Only the moderators can delete this post. However, they can be extremely slow and unresponsive. One day, you unintentionally shared/posted this PII about yourself. Immediately, you realize that you cannot delete this post.

 On a scale of 1-5, please rate how much you feel your privacy is violated by this action, where: \\
 1 - I feel my privacy is not violated. So, I wouldn't care. \\
 2 - I feel my privacy is slightly violated. However, it's not worth taking any action. \\
 3 - I feel my privacy is somewhat violated. I will message the moderator. In case there's no response, I will give up. \\
 4 - I feel my privacy is violated. I will inform the moderator and follow up for a few days. In case there's no response after that, I will give up. \\
 5 - I feel my privacy is extremely violated. I will not give up until this post is deleted.
\end{em}
 

\subsection{Users and Visual Privacy Judgment}

\begin{figure}
\begin{center}
   \includegraphics[width=\linewidth]{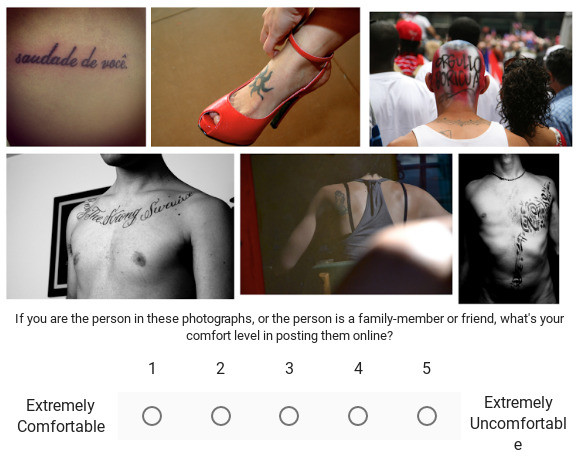}
   \includegraphics[width=\linewidth]{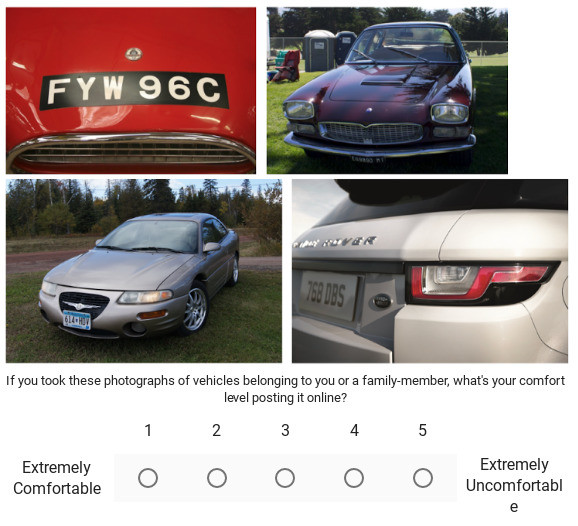}
\end{center}
   \caption{Questions from the user study to evaluate user privacy judgment}
\label{fig:p2_3}
\end{figure}

In order to understand how good are users at identifying privacy risks from images, we conduct this user study in two parts.
In the first part, we instruct users on a fictitious photo-sharing website, where images shared are publicly available.
For each of the 68 privacy attributes, we present a question on a group of images from the dataset representing this attribute (Figure \ref{fig:p2_3}).
The user responds how comfortable he/she is posting such images on the website.
The exact instructions for this part is provided below.

In the second part, we obtain user preferences over the attributes following the exact instructions in the previous section.

\subsubsection*{Instructions provided to the Users}

\begin{em}
In this academic survey we want to understand your comfort level sharing things on the internet. 

 Following this description are groups  of images. For each of these groups of images, consider the following situation:
 On an online public platform, you create an account. On this platform, you are allowed to post photographs, which anyone can view. Moreover, you can also interact with other users who shared their photographs and can comment on or like them.

 Important: For each of the below groups of images, picture yourself as either being the subject in the photograph, or the one who took the photograph of a family-member.

 On a scale of 1-5, rate how comfortable you are sharing such photographs, where: \\
 1 - You are extremely comfortable sharing such photographs \\
 2 - You are slightly comfortable sharing such photographs \\
 3 - You are somewhat comfortable sharing such photographs \\
 4 - You are not comfortable sharing such photographs \\
 5 - You are extremely uncomfortable sharing such photographs
\end{em}


\section{Additional Qualitative Examples for Privacy Attribute Prediction}
\label{sec:appendix_pap_qual}

\begin{figure*}
\centering
\begin{tabular}{ @{} c  >{\centering\arraybackslash} m{4cm}  >{\centering\arraybackslash} m{4cm} >{\centering\arraybackslash} m{4cm} @{}}
 & True Positives & False Positives & False Negatives \\
Credit Card 
&  \includegraphics[width=0.11\textwidth]{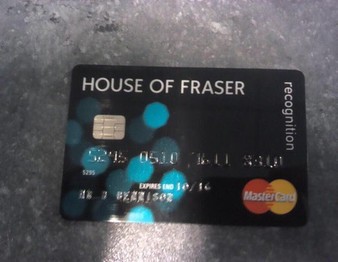}
   \includegraphics[width=0.11\textwidth]{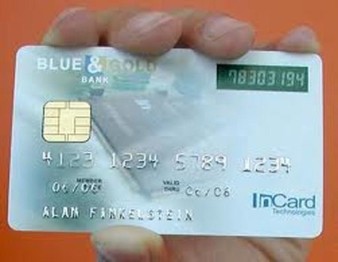} \qquad 
&  \includegraphics[width=0.11\textwidth]{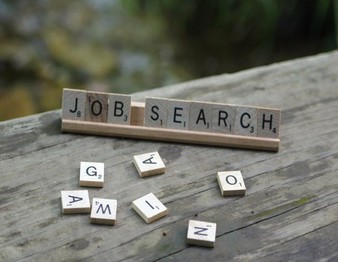}
   \includegraphics[width=0.11\textwidth]{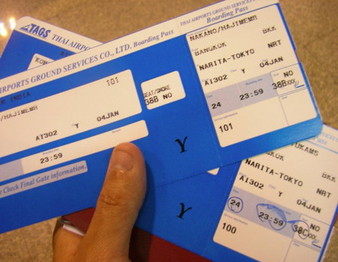}
&  \includegraphics[width=0.11\textwidth]{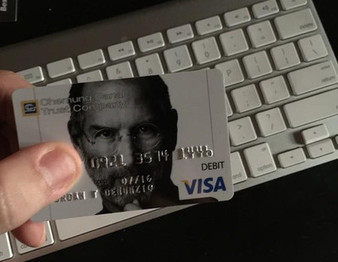}
   \includegraphics[width=0.11\textwidth]{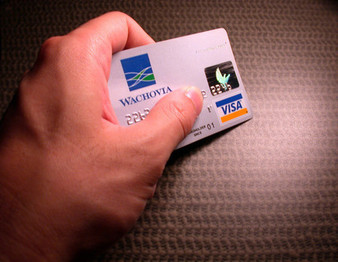} \\
\begin{tabular}[c]{@{}c@{}}Ethnic\\ Clothing\end{tabular} 
&  \includegraphics[width=0.11\textwidth]{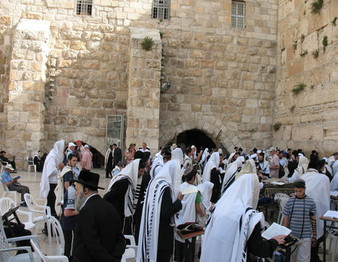}
   \includegraphics[width=0.11\textwidth]{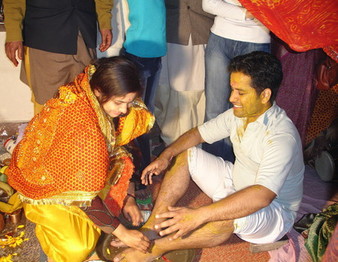} \qquad
&  \includegraphics[width=0.11\textwidth]{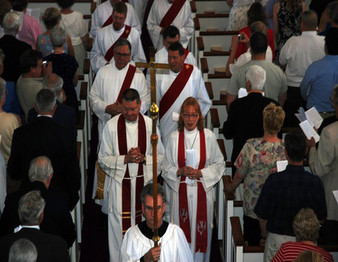}
   \includegraphics[width=0.11\textwidth]{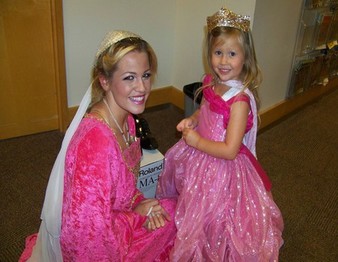}
&  \includegraphics[width=0.11\textwidth]{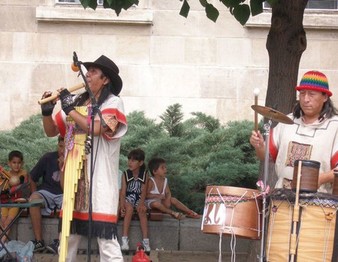}
   \includegraphics[width=0.11\textwidth]{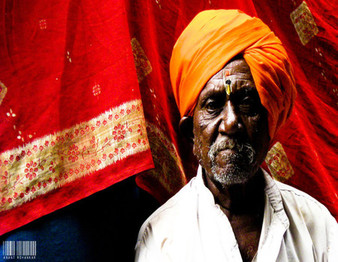} \\
Full Name 
&  \includegraphics[width=0.11\textwidth]{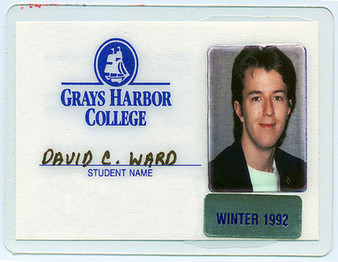}
   \includegraphics[width=0.11\textwidth]{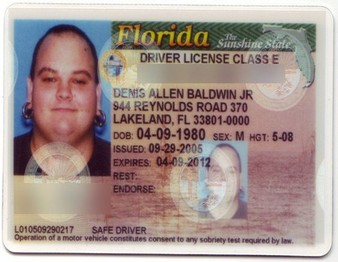} \qquad
&  \includegraphics[width=0.11\textwidth]{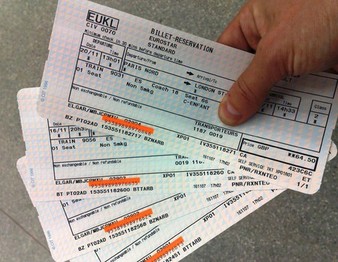}
   \includegraphics[width=0.11\textwidth]{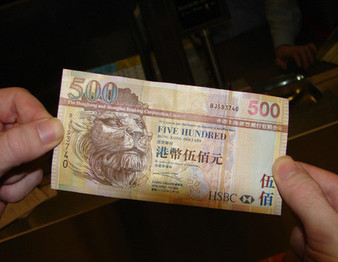}
&  \includegraphics[width=0.11\textwidth]{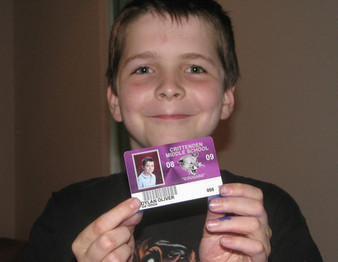}
   \includegraphics[width=0.11\textwidth]{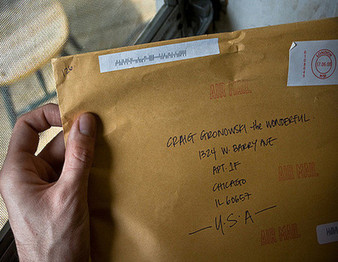} \\
Hobbies 
&  \includegraphics[width=0.11\textwidth]{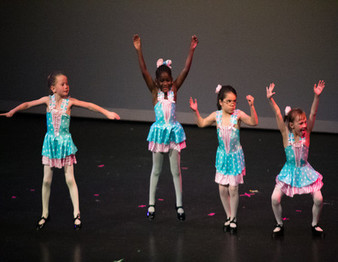}
   \includegraphics[width=0.11\textwidth]{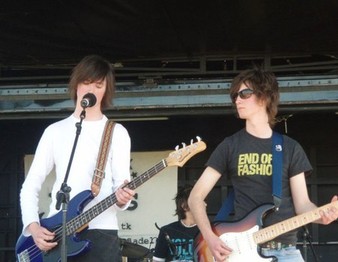} \qquad
&  \includegraphics[width=0.11\textwidth]{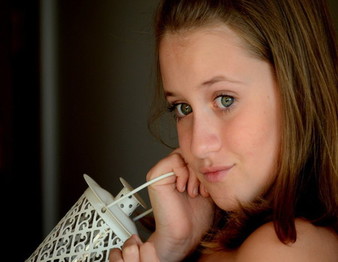}
   \includegraphics[width=0.11\textwidth]{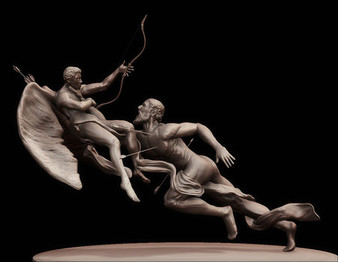}
&  \includegraphics[width=0.11\textwidth]{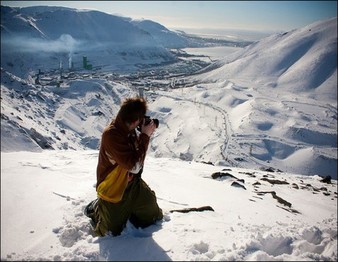}
   \includegraphics[width=0.11\textwidth]{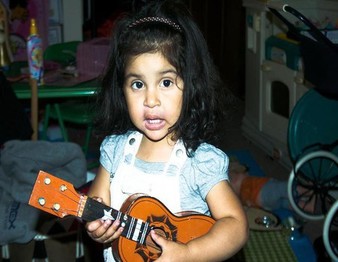} \\
Passport 
&  \includegraphics[width=0.11\textwidth]{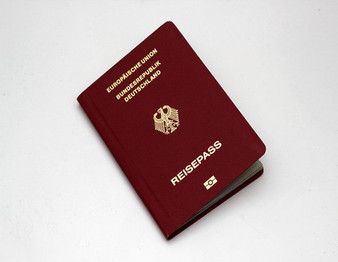}
   \includegraphics[width=0.11\textwidth]{fig/pap_qual/pport_tp_2.jpg} \qquad
&  \includegraphics[width=0.11\textwidth]{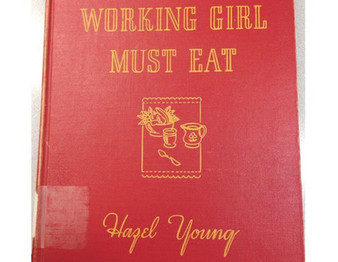}
   \includegraphics[width=0.11\textwidth]{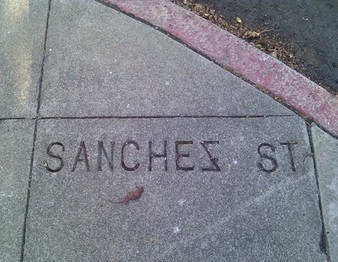}
&  \includegraphics[width=0.11\textwidth]{fig/pap_qual/pport_fn_1.jpg}
   \includegraphics[width=0.11\textwidth]{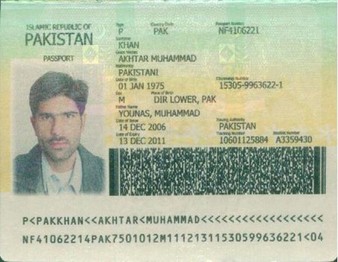} \\
\begin{tabular}[c]{@{}c@{}}Sexual\\ Orientation\end{tabular} 
&  \includegraphics[width=0.11\textwidth]{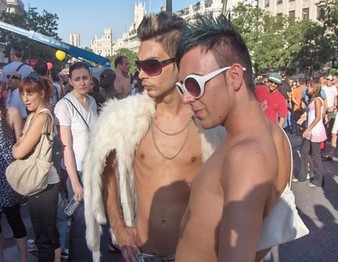}
   \includegraphics[width=0.11\textwidth]{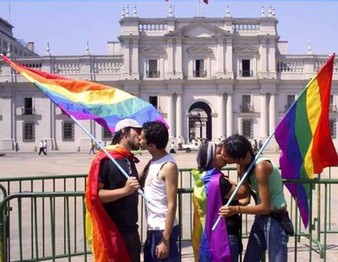} \qquad
&  \includegraphics[width=0.11\textwidth]{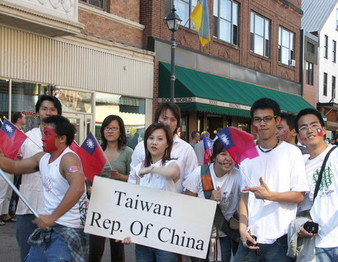}
   \includegraphics[width=0.11\textwidth]{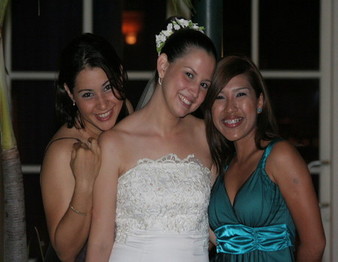}
&  \includegraphics[width=0.11\textwidth]{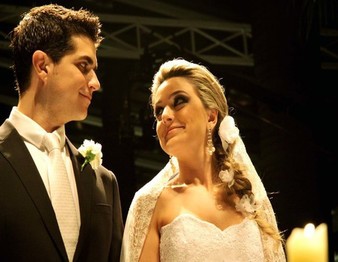}
   \includegraphics[width=0.11\textwidth]{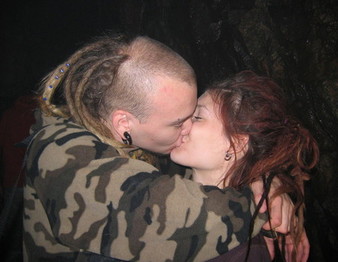} \\
\begin{tabular}[c]{@{}c@{}}Medical\\ History\end{tabular} 
&  \includegraphics[width=0.11\textwidth]{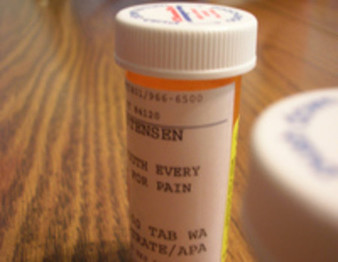}
   \includegraphics[width=0.11\textwidth]{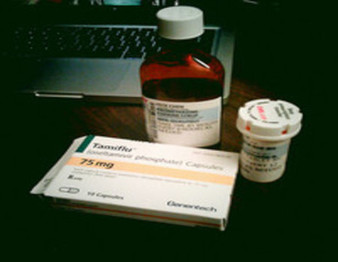}
&  \includegraphics[width=0.11\textwidth]{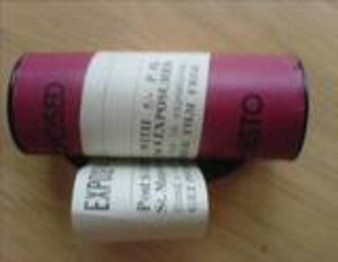}
   \includegraphics[width=0.11\textwidth]{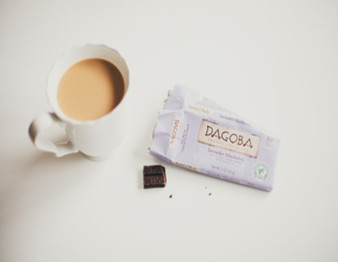}
&  \includegraphics[width=0.11\textwidth]{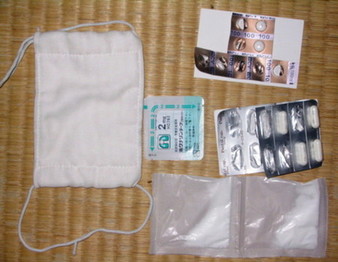}
   \includegraphics[width=0.11\textwidth]{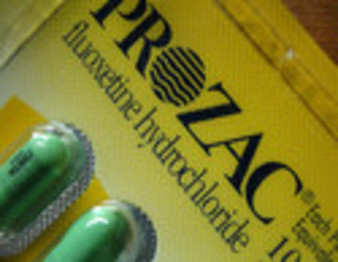} \\
\begin{tabular}[c]{@{}c@{}}Drivers\\ License\end{tabular} 
&  \includegraphics[width=0.11\textwidth]{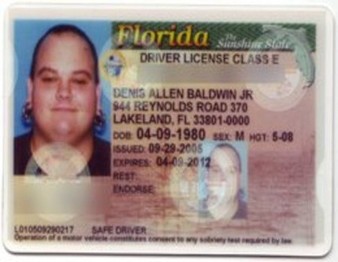}
   \includegraphics[width=0.11\textwidth]{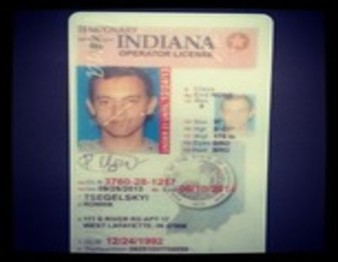}
&  \includegraphics[width=0.11\textwidth]{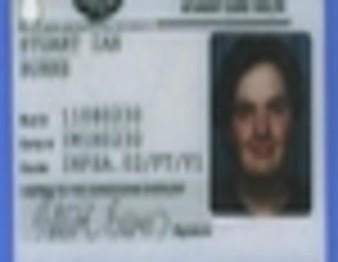}
   \includegraphics[width=0.11\textwidth]{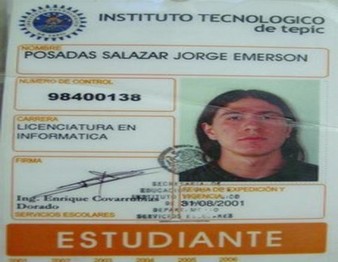}
&  \includegraphics[width=0.11\textwidth]{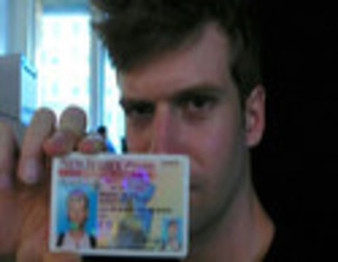}
   \includegraphics[width=0.11\textwidth]{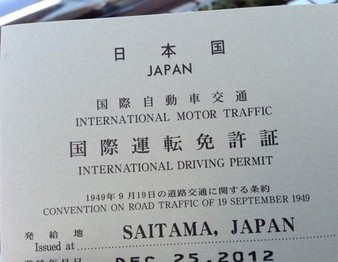} \\
Handwriting
&  \includegraphics[width=0.11\textwidth]{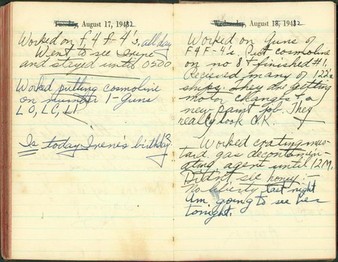}
   \includegraphics[width=0.11\textwidth]{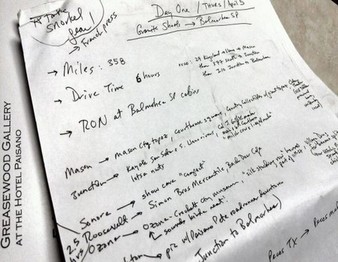}
&  \includegraphics[width=0.11\textwidth]{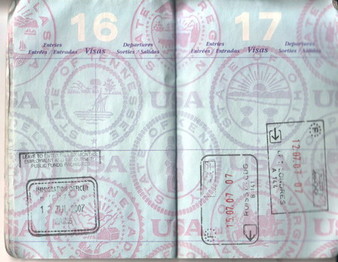}
   \includegraphics[width=0.11\textwidth]{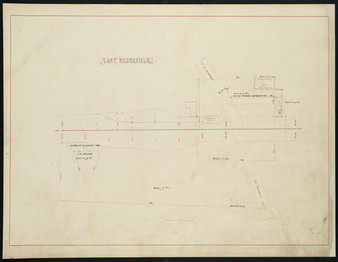}
&  \includegraphics[width=0.11\textwidth]{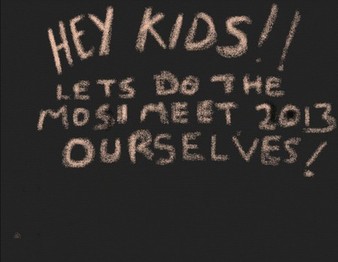}
   \includegraphics[width=0.11\textwidth]{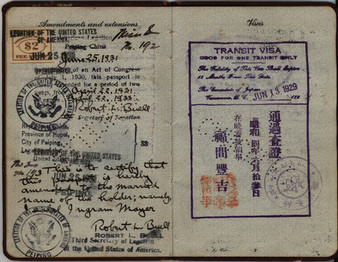} \\
Occupation 
&  \includegraphics[width=0.11\textwidth]{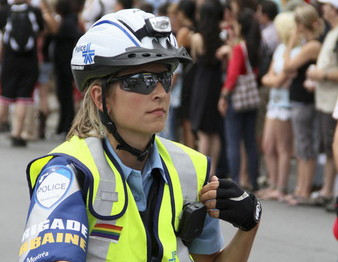}
   \includegraphics[width=0.11\textwidth]{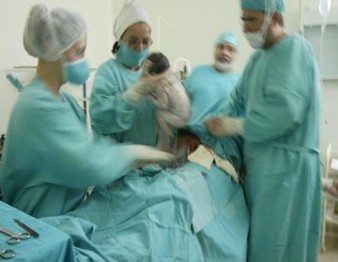}
&  \includegraphics[width=0.11\textwidth]{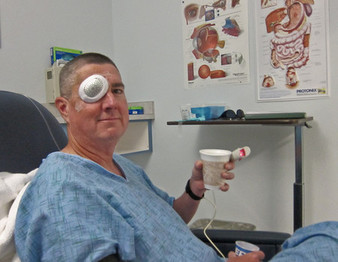}
   \includegraphics[width=0.11\textwidth]{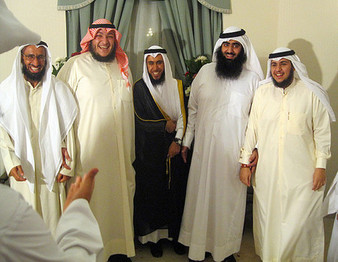}
&  \includegraphics[width=0.11\textwidth]{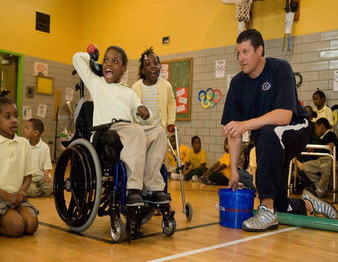}
   \includegraphics[width=0.11\textwidth]{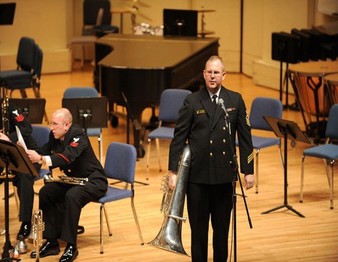} \\
\begin{tabular}[c]{@{}c@{}}Personal\\ Relationships\end{tabular} 
&  \includegraphics[width=0.11\textwidth]{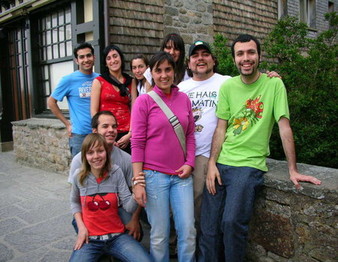}
   \includegraphics[width=0.11\textwidth]{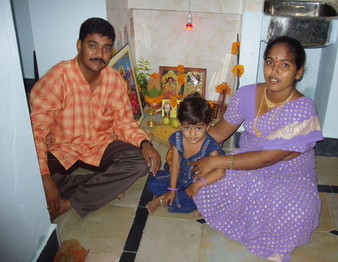}
&  \includegraphics[width=0.11\textwidth]{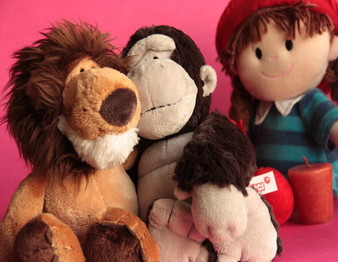}
   \includegraphics[width=0.11\textwidth]{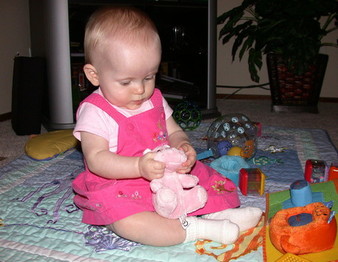}
&  \includegraphics[width=0.11\textwidth]{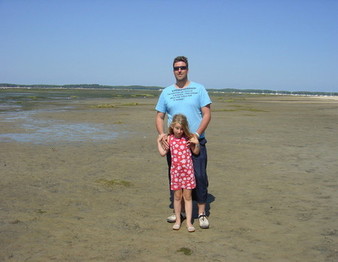}
   \includegraphics[width=0.11\textwidth]{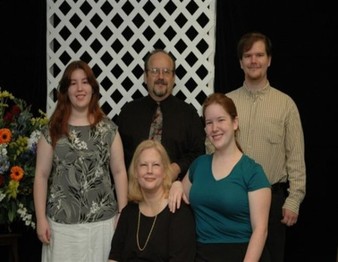} \\
Username 
&  \includegraphics[width=0.11\textwidth]{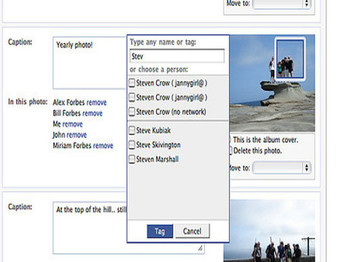}
   \includegraphics[width=0.11\textwidth]{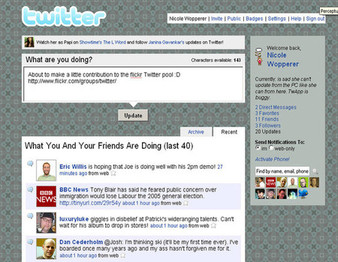}
&  \includegraphics[width=0.11\textwidth]{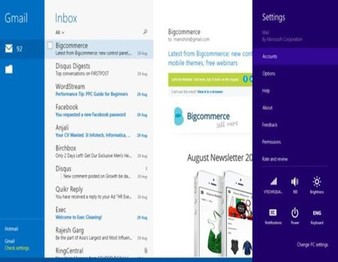}
   \includegraphics[width=0.11\textwidth]{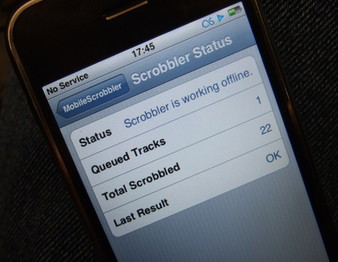}
&  \includegraphics[width=0.11\textwidth]{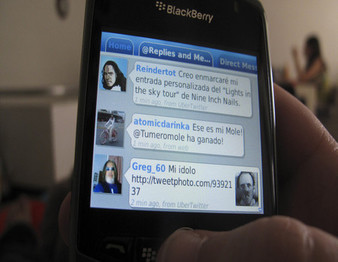}
   \includegraphics[width=0.11\textwidth]{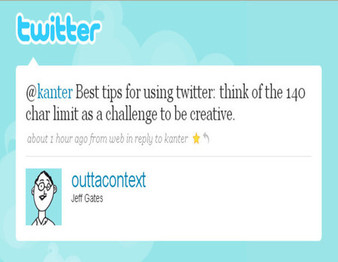} \\
\begin{tabular}[c]{@{}c@{}}License Plate\\ (Complete)\end{tabular} 
&  \includegraphics[width=0.11\textwidth]{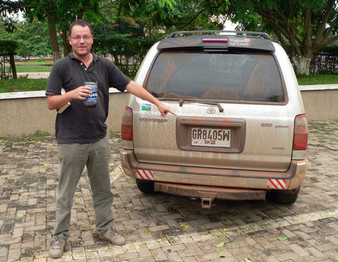}
   \includegraphics[width=0.11\textwidth]{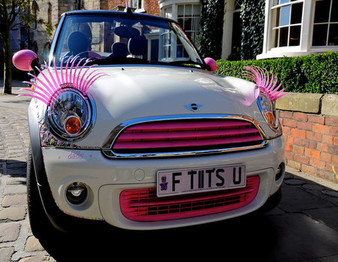}
&  \includegraphics[width=0.11\textwidth]{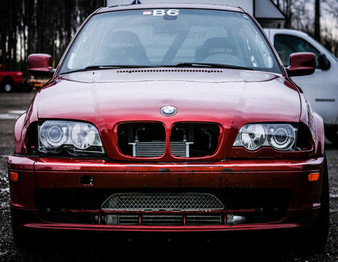}
   \includegraphics[width=0.11\textwidth]{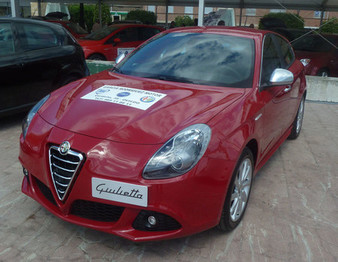}
&  \includegraphics[width=0.11\textwidth]{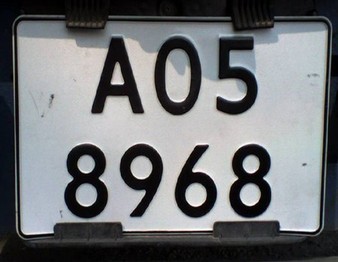}
   \includegraphics[width=0.11\textwidth]{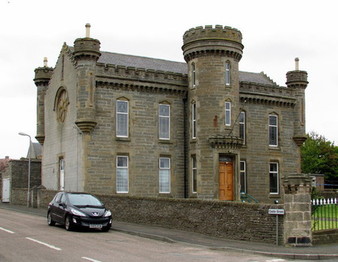} \\
\end{tabular}
\caption{Additional Qualitative Results of our Privacy Attribute Prediction method}
\label{fig:pap_qualitative_sup}
\end{figure*}

In Section 5.1 we discussed our approach to \emph{Privacy Attribute Prediction} -- a user-independent method of predicting  multiple privacy attributes given an image.
In this section, in addition to Figure 6, we present additional qualitative examples in \autoref{fig:pap_qualitative_sup}.
Each row represents images of a particular privacy attribute.
The True Positives column indicate the case when this attribute is in both the ground-truth and predicted set of privacy attributes.
The False Positives column indicate images when the attribute is incorrectly predicted.
The False Negatives column indicate images when the attribute is in ground-truth, but is not predicted.

We observe our method associates privacy attributes to distinctive visual cues such as clothing (for occupation and ethnic clothing), exposed skin (for tattoos, nudity), metallic objects with wheels (for physical disability, license plates) and text (for names, drivers license, username, handwriting).
As a result, apart from correct predictions, we find that this also leads to incorrectly predicting attributes (\eg predicting card-shaped identification documents as drivers licenses, cars for license plates) or failing to recognize attributes in a different context (\eg handwriting on a wall instead of documents, new types of drivers licenses).
We also observe our approach underperform in differentiating between full, first and last names, or usernames and email addresses (which requires text-based reasoning), identifying relationships and sexual orientation (which requires interpreting interaction between multiple people) and differentiating occupations, religion and ethnic clothing (which requires fine-grained recognition).

\section{Additional Results for Personalized Privacy Prediction}
\label{sec:appendix_profile_pr}

\begin{figure*}
\begin{center}
   \includegraphics[width=\textwidth]{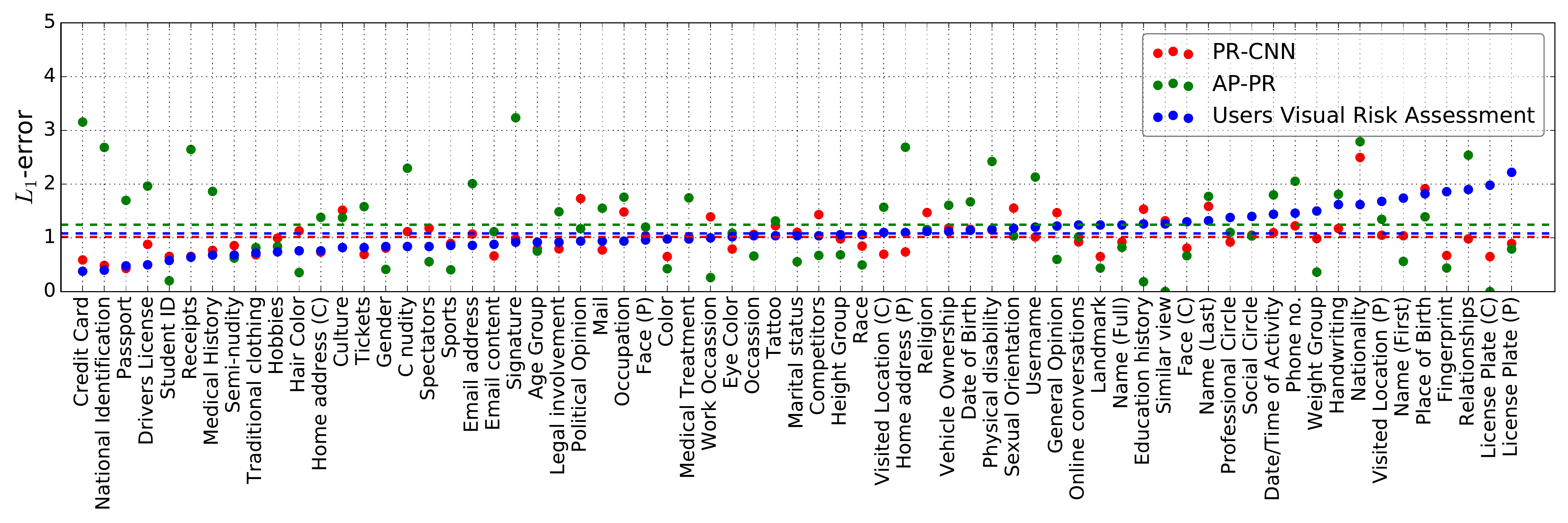}
\end{center}
   \caption{$L_1$ errors over attributes}
\label{fig:l1_attributes}
\end{figure*}

\subsection{Qualitative Results}
In this section, we discuss additional results for Section 5.2: Personalizing Privacy Risk Prediction.

\begin{figure*}
\centering
\begin{tabular}{ccccccc}
   \includegraphics[width=0.15\textwidth]{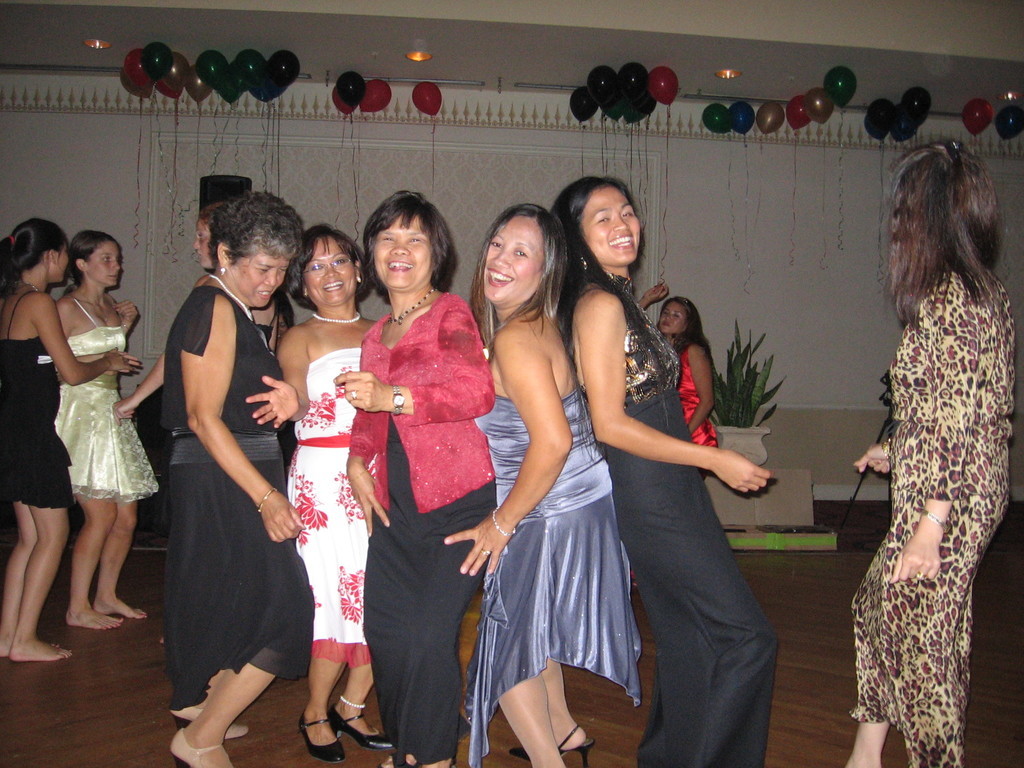}
 &  
 & \includegraphics[width=0.15\textwidth]{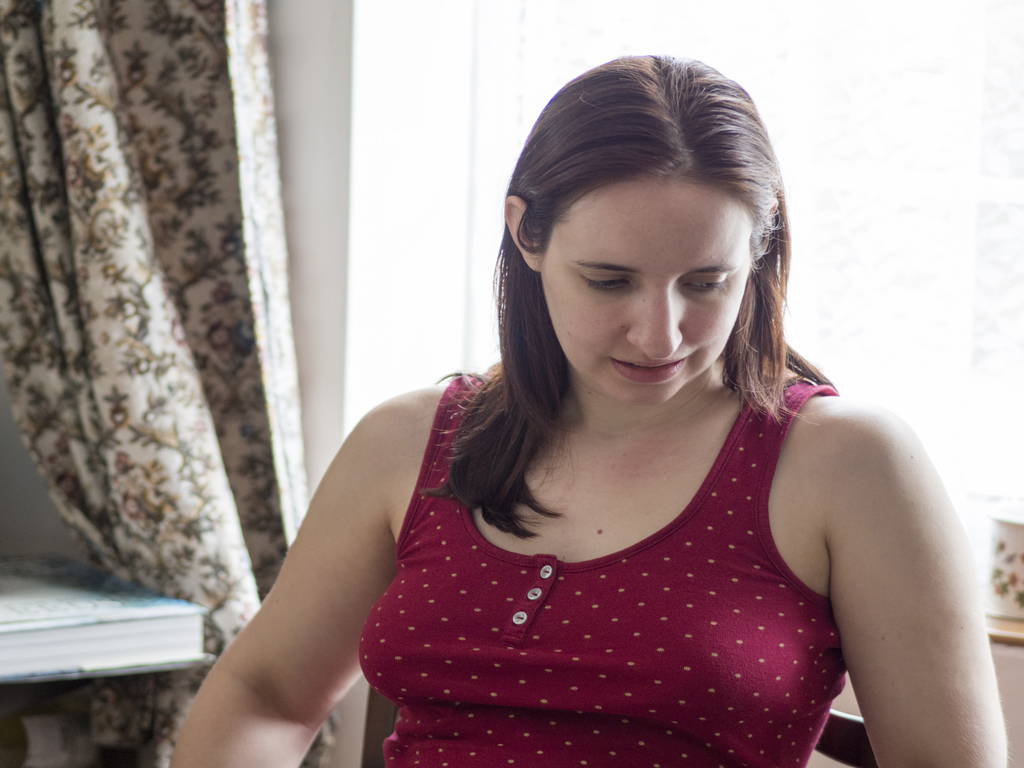} 
 &  
 & \includegraphics[width=0.15\textwidth]{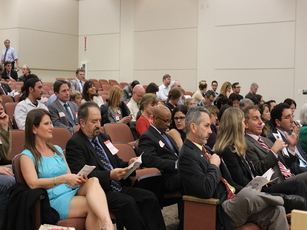}
 & 
 & \includegraphics[width=0.15\textwidth]{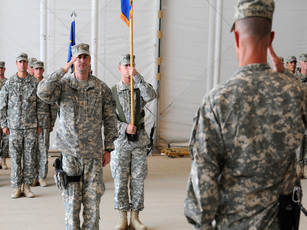} \\
   \includegraphics[width=0.15\textwidth]{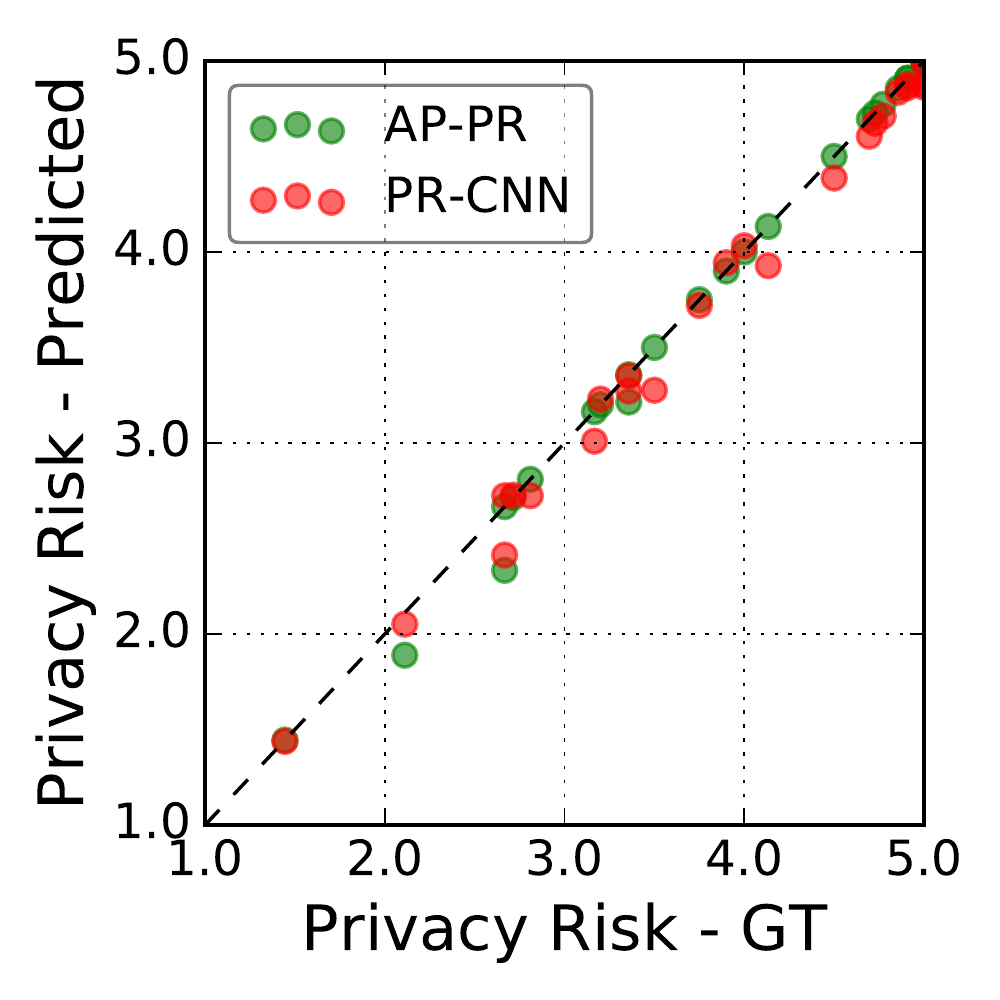} 
 & 
 &  \includegraphics[width=0.15\textwidth]{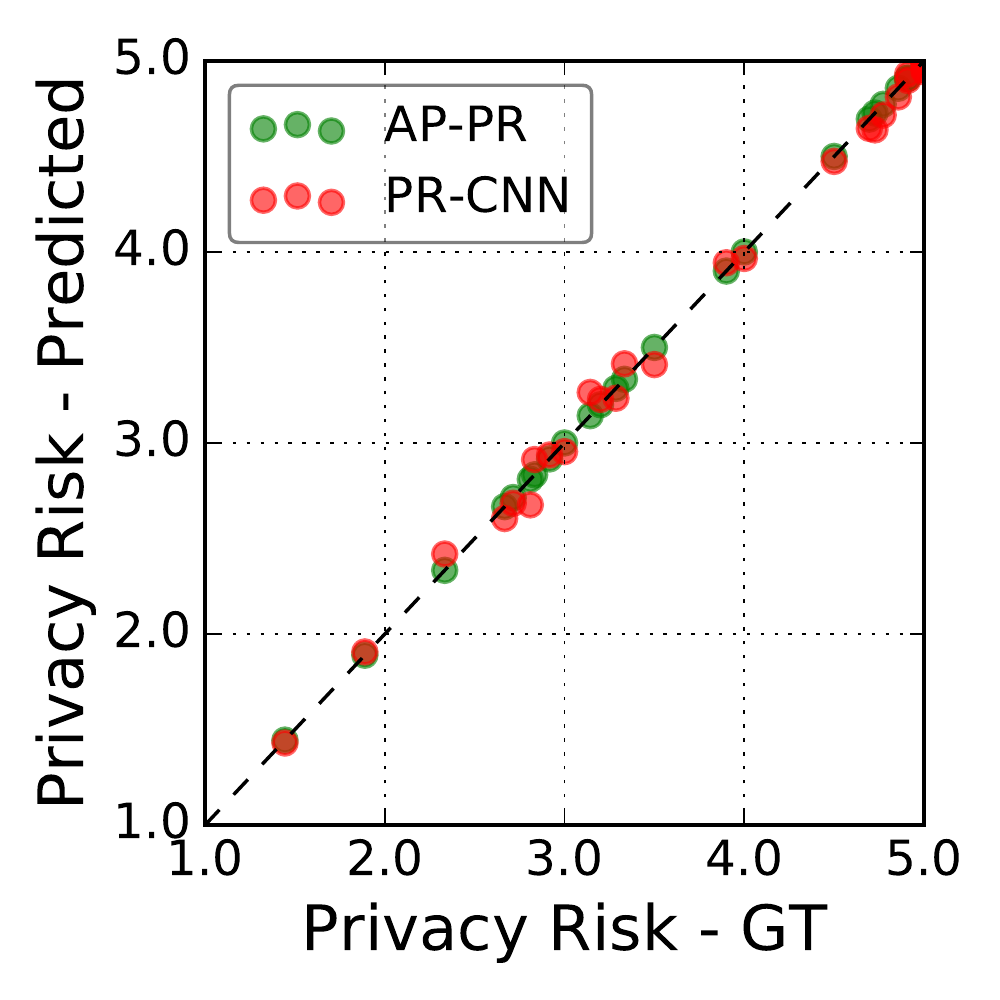} 
 & 
 & \includegraphics[width=0.15\textwidth]{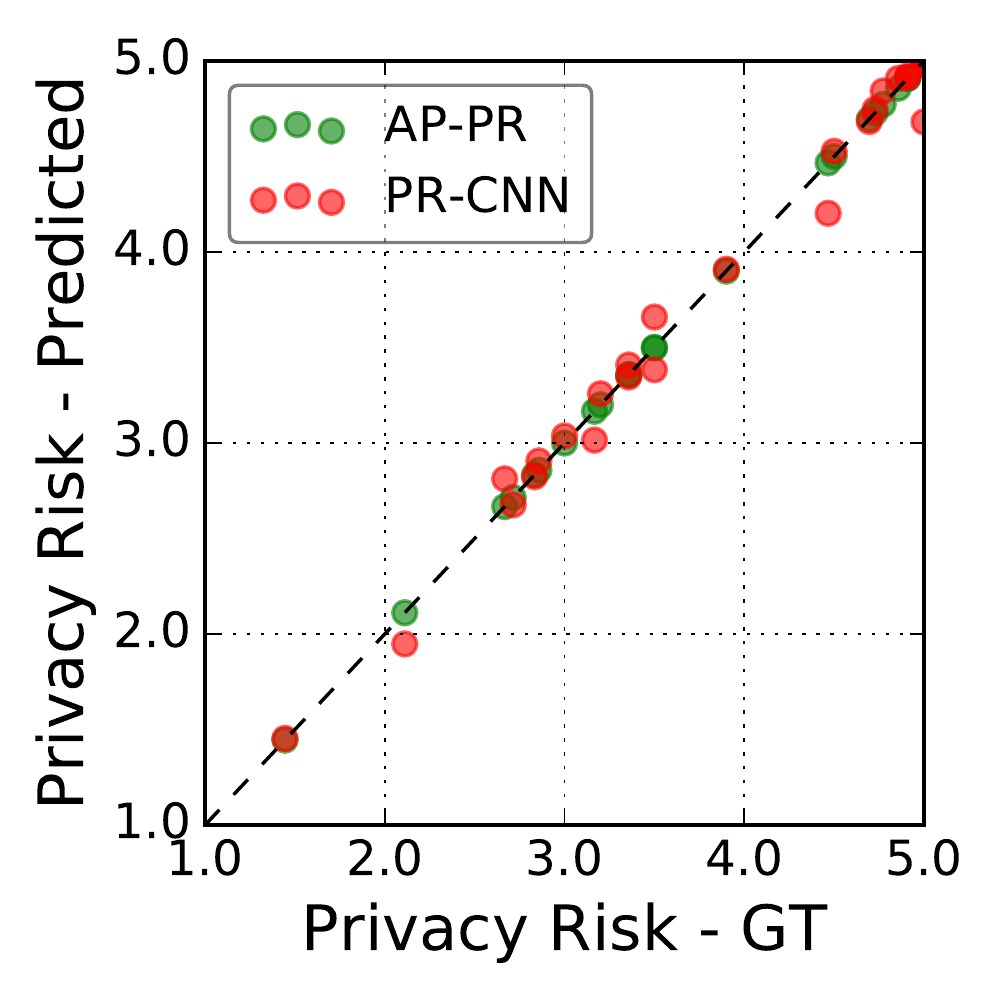} 
 & 
 & \includegraphics[width=0.15\textwidth]{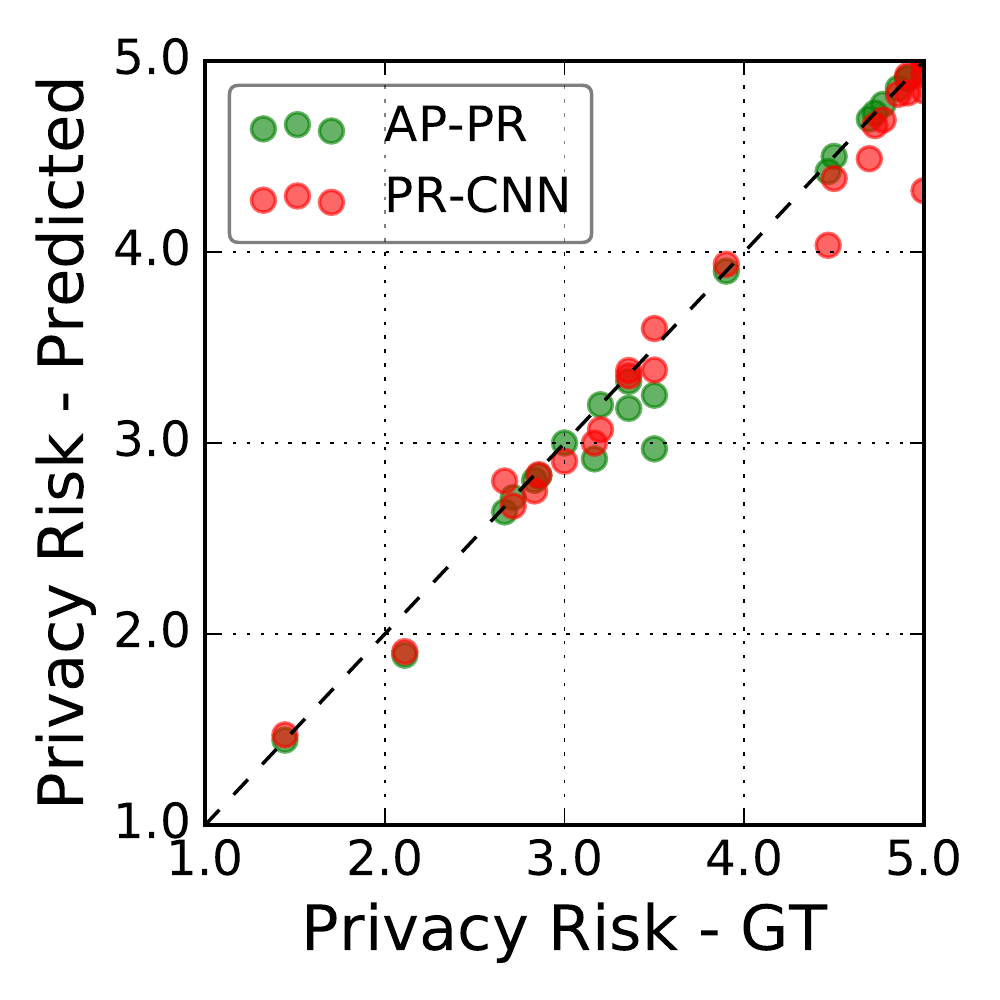}  \\
 & \vspace{1em} & & & & & \\
   \includegraphics[width=0.15\textwidth]{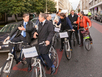}
 &  
 & \includegraphics[width=0.15\textwidth]{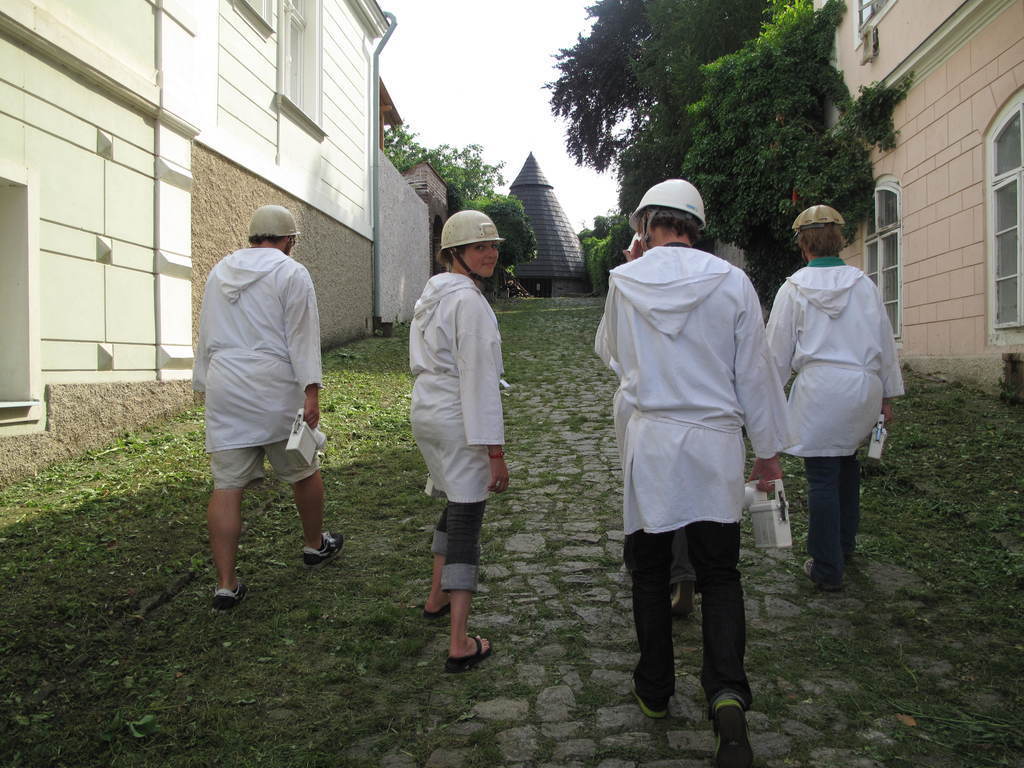} 
 &  
 & \includegraphics[width=0.15\textwidth]{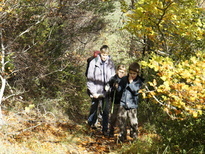}
 & 
 & \includegraphics[width=0.15\textwidth]{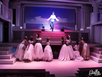} \\
 
   \includegraphics[width=0.15\textwidth]{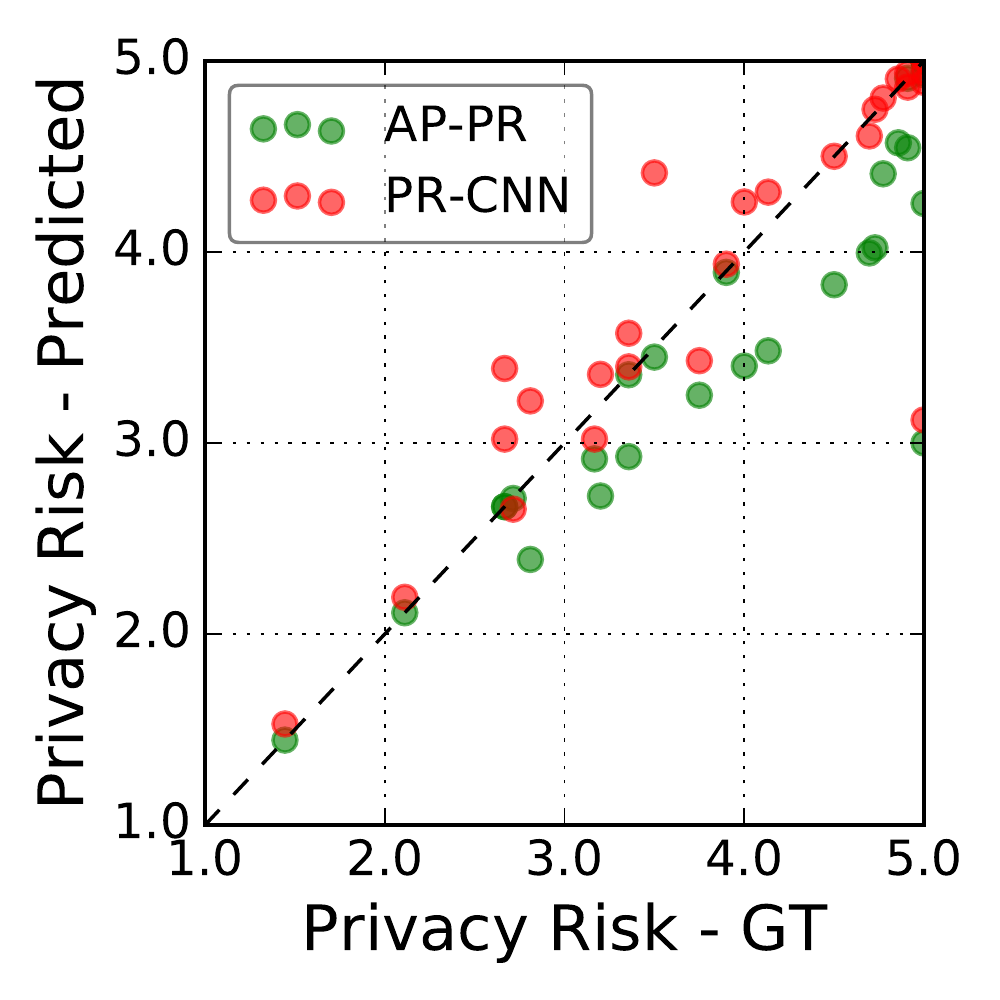} 
 & 
 &  \includegraphics[width=0.15\textwidth]{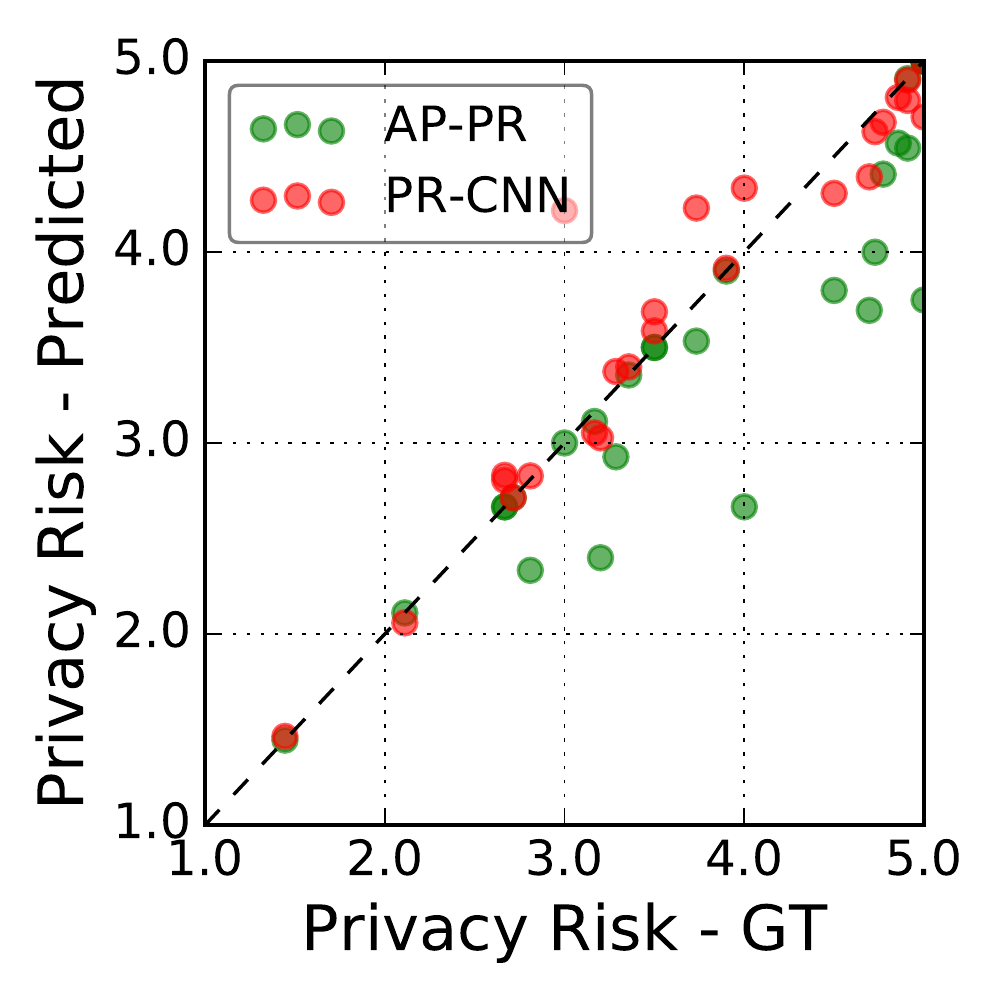} 
 & 
 & \includegraphics[width=0.15\textwidth]{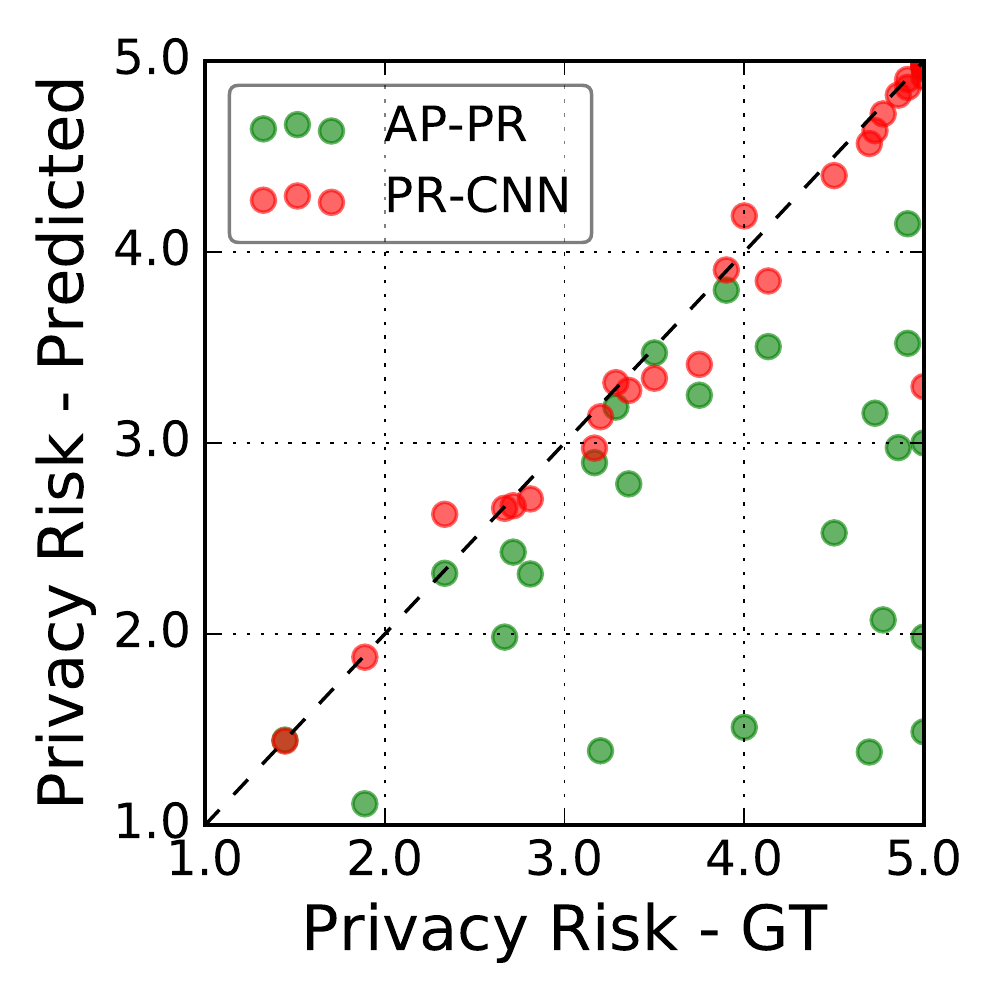} 
 & 
 & \includegraphics[width=0.15\textwidth]{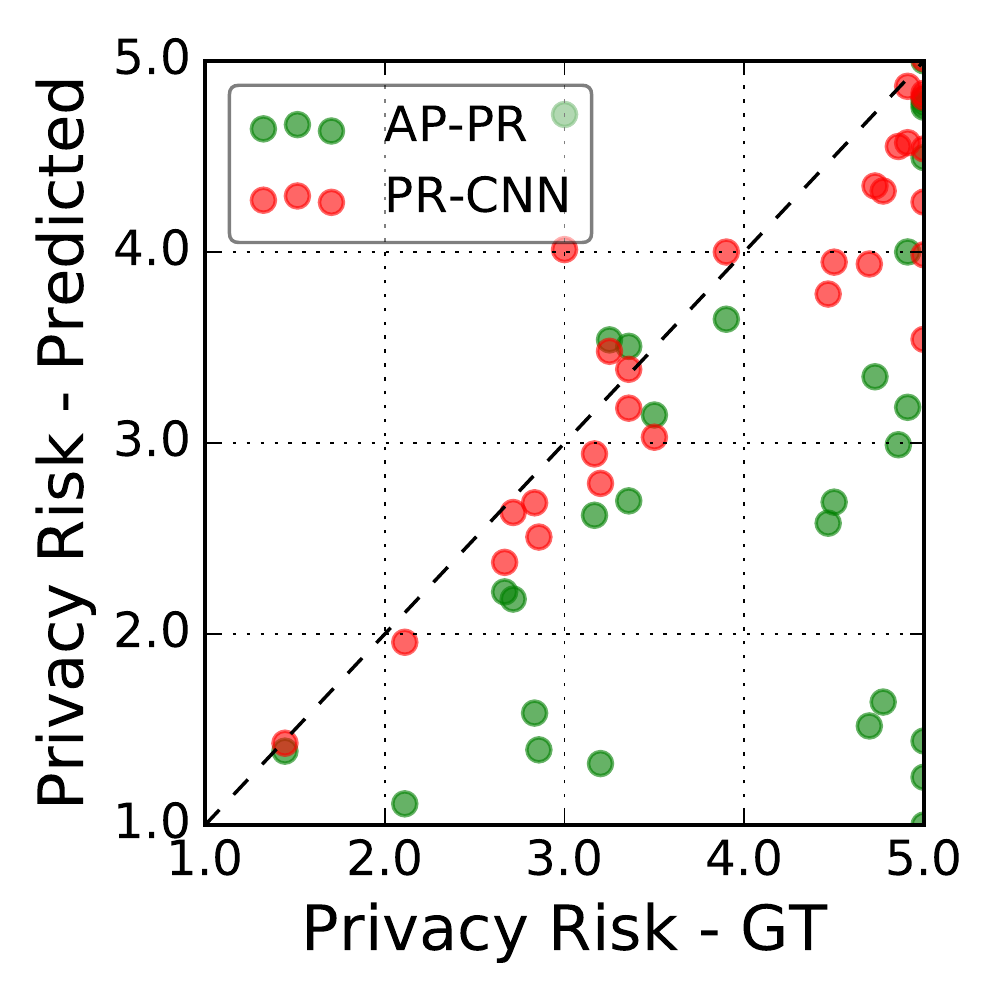}  \\
 & \vspace{1em} & & & & & \\
   \includegraphics[width=0.15\textwidth]{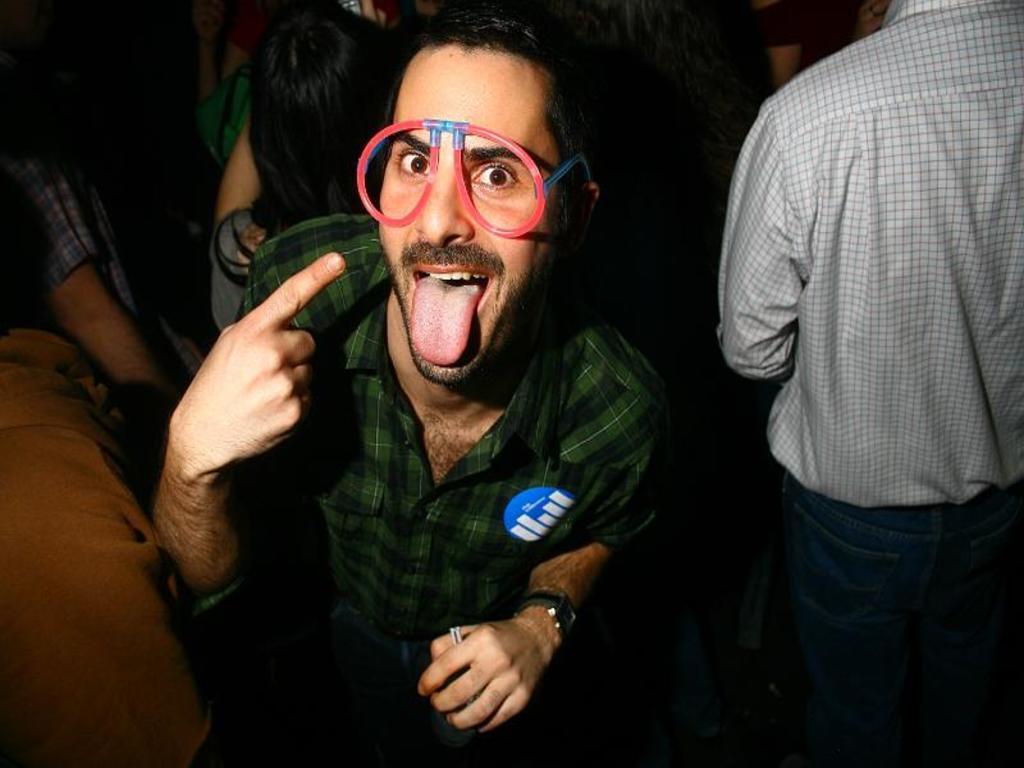}
 &  
 & \includegraphics[width=0.15\textwidth]{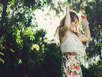} 
 &  
 & \includegraphics[width=0.15\textwidth]{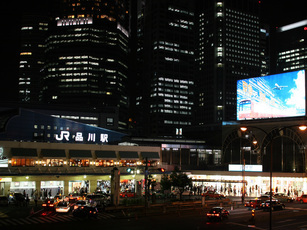}
 & 
 & \includegraphics[width=0.15\textwidth]{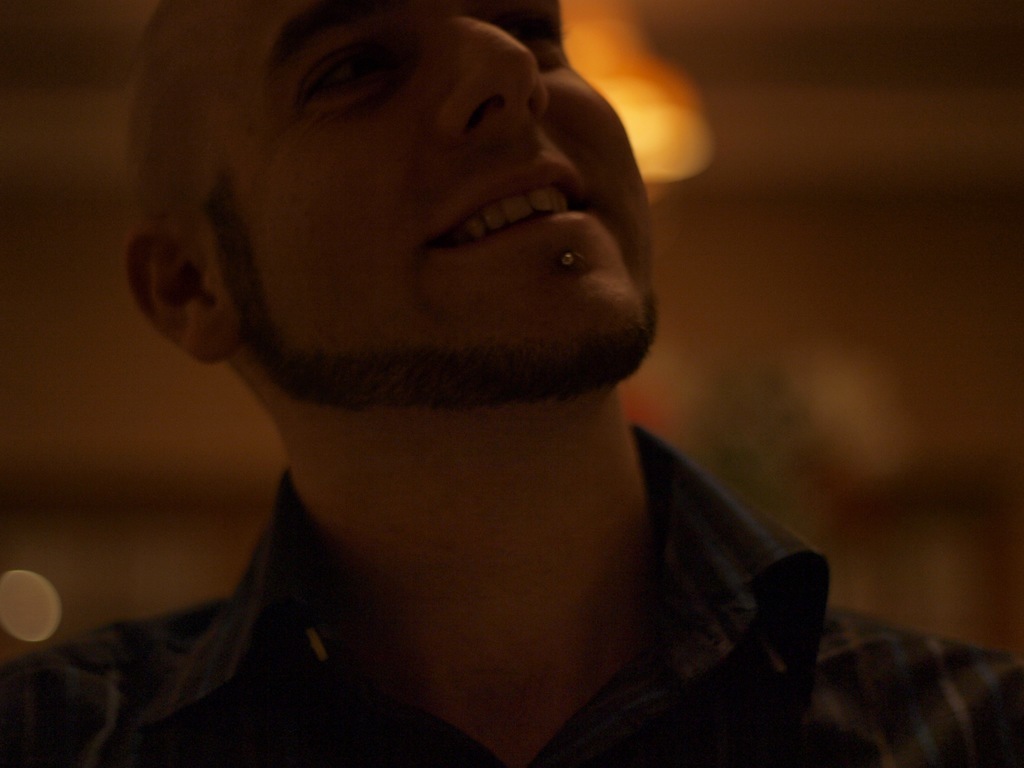} \\
 
   \includegraphics[width=0.15\textwidth]{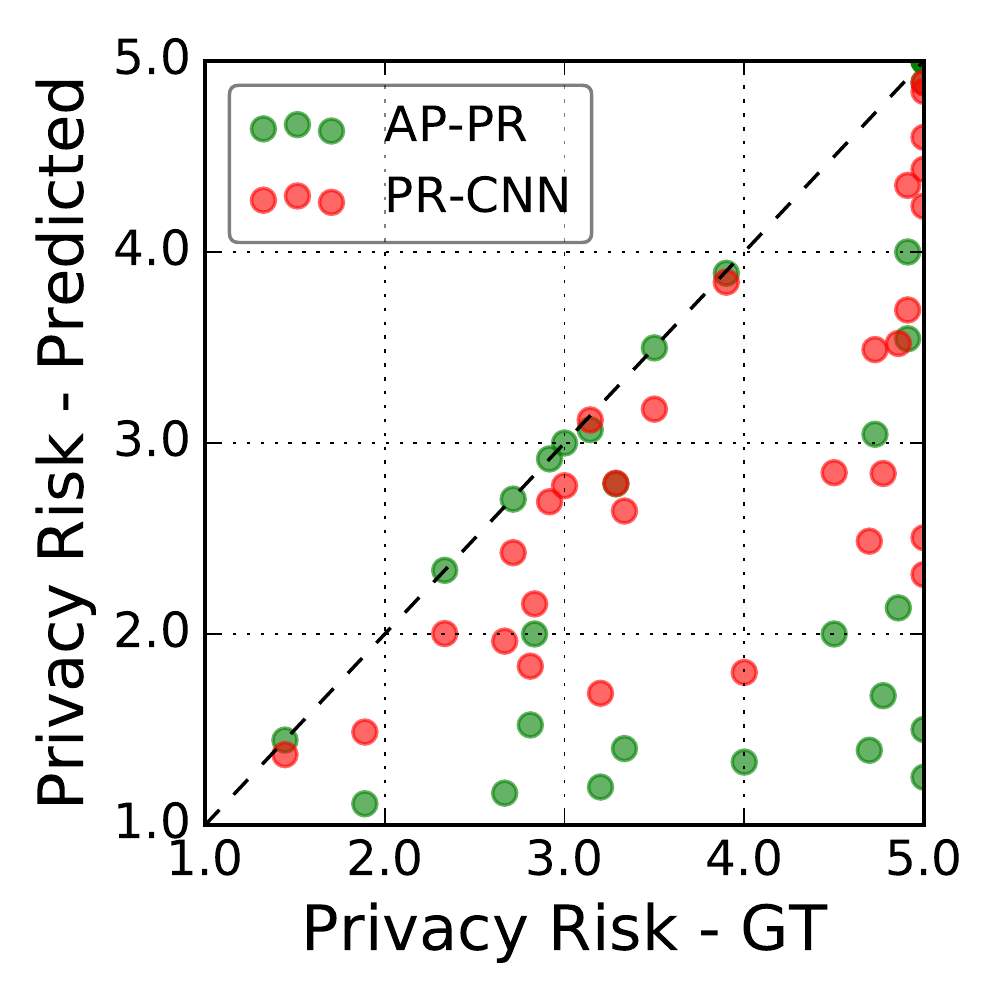} 
 & 
 &  \includegraphics[width=0.15\textwidth]{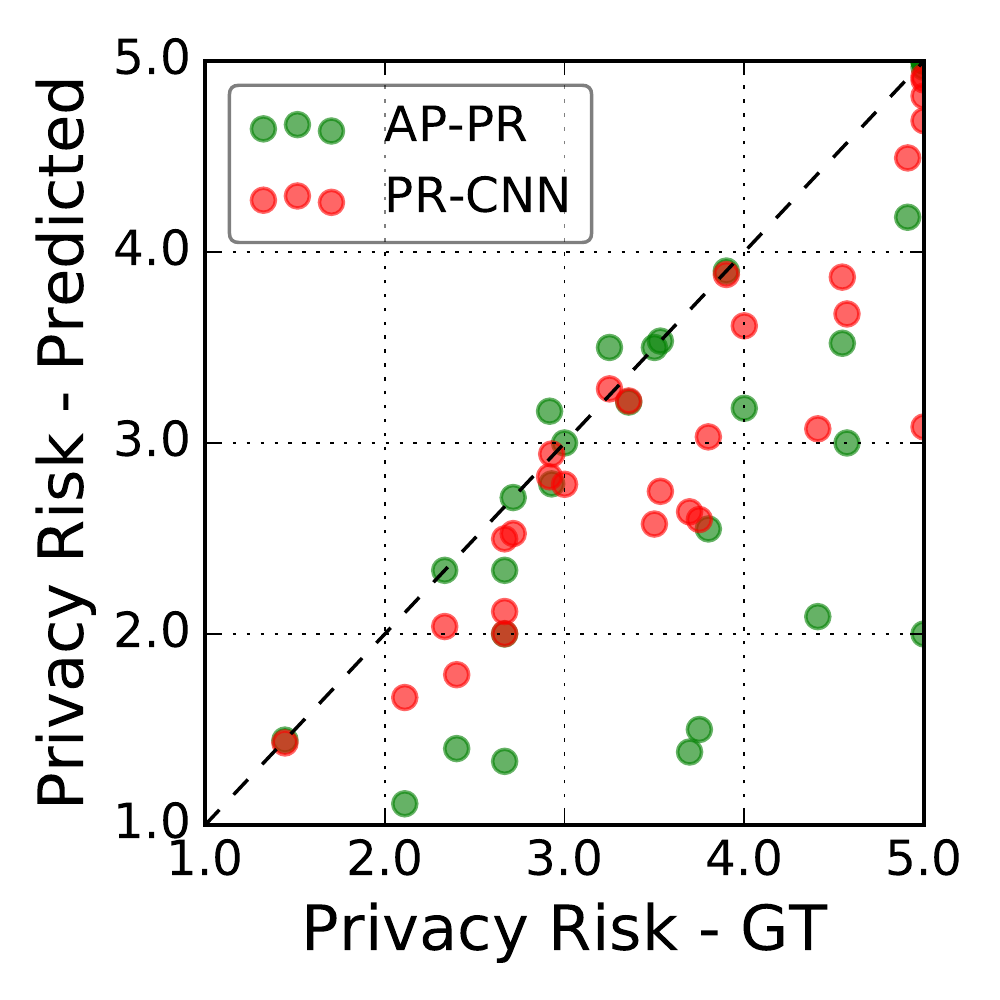} 
 & 
 & \includegraphics[width=0.15\textwidth]{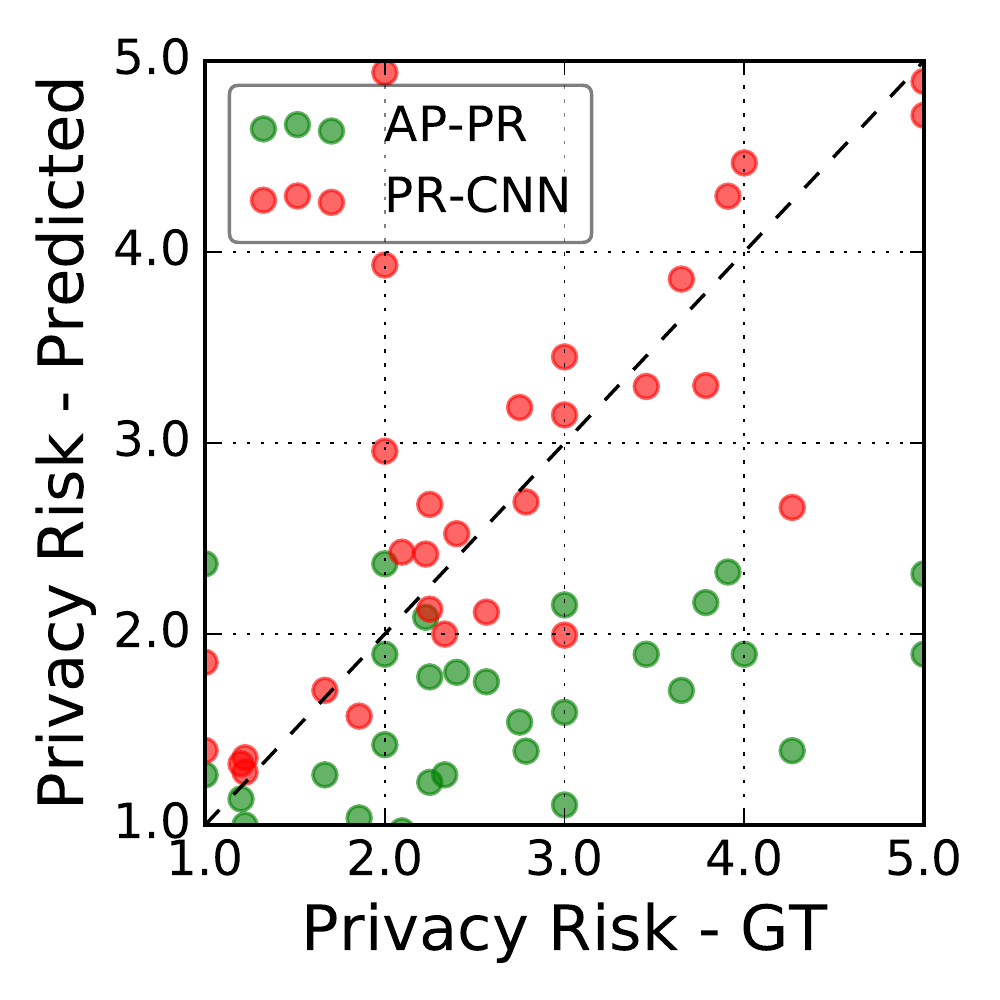} 
 & 
 & \includegraphics[width=0.15\textwidth]{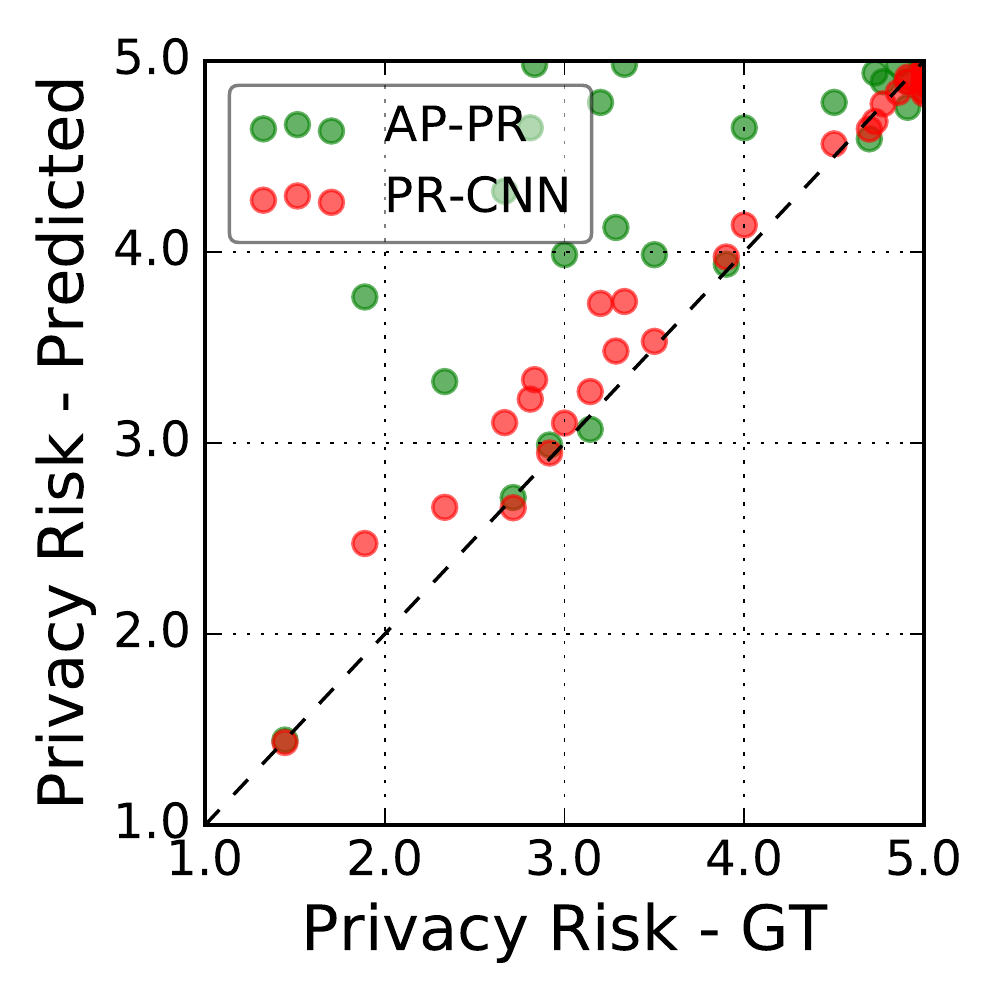}  \\
 & \vspace{1em} & & & & & \\
   \includegraphics[width=0.15\textwidth]{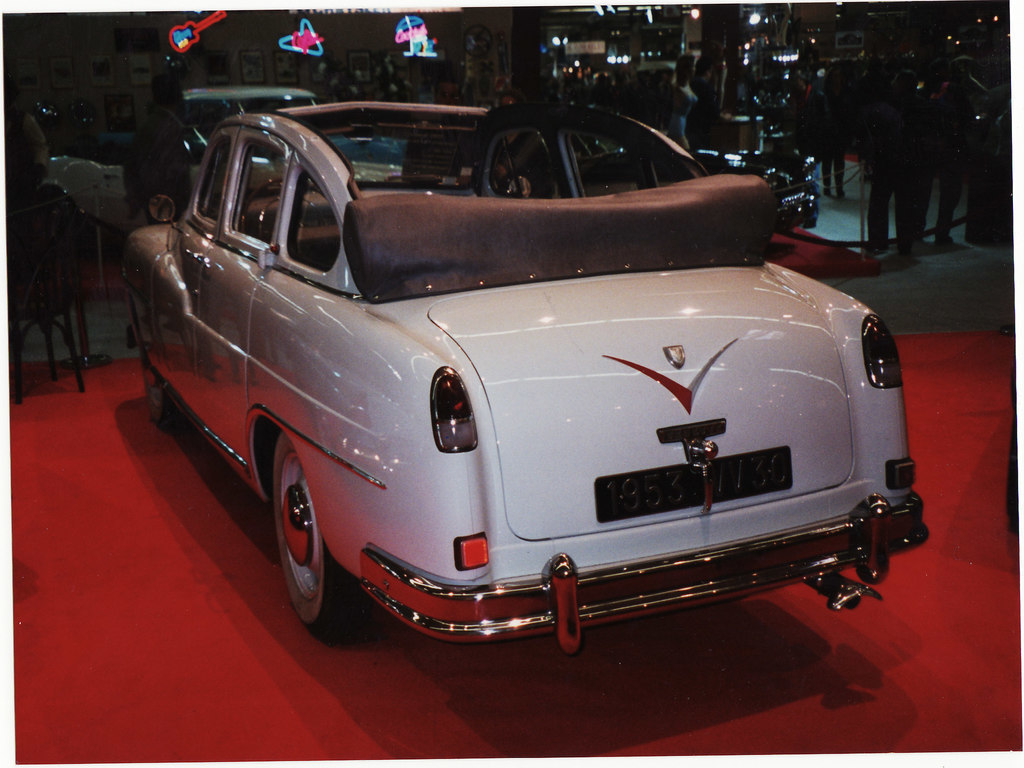}
 &  
 & \includegraphics[width=0.15\textwidth]{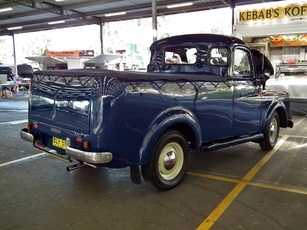} 
 &  
 & \includegraphics[width=0.15\textwidth]{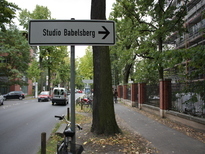}
 & 
 & \includegraphics[width=0.15\textwidth]{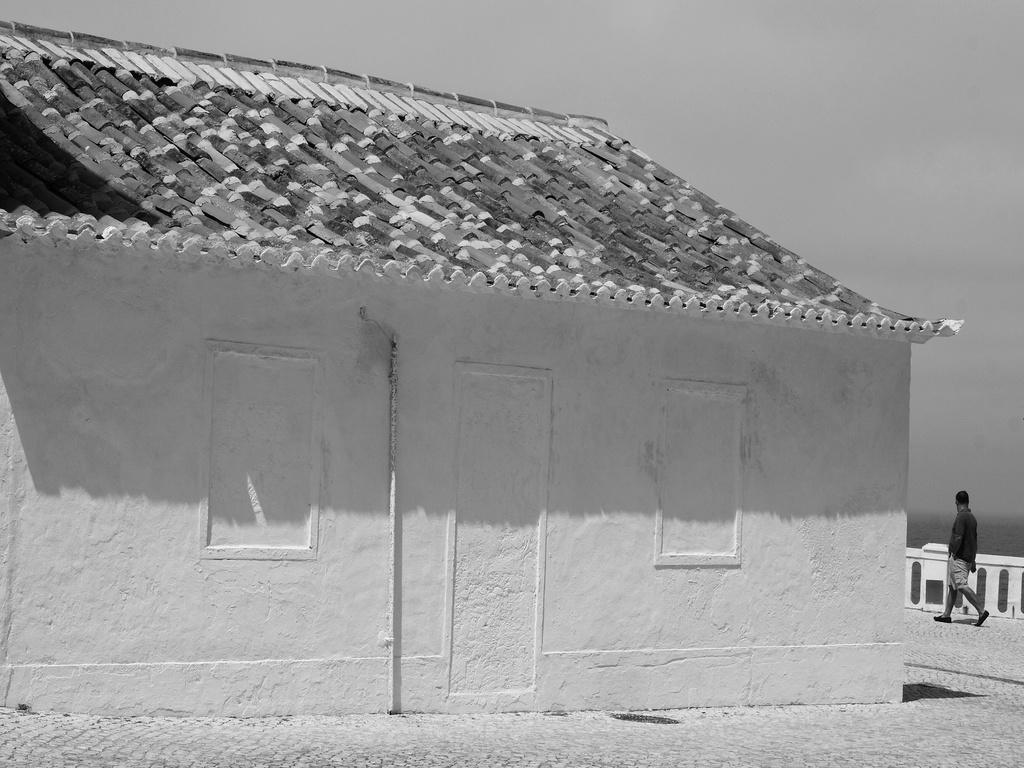} \\
 
   \includegraphics[width=0.15\textwidth]{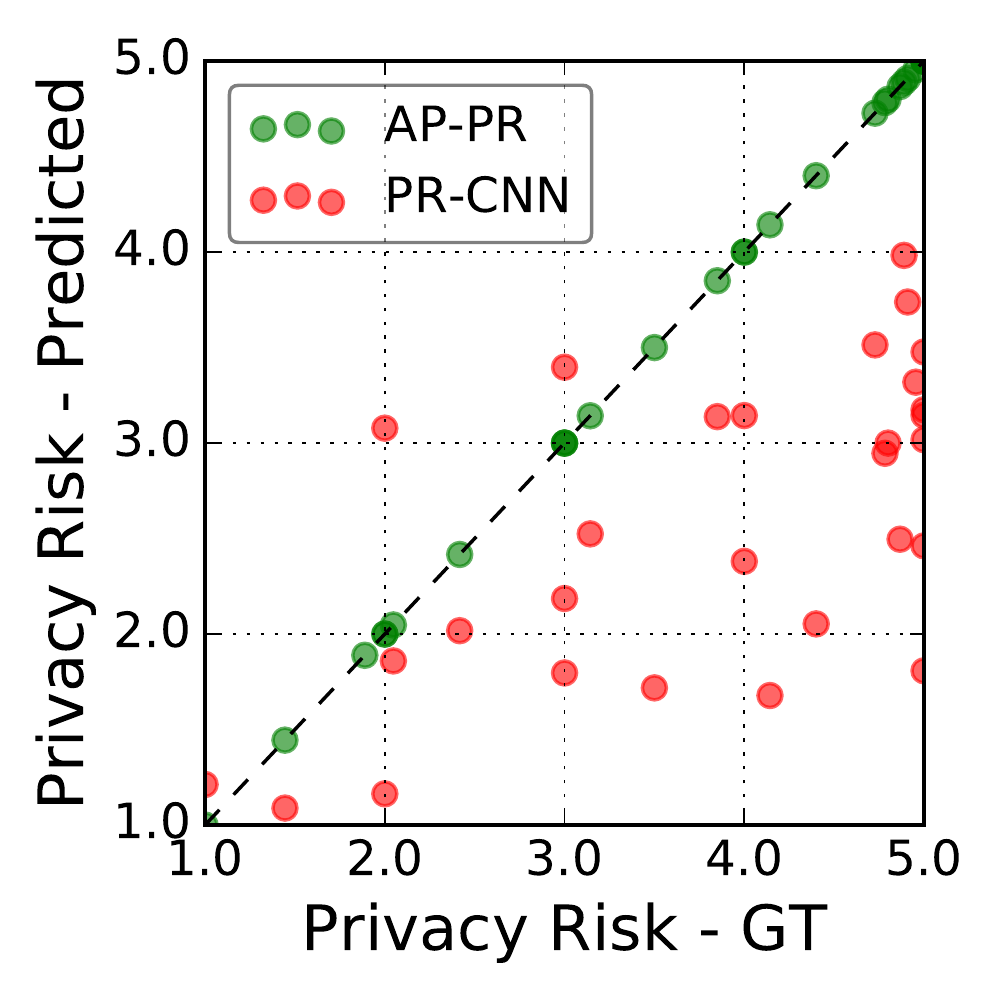} 
 & 
 &  \includegraphics[width=0.15\textwidth]{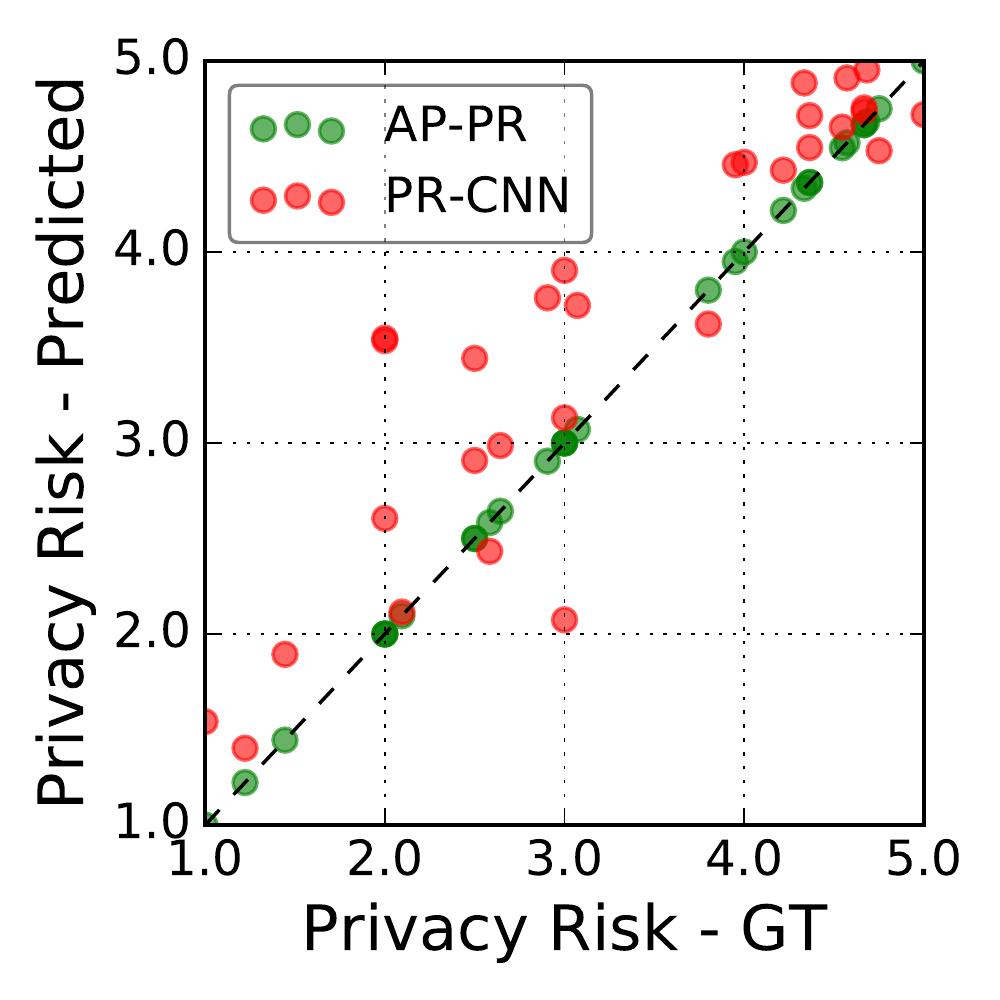} 
 & 
 & \includegraphics[width=0.15\textwidth]{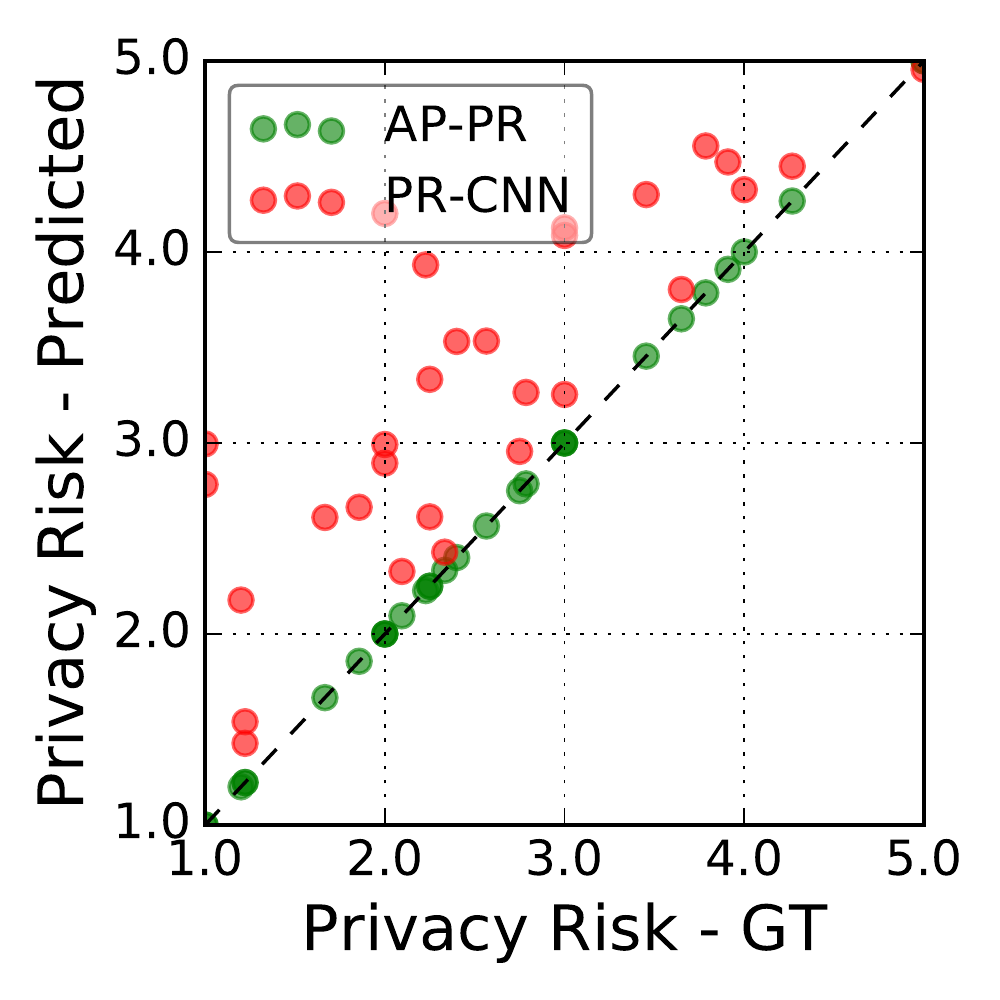} 
 & 
 & \includegraphics[width=0.15\textwidth]{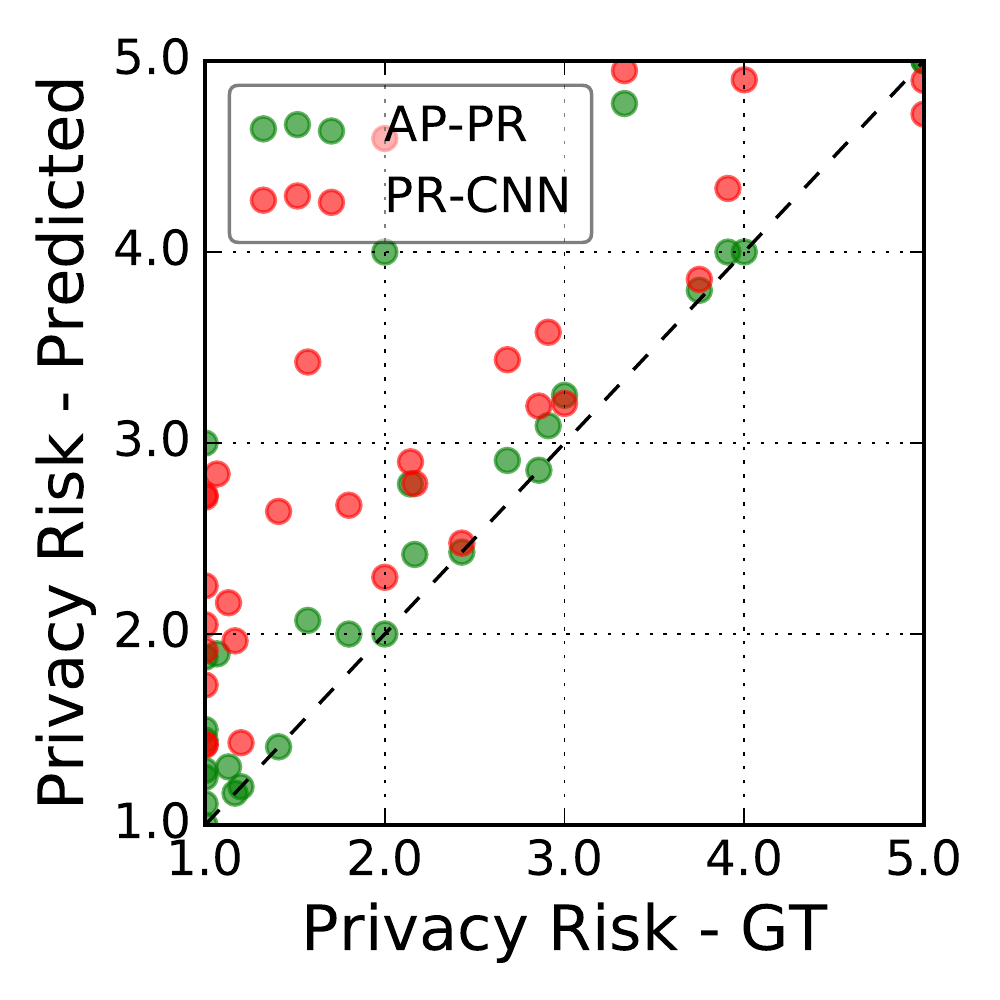}  \\
 & & & & & & \\
\end{tabular}
\caption{Qualitative results for Personalized Privacy Risk Prediction}
\label{fig:ppp_qual}
\end{figure*}

\autoref{fig:ppp_qual} presents qualitative results for our approach to user-specific \emph{Personalized Privacy Risk Prediction} discussed in Section 5.2.
To visualize the qualitative results over all 30 user profiles simultaneously, we present a scatter plot of ground-truth vs. predicted scores for each image.
Each point in the scatter plot represents one user-profile.
In these plots, points closer to the diagonal (dotted line) indicate lower errors.
Points above the diagonal indicate risk over-estimation and under the diagonal indicate risk under-estimation.

We observe from the qualitative results and w.r.t each row in \autoref{fig:ppp_qual}:
\begin{enumerate*}[label=(\roman*)]
	\item (First row) presents examples with correct high confidence attribute predictions according to the posterior probability. Here, both AP-PR and PR-CNN perform equally well.
    \item (Second row) presents examples where attribute predictions are noisy. In these, PR-CNN outperforms AP-PR.
    \item (Third row) Both AP-PR and PR-CNN are challenged by difficult images (low contrast, unnatural angles, low lighting, occlusion). However, we see that PR-CNN often performs slightly better than AP-PR in these cases.
    \item (Fourth row) presents examples where AP-PR with correct attribute predictions performs better than PR-CNN.
\end{enumerate*}

\begin{figure*}
\begin{center}
   \includegraphics[width=\textwidth]{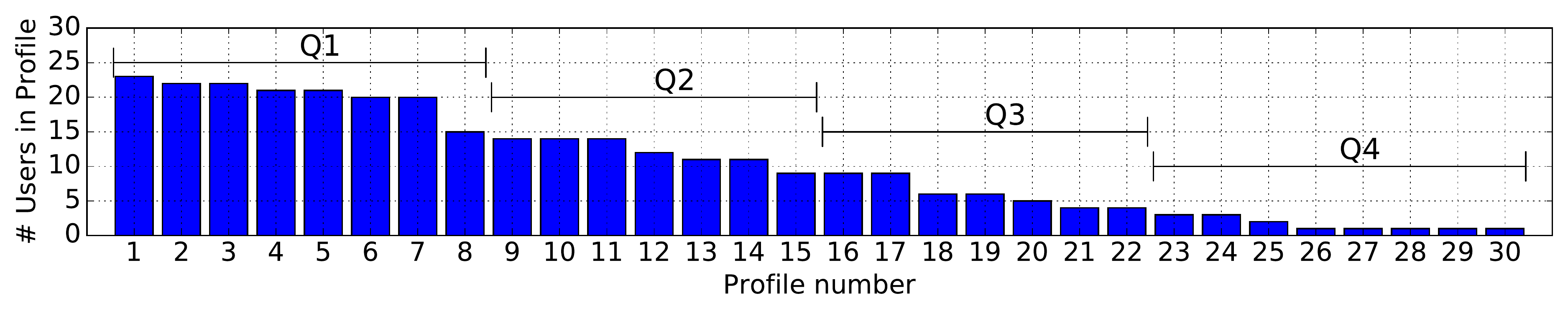}
   \includegraphics[width=\textwidth]{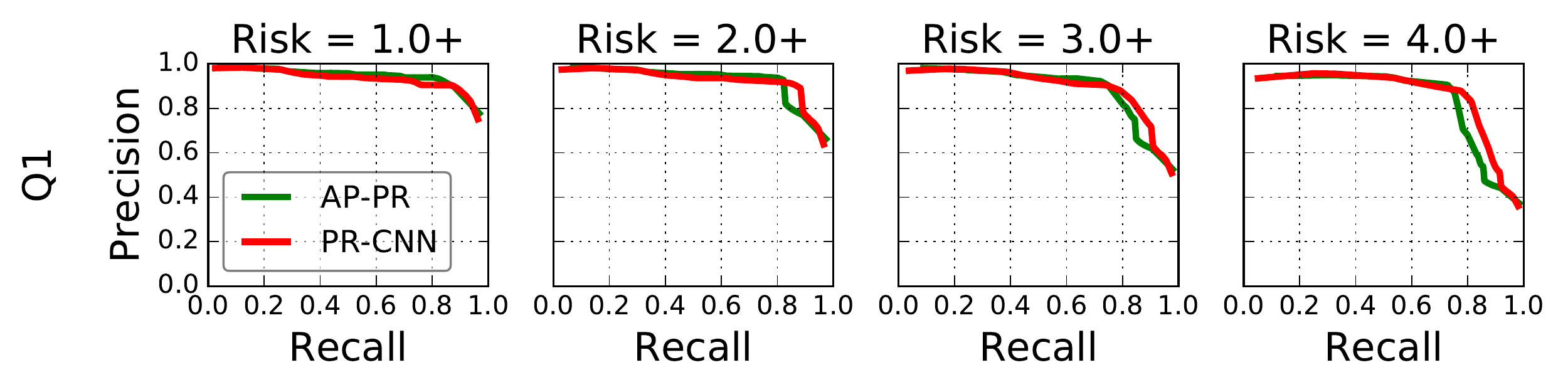}
   \includegraphics[width=\textwidth]{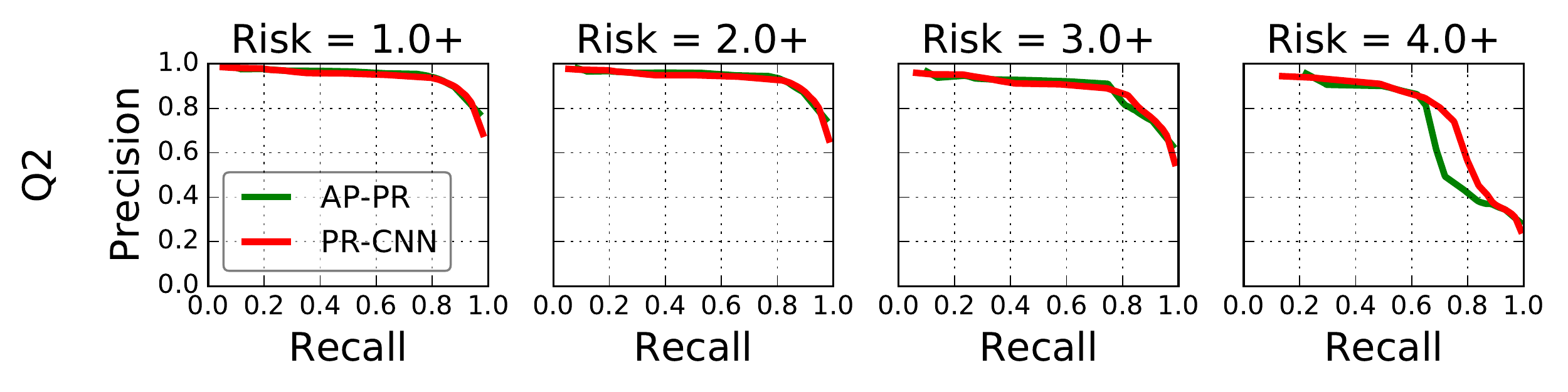}
   \includegraphics[width=\textwidth]{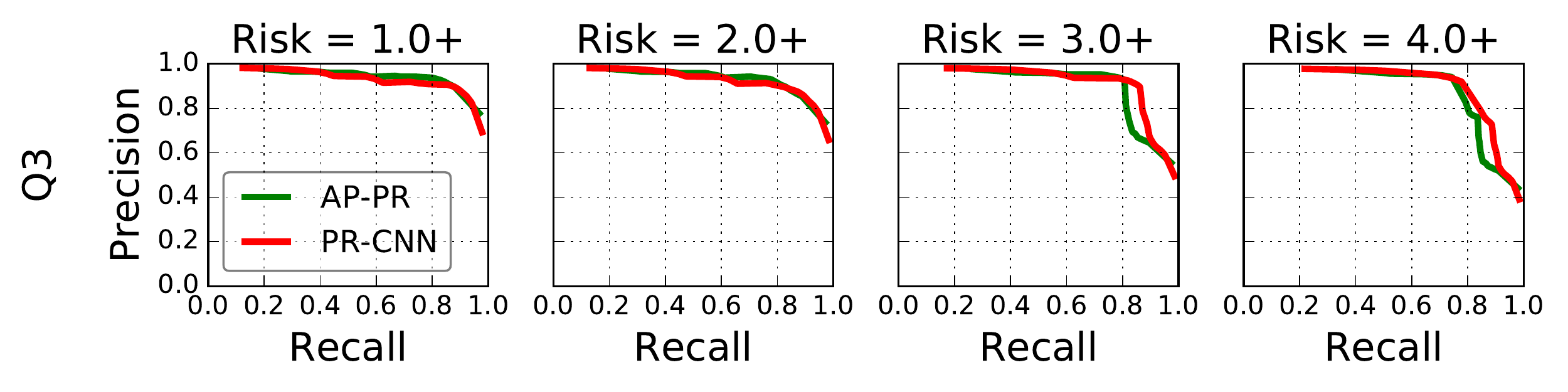}
   \includegraphics[width=\textwidth]{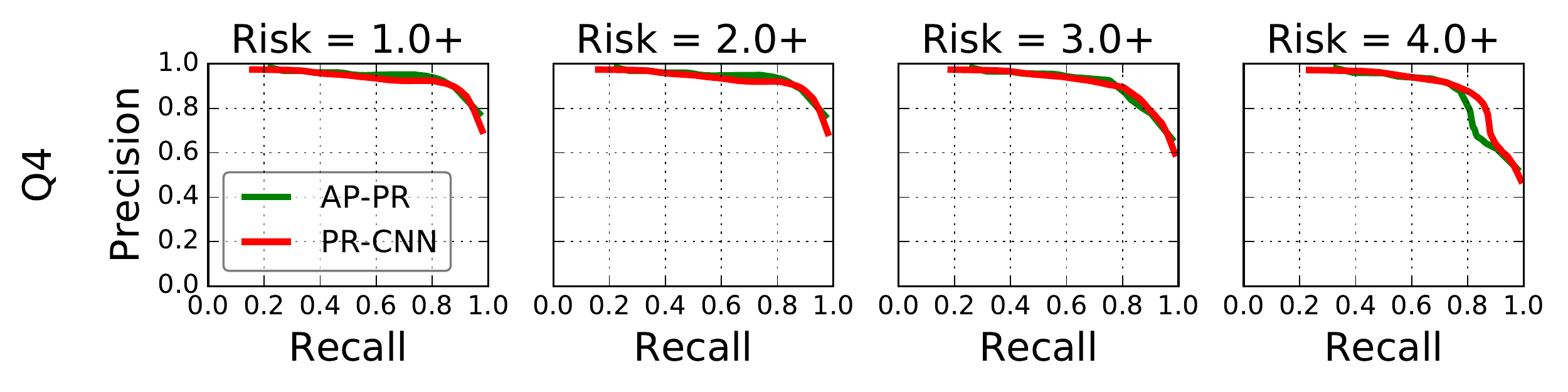}
\end{center}
   \caption{Precision-Recall curves when visualized over groups of user profiles}
\label{fig:pap_prec_recall_sup}
\end{figure*}

\subsection{Precision-Recall Curves for User Profiles}

Section 5.2 discussed Precision-Recall curves evaluated over all profiles.
These were obtained by treating the privacy risk-prediction as a binary classification problem, where images above a certain risk score (3+ and 4+ previously) is considered private per user profile.

In \autoref{fig:pap_prec_recall_sup}, we present the Precision-Recall curves evaluated over groups of profiles and additional risk thresholds.
To generate the curves in these figures, we first create four groups of profiles, with an equal number of profiles in each group.
We refer to these groups as quartiles Q1-Q4. 
We then obtain the Precision-Recall curves for each of these quartiles.

We observe that PR-CNN displays better performance for high-risk images over \emph{all} quartiles of the 30 user profiles and hence contributing to an overall better performance.

Additionally, we observe a similar pattern with the $L_1$-error metric (the absolute difference in scores), where PR-CNN (error = 0.67) incurs lower error in scores for private images compared to AP-PR (error = 0.84).
However, AP-PR (error = 0.34) performs better for safe images in comparison to PR-CNN (error = 0.58).

\section{Additional Results for Humans vs. Machine}
\label{sec:appendix_humans_machines}

In Section 5.3, we discussed the performance of our Privacy Risk Evaluation Methods when compared to the users themselves.
The performance evaluation was primarily with Precision-Recall curves.

In this section, we discuss performance when evaluated using $L_1$ as a distance metric between the ground-truth privacy scores (user's specified preferences) and the privacy risk estimation using three approaches (user's visual risk assessment and our two proposed approaches -- AP-PR and PR-CNN).
The $L_1$ distance here measures the absolute difference in risk score (where risk scores are between 1--5).
\autoref{fig:l1_attributes} presents these errors per attribute.

We observe from these results: 
\begin{enumerate*}[label=(\roman*)]
	\item On average (horizontal lines), the PR-CNN estimates privacy risks  ($L_1$ error = 1.03) slightly better than the user's image-based judgment ($L_1$ error = 1.1)
    \item Users often misjudge the risk (right end of figure) from natural-looking images such as cars with visible license plates or family photographs depicting relationships. In these cases, PR-CNN is better at evaluating risks.
    \item Considering the attributes in which AP-PR incurs high errors (\eg relationships, addresses, username, signature, credit card), we see that PR-CNN outperforms in all these cases bypassing incorrect attribute predictions.
\end{enumerate*}


\end{document}